\documentclass{article}

\usepackage{microtype}
\usepackage{graphicx}
\usepackage{subfigure}
\usepackage{booktabs} 
\usepackage[dvipsnames, svgnames, x11names]{xcolor}

\usepackage{hyperref}

\usepackage[accepted]{icml2024}

\usepackage{amsmath}
\usepackage{amssymb}
\usepackage{mathtools}
\usepackage{amsthm}
\usepackage{mathtools}

\usepackage[capitalize,noabbrev]{cleveref}

\usepackage{url}
\usepackage{csquotes}
\usepackage{multirow}
\usepackage{makecell}
\usepackage[T1]{fontenc}
\usepackage{amsmath,amsthm}
\usepackage{amssymb,mathrsfs}
\usepackage{bm}
\usepackage{algorithm}
\usepackage{algorithmic}
\usepackage{bbding}
\usepackage{enumitem}
\usepackage{tcolorbox}

\theoremstyle{plain}
\newtheorem{theorem}{Theorem}[section]
\newtheorem{proposition}[theorem]{Proposition}

\theoremstyle{definition}

\theoremstyle{remark}

\usepackage[textsize=tiny]{todonotes}

\icmltitlerunning{Configurable Mirror Descent: Towards a Unification of Decision Making}

\begin{document}

\twocolumn[
\icmltitle{Configurable Mirror Descent: Towards a Unification of Decision Making}



\icmlsetsymbol{equal}{*}

\begin{icmlauthorlist}
\icmlauthor{Pengdeng Li}{a}
\icmlauthor{Shuxin Li}{a,equal}
\icmlauthor{Chang Yang}{b,equal}
\icmlauthor{Xinrun Wang}{a}
\icmlauthor{Shuyue Hu}{c}
\icmlauthor{Xiao Huang}{b}
\icmlauthor{Hau Chan}{d}
\icmlauthor{Bo An}{a,e}
\end{icmlauthorlist}

\icmlaffiliation{a}{Nanyang Technological University}
\icmlaffiliation{b}{The Hong Kong Polytechnic University}
\icmlaffiliation{c}{Shanghai Artifcial Intelligence Laboratory}
\icmlaffiliation{d}{University of Nebraska-Lincoln}
\icmlaffiliation{e}{Skywork AI}

\icmlcorrespondingauthor{Xinrun Wang}{xinrun.wang@ntu.edu.sg}

\icmlkeywords{Machine Learning, ICML}

\vskip 0.3in
]



\printAffiliationsAndNotice{\icmlEqualContribution} 

\begin{abstract}
Decision-making problems, categorized as single-agent, e.g., Atari, cooperative multi-agent, e.g., Hanabi, competitive multi-agent, e.g., Hold'em poker, and mixed cooperative and competitive, e.g., football, are ubiquitous in the real world. Various methods are proposed to address the specific decision-making problems. Despite the successes in specific categories, these methods typically evolve independently and cannot generalize to other categories. Therefore, a fundamental question for decision-making is: \emph{Can we develop \textbf{a single algorithm} to tackle \textbf{ALL} categories of decision-making problems?} There are several main challenges to address this question: i) different decision-making categories involve different numbers of agents and different relationships between agents, ii) different categories have different solution concepts and evaluation measures, and iii) there lacks a comprehensive benchmark covering all the categories. This work presents a preliminary attempt to address the question with three main contributions. i) We propose the generalized mirror descent (GMD), a generalization of MD variants, which considers multiple historical policies and works with a broader class of Bregman divergences. ii) We propose the configurable mirror descent (CMD) where a meta-controller is introduced to dynamically adjust the hyper-parameters in GMD conditional on the evaluation measures. iii) We construct the \textsc{GameBench} with 15 academic-friendly games across different decision-making categories. Extensive experiments demonstrate that CMD achieves empirically competitive or better outcomes compared to baselines while providing the capability of exploring diverse dimensions of decision making.
\end{abstract}

\section{Introduction}

Decision-making problems are pervasive in the real world \cite{sutton2018reinforcement,shoham2008multiagent}, which can be generally categorized into single-agent, e.g., Atari~\cite{mnih2015human}, cooperative multi-agent, e.g., Hanabi game~\cite{bard2020hanabi}, competitive multi-agent, e.g., Hold'em poker~\cite{brown2018superhuman,brown2019superhuman}, and mixed cooperative and competitive (MCC), e.g., football~\cite{kurach2020google,liu2022motor}. To solve these problems, various methods are proposed where notable examples include PPO~\cite{schulman2017proximal} for single-agent category, QMIX~\cite{rashid2018qmix} for cooperative multi-agent category and PSRO~\cite{lanctot2017unified} for competitive category. Despite the successes in specific categories, these methods are developed almost independently and cannot generalize to other categories. Therefore, a fundamental question for decision making to answer is:
\begin{displayquote}
\vspace{-10pt}
\emph{Can we develop \textbf{a single algorithm} to  tackle \textbf{ALL} categories of decision-making problems?}
\vspace{-10pt}
\end{displayquote}

\begin{figure}[htbp]
    \centering
    \includegraphics[width=0.4\textwidth]{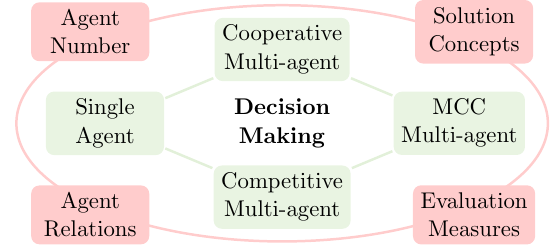}
    \caption{Overview of the categories of decision making and the four desiderata for the required method to satisfy.}
    \label{fig:overall_dm}
\end{figure}

There are several critical challenges to address this fundamental question. First, the different categories of decision-making problems include different numbers of agents and different relationships between agents. There is one agent for the single-agent category, while multiple agents for the other three categories, therefore, the reinforcement learning methods, e.g., PPO, mainly developed for single-agent decision-making problems, cannot be directly applied to multi-agent categories. Furthermore, even for multi-agent categories, QMIX~\cite{rashid2018qmix} is developed to handle the cooperative multi-agent category and cannot be applied to the competitive category. Second, different decision-making categories have different solution concepts, where the optimal (joint) policy is considered in the single-agent and cooperative multi-agent categories, while for the competitive and MCC multi-agent categories, Nash equilibrium (NE)~\cite{nash1951non} is the canonical solution concept and other solution concepts, e.g., correlated equilibrium~\cite{aumann1987correlated} are also considered. Furthermore, even for one solution concept, e.g., NE, there are different evaluation measures, e.g., NashConv or NashConv with social welfare and fairness\footnote{This is related to the equilibrium selection problem~\cite{harsanyi1988general} and different measures lead to different equilibria.}. To summarize the challenges, we propose the four desiderata that the methods should satisfy:
\begin{tcolorbox}[left=0cm, right=0cm, bottom=0cm, top=0cm, colback=red!5, colframe=red!20, boxrule=1.5pt]
\begin{itemize}[left=0.2mm, itemsep=0mm, parsep=0.5mm]
    \item \textbf{D1}: Applicable to single- and multi-agent categories
    \item \textbf{D2}: Applicable to coop., comp., \& MCC categories
    \item \textbf{D3}: Applicable to different solution concepts
    \item \textbf{D4}: Applicable to different evaluation measures
\end{itemize}
\end{tcolorbox}
An overall illustration of the categories of the decision making and the desiderata is displayed in Figure~\ref{fig:overall_dm}. Third, existing benchmarks are typically specialized for specific decision-making categories, while a comprehensive benchmark that satisfies the following two desiderata is lacking.
\begin{tcolorbox}[left=0cm, right=0cm, bottom=0cm, top=0cm, colback=red!5, colframe=red!20, boxrule=1.5pt]
\begin{itemize}[left=0.2mm, itemsep=0mm, parsep=0.5mm]
    \item \textbf{D5}: (Comprehensive) It covers all categories
    \item \textbf{D6}: (Academic-friendly) It is less resource-intensive
\end{itemize}
\end{tcolorbox}

In this work, we make a preliminary attempt to address these challenges and provide three main contributions. i) We propose the generalized mirror descent (GMD), a generalization of existing MD algorithms~\cite{nemirovskij1983problem,beck2003mirror}, which incorporates multiple historical policies into the policy updating and is able to explore a broader class of Bregman divergence by addressing the Karush–Kuhn–Tucker (KKT) conditions at each iteration. As GMD is adopted by each agent independently, it can be applied to different decision-making categories involving different numbers of agents and different relationships between agents (\textbf{D1} and \textbf{D2}). ii) We propose the configurable mirror descent (CMD) by introducing a meta-controller to dynamically adjust the hyper-parameters in GMD conditional on the evaluation measures, allowing us to study different solution concepts as well as evaluation measures (\textbf{D3} and \textbf{D4}), with minimal modifications. iii) We construct the \textsc{GameBench} consisting of 15 games which cover all the decision-making categories (\textbf{D5}) and are deliberately constructed with the principle that running algorithms on these games does not require much computational resource (\textbf{D6}), and hence, forming a comprehensive and academic-friendly testbed for researchers to efficiently develop and test novel algorithms. Extensive experiments on the \textsc{GameBench} show that CMD achieves empirically competitive or better outcomes compared to baselines while offering the ability to investigate diverse dimensions of decision making.

\section{A Real-World Motivating Scenario}

We provide an illustrative example to highlight the importance and real-world implications of a unified algorithm framework. Consider that a robotic company is developing and selling \textit{generalist domestic robots} to users. The user may ask the robot to learn to complete different novel tasks, including single-agent, cooperative, competitive, and MCC categories, by specifying the objective. Therefore, if we can deploy a unified algorithm into the robot, the robot can learn to complete different novel tasks with a single algorithm. 

Developing and deploying such a unified algorithm would benefit both the development and users. \textit{For the development side}, as only a single policy learning rule is required, the deployment and user interface design could be largely simplified, which would be more cost-efficient than deploying different specialized algorithms such as MAPPO and PSRO as they may complicate the development pipeline and user interface design. \textit{For the user side}, the user only needs to configure one set of parameters for different novel tasks, e.g., only needs to specify the optimization objective of the meta-controller in our proposed CMD algorithm.

\section{Related Work\label{sec:related_work}}

The related literature is too vast to cover in its entirety. We present an overview below to emphasize our contributions while more related works can be found in Appendix~\ref{app:more_related_works}.

\textbf{Decision Making.} Substantial progress has been achieved in developing algorithms to address different categories of decision-making problems, e.g., DQN~\cite{mnih2015human} and PPO~\cite{schulman2017proximal} for single-agent category, QMIX~\cite{rashid2018qmix} and MAPPO~\cite{yu2022surprising} for cooperative multi-agent category, self-play~\cite{tesauro1995temporal} and PSRO~\cite{lanctot2017unified} for competitive and MCC categories, to name just a few. Despite the successes in specific categories, these methods often cannot directly generalize to different categories. In this work, we make a preliminary attempt to develop \textit{a single algorithm} capable of \textit{tackling all categories of decision-making problems} which typically involve different numbers of agents, different relationships between agents, different solution concepts as well as different evaluation measures. 

\textbf{Mirror Descent.} Mirror descent (MD)~\cite{nemirovskij1983problem,beck2003mirror,vural2022mirror} has shown effectiveness in learning optimal policies in single-agent RL~\cite{tomar2022mirror} and proved the last-iterate convergence in learning approximate equilibrium in zero-sum games~\cite{bailey2018multiplicative,kangarshahi2018let,wibisono2022alternating,kozuno2021learning,lee2021lastiterate,jain2022matrix,ao2023asynchronous,liu2023the,cen2023faster,sokota2023unified} and some classes of general-sum games, e.g., polymatrix and potential games~\cite{anagnostides2022last}. Despite the progress, existing works typically focus on some specific Bregman divergence such as the KL divergence. We relax this premise by addressing the KKT conditions at each iteration, enabling us to \textit{explore a broader class of Bregman divergence}. Moreover, by introducing a meta-controller to dynamically adjust the hyper-parameters, our CMD can be applied to \textit{different solution concepts and evaluation measures} with minimal modifications.

\textbf{Hyper-Parameter Tuning.} Existing works typically determine the hyper-parameter values of the MD algorithms depending on domain knowledge~\cite{sokota2023unified,anagnostides2022last,hsieh2021adaptive}, which may not be easy to generalize to different games. On the other hand, gradient-based hyper-parameter tuning methods such as STAC~\cite{zahavy2020self} are less applicable as the evaluation measures, e.g., NashConv, could be non-differentiable with respect to the hyper-parameters. To address the issue, we propose a simple yet effective zero-order optimization method where the performance difference between two candidates is used to only determine the \textit{update direction} of the hyper-parameters rather than the update magnitude, which is more effective than existing methods~\cite{wang2022zarts} when the value of the performance is extremely small.

\section{Preliminaries}

In this section, we first introduce the model of decision making and the solution concepts and evaluation measures considered in our work. Then, we present the classic mirror descent algorithm~\cite{beck2003mirror}.

\subsection{Decision Making\label{sec:problem_setting}}
\textbf{POSG.} A decision-making problem, either single-agent, cooperative, competitive, or mixed cooperative and competitive category, can be described as a partially observable stochastic game (POSG)~\cite{oliehoek2016concise} denoted as $\langle\mathcal{N}, \mathcal{S}, \mathcal{A}, \mathcal{O}, \Omega,  P, R, \gamma, \nu\rangle$. $\mathcal{N}=\{1, \cdots, N\}$ is the set of agents. $\mathcal{S}$ is the \textit{finite} set of the states. $\mathcal{A}=\times_{i\in\mathcal{N}}\mathcal{A}_{i}$ and $\mathcal{O}=\times_{i\in\mathcal{N}}\mathcal{O}_{i}$ where $\mathcal{A}_{i}$ and $\mathcal{O}_{i}$ are the \textit{finite} set of actions and observations of agent $i$, respectively. Let $\bm{a}\in\mathcal{A}$ denote the joint action of agents where $a_{i}\in\mathcal{A}_{i}$ is agent $i$'s action. $\Omega=\times_{i\in\mathcal{N}}\Omega_{i}$ where $\Omega_{i}:\mathcal{S}\times\mathcal{A}\rightarrow\mathcal{O}_{i}$ is the observation function specifying agent $i$'s observation $o_{i}\in\mathcal{O}_{i}$ when all agents take $\bm{a}\in\mathcal{A}$ at state $s\in\mathcal{S}$. $P:\mathcal{S}\times\mathcal{A}\rightarrow\Delta(\mathcal{S})$ is the transition function which specifies the probability of transiting to $s^{\prime}\in\mathcal{S}$ when agents take $\bm{a}\in\mathcal{A}$ at state $s\in\mathcal{S}$. $\Delta(\cdot)$ denotes the simplex. $R=\{r_{i}\}_{i\in\mathcal{N}}$ where $r_{i}:\mathcal{S}\times\mathcal{A}\rightarrow\mathbb{R}$ is the reward function of agent $i$ and $\gamma\in[0, 1)$ is the discount factor. $\nu\in\Delta(\mathcal{S})$ denotes the distribution over initial states. At time step $t\ge0$, each agent has an action-observation history (i.e., a decision point) $\tau_{i}^{t}\in\mathcal{T}_{i}^{t}$ where $\mathcal{T}_{i}^{t}=(\mathcal{O}_{i}\times\mathcal{A}_{i})^{t}$ and constructs its policy $\pi_{i}:\mathcal{T}_{i}^{t}\rightarrow\Delta(\mathcal{A}_{i})$ to maximize its own return. Let $\Pi_{i}$ denote the policy space of agent $i$, that is, we have $\pi_{i}\in\Pi_{i}$. The joint policy of all agents is denoted as $\bm{\pi}=\pi_1\odot\cdots\odot\pi_N$ and $\bm{\pi}\in \Pi$ where $\Pi$ denotes the joint policy space of all agents. A special case of joint policy is the \textit{product policy} denoted as $\bm{\pi}=\pi_1\times\cdots\times\pi_N$. Also, let $\bm{\pi}_{-i}=\pi_1\odot\cdots\pi_{i-1}\odot\pi_{i+1}\cdots\odot\pi_N$ denote the joint policy of all agents except $i$. Given the initial state $s_0=s$, the value function of agent $i$ is $V_{i}(s, \bm{\pi})\coloneqq\mathbb{E}[\sum_{t=0}^{\infty}\gamma^{t}r_{i}^{t}|s, \bm{\pi}]$ where $r_{i}^{t}$ is the agent $i$'s reward at time $t\ge 0$. Furthermore, we have $V_i(\nu, \bm{\pi})\coloneqq\mathbb{E}_{s\sim\nu}\left[V_{i}(s, \bm{\pi})\right]$. 

\textbf{Solution Concepts.} The policy of an agent is said to be optimal if it is optimal in every decision point belonging to the agent. In single-agent and cooperative categories, this \textit{optimal policy} maximizes the expected return for the agent or the team. In multi-agent competitive and mixed cooperative and competitive categories, we consider two common equilibrium concepts: \textit{Nash equilibrium (NE)}~\cite{nash1951non} and \textit{coarse correlated equilibrium (CCE)}~\cite{moulin1978strategically}. Let $\pi_i\times\bm{\pi}_{-i}$ denote the \textit{product policy} and $\pi_i\odot\bm{\pi}_{-i}$ denote the joint policy. Then, $\bm{\pi}^{*}$ is called an NE if for each agent $i$ it satisfies: $\forall\pi_i^{\prime}\in\Pi_i$, $V_i(\nu, \bm{\pi}^{*}) \ge V_i(\nu, \pi_i^{\prime}\times\bm{\pi}_{-i}^{*})$. Similarly, $\bm{\pi}^{*}$ is called a CCE if for each agent $i$ it satisfies: $\forall\pi_i^{\prime}\in\Pi_i$, $V_i(\nu, \bm{\pi}^{*}) \ge V_i(\nu, \pi_i^{\prime}\odot\bm{\pi}_{-i}^{*})$.

\textbf{Evaluation Measures.} Let $\mathcal{L}(\bm{\pi})$ denote measures used to evaluate a (joint) policy $\bm{\pi}$. In single-agent and cooperative categories, the measure is the distance of the (joint) policy to the optimal (joint) policy $\bm{\pi}^{*}$, which is defined as $\mathcal{L}(\bm{\pi})=\text{OptGap}(\bm{\pi}) = V(\nu, \bm{\pi}^{*}) - V(\nu, \bm{\pi})$. In other categories, we consider multiple evaluation measures. The first one is the distance of the joint policy to the equilibrium (NE or CCE). For NE, we refer to this distance as NashConv, and for CCE, we refer to it as CCEGap, as is convention in previous works~\cite{lanctot2017unified,marris2021multi}. More specifically, we have $\text{NashConv}(\bm{\pi}) = \sum_{i\in\mathcal{N}}[V_{i}(\nu, \pi_{i}^{\text{BR}}\times\bm{\pi}_{-i}) - V_{i}(\nu, \bm{\pi})]$ and $\text{CCEGap}(\bm{\pi}) = \sum_{i\in\mathcal{N}}[V_{i}(\nu, \pi_{i}^{\text{BR}}\odot\bm{\pi}_{-i}) - V_{i}(\nu, \bm{\pi})]$, where $\pi_{i}^{\text{BR}}$ is the best response (BR) policy of agent $i$ against all other agents. The second evaluation measure we consider is the social welfare (SW)~\cite{davis1962externalities}, denoted as $\mathcal{L}(\bm{\pi}) = \sum_{i\in\mathcal{N}}V_{i}(\nu, \bm{\pi})$. 

\begin{table*}[ht]
    \centering
    \caption{Comparison of different methods. $^*$Note that MMD can be regarded as the method that can consider multiple previous policies by setting the magnet policy to the initial policy (typically a uniform policy).}
    \label{tab:compare_table}
    \vskip 0.1in
    \begin{tabular}{l|c|c|c|c}
    \toprule
    \multirow{2}*{Method} & Multiple previous & Working on any  & Working on any&Configurable for \\
         & policies & Bregman divergence & solution concept&any measure\\
    \midrule
    MD~\cite{nemirovskij1983problem}  & {\color{red!75}\XSolidBrush} & {\color{red!75}\XSolidBrush} &{\color{red!75}\XSolidBrush} &{\color{red!75}\XSolidBrush} \\
    MMD$^*$~\cite{sokota2023unified} & {\color{Green3}\Checkmark} & {\color{red!75}\XSolidBrush} & {\color{red!75}\XSolidBrush}&{\color{red!75}\XSolidBrush} \\
    GMD (This work) & {\color{Green3}\Checkmark} & {\color{Green3}\Checkmark} & {\color{red!75}\XSolidBrush} &{\color{red!75}\XSolidBrush} \\
    CMD (This work) & {\color{Green3}\Checkmark} & {\color{Green3}\Checkmark} & {\color{Green3}\Checkmark}&{\color{Green3}\Checkmark}\\
    \bottomrule
    \end{tabular}
\end{table*}
\subsection{Mirror Descent}
From a single agent's perspective, the condition for the optimal or equilibrium policy can be expressed by the following optimization problem at each decision point of the agent~\cite{tomar2022mirror,sokota2023unified}: $\forall \tau_{i}^{t} \in \mathcal{T}_{i}^{t}$,
\begin{equation}
    \pi_{i}^{*}(\tau_{i}^{t}) = \operatorname*{arg\,max}_{\pi_{i} \in \Pi_{i}} \mathbb{E}_{a \sim \pi_{i}(\tau_{i}^{t})} Q(\tau_{i}^{t}, a, \pi_{i}\odot\bm{\pi}_{-i}^{*}), \label{eq:ne_condition}
\end{equation}
where $Q(\tau_{i}^{t}, a, \bm{\pi}) = \mathbb{E}[\sum_{h=t+1}^{\infty}\gamma^{h}r_{i}^{h}|\tau_{i}^{t}, a_i^t = a, \bm{\pi}]$ is the action-value function of the action $a\in\mathcal{A}_i$ at the decision point $\tau_{i}^{t}$. Without loss of generality, we will only focus on the policy learning of a single agent $i$ in a single decision point $\tau_{i}^{t}\in\mathcal{T}_{i}^{t}$ and henceforth, the index $i$ and $\tau_{i}^{t}$ are ignored as they are clear from the context, and with a slight abuse of notation, we denote $\mathcal{A}$ the action set $\mathcal{A}_i$ of agent $i$, $\pi \in \Pi$ the agent's policy, and $Q(a)$ (or $Q(a, \pi)$) the action-value of the action $a\in\mathcal{A}$. Then, to learn the optimal or equilibrium policy, we aim to solve the optimization problem: $\pi^{*} = \arg\max\nolimits_{\pi \in \Pi} \mathbb{E}_{a \sim \pi} Q(a)$. A feasible method to solve this problem is the mirror descent (MD), which takes the form~\cite{beck2003mirror,tomar2022mirror}:
\begin{equation}
\begin{aligned}
    &\pi_{k+1} = \arg\max\nolimits_{\pi\in\Pi} \langle Q(\pi_{k}), \pi\rangle  - \alpha \mathcal{B}_{\phi}(\pi, \pi_{k}),
\end{aligned}
\label{eq:mirror-descent}
\end{equation}
where $1\leq k \leq K$ is the iteration, $Q(\pi_{k})$ is the action-value vector induced by $\pi_{k}$ (for simplicity, we let $Q(\pi_{k})=(Q(a, \pi_{k}))_{a\in\mathcal{A}}$), $\alpha$ is the regularization intensity, $\mathcal{B}_{\phi}$ is the Bregman divergence with respect to the mirror map $\phi$, defined as $\mathcal{B}_{\phi}(x;y)=\phi(x)-\phi(y)-\langle\nabla\phi(y), x-y\rangle$ with $\langle\cdot\rangle$ being the standard inner product and $x, y\in \Delta(\mathcal{A})$.

\section{Configurable Mirror Descent}

\begin{figure}[htbp]
    \centering
    \includegraphics[width=\columnwidth]{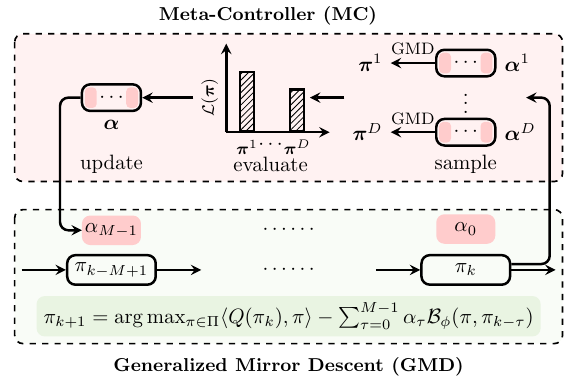}
    \vspace{-15pt}
    \caption{Overview of CMD.}
    \label{fig:cmd-overview}
\end{figure}

In this section, we propose a novel algorithm which satisfies the four desiderata (\textbf{D1}--\textbf{D4}) presented in the Introduction. First, we propose the generalized mirror descent (GMD), a generalization of existing MD algorithms, which when independently executed by each agent, can effectively tackle different decision-making categories involving different numbers of agents and different relationships between agents (\textbf{D1} and \textbf{D2}). Second, we propose the configurable mirror descent (CMD) where a meta-controller is introduced to dynamically adjust the hyper-parameters of GMD conditional on the evaluation measures, which can be configured to account for different solution concepts as well as evaluation measures (\textbf{D3} and \textbf{D4}), with minimal modifications. CMD shares similarities with the centralized training and decentralized execution (CTDE)~\cite{lowe2017multi} since the meta-controller considers all agents to optimize the targeted evaluation measures (\enquote{centralized} training from the controller's perspective) while GMD is executed by each agent independently (\enquote{decentralized} execution from each agent's perspective). The overview of CMD is shown in Figure~\ref{fig:cmd-overview} and Table~\ref{tab:compare_table} presents a comparison to more clearly position our methods in the context of related literature.

\subsection{Generalized Mirror Descent \label{sec:gmd}}

Though existing MD algorithms, e.g., Eq. (\ref{eq:mirror-descent}), can be also executed by each agent independently, they could not generalize well to satisfy the desiderata \textbf{D1} and \textbf{D2}. The main reasons are two-fold. First, classic MD algorithms~\cite{beck2003mirror,tomar2022mirror} typically only consider the current policy when deriving the new policy at each iteration. However, it has been shown that incorporating multiple previous policies (e.g., the initial and current policies) could be powerful in solving two-player zero-sum games~\cite{sokota2023unified,liu2023the}. Second, even though multiple previous policies are considered, existing MD algorithms typically focus on some specific Bregman divergences by restricting $\phi$ to certain convex functions which may not be the optimal choices across different decision-making categories. To address these challenges, we propose a more general MD method satisfying the desiderata \textbf{D1} and \textbf{D2}.

\textbf{A General MD Method.} We propose a more general MD approach which takes multiple historical policies into account when deriving the new policy, as given below:
\begin{align}
    & \pi_{k+1} = \operatorname*{arg\,max}_{\pi \in \Pi} \langle Q(\pi_{k}), \pi\rangle  - \sum\nolimits_{\tau = 0}^{M-1} \alpha_{\tau}\mathcal{B}_{\phi} (\pi, \pi_{k-\tau}), \nonumber\\
    & \textit{s.t. } \sum\nolimits_{a\in\mathcal{A}}\pi_k(a) = 1 \text{ and } \pi_k(a)\geq 0, \forall a \in\mathcal{A},\label{eq:gmd}
\end{align}
where $M\ge1$ is the number of historical policies, $\alpha_{\tau} \in (0, 1]$ is the regularization intensity of $\pi_{k-\tau}$, $0\leq\tau\leq M-1$, and let $\bm{\alpha}=(\alpha_{\tau})_{0\leq\tau\leq M-1}$. Note that solving the problem (\ref{eq:gmd}) to derive the policy updating rule could be challenging. In practice, the problem could have a closed-form solution only in certain settings such as the convex function $\phi$ is the negative entropy and the constraints are removed (i.e., the unconstrained domains~\cite{sokota2023unified}). To address this issue, we propose a novel method to solve the problem (\ref{eq:gmd}), which does not rely on the availability of the closed-form solution and hence, can consider more possible options of $\phi$. To this end, first, we have the following result:
\begin{proposition}
\label{prop:kkt_problem}
    Assume that i) $\pi(a) \geq \epsilon$, $\forall a \in \mathcal{A}$, where $\epsilon$ is a small positive value and ii) the $\phi(\pi)$ defined on $\Pi$ can be decomposed to\footnote{The sum of convex functions is still a convex function. Furthermore, the negative entropy and squared Euclidean norm are two special variants that have been extensively adopted in literature.} $\phi(\pi)=\sum_{a\in\mathcal{A}}\psi(\pi(a))$ where $\psi$ is some convex function defined on $[0, 1]$. Then, solving the problem (\ref{eq:gmd}) can be converted to solve the following equation:
    \begin{align}
        \sum\nolimits_{a\in\mathcal{A}}\pi(a) = \sum\nolimits_{a\in\mathcal{A}}\psi^{\prime-1}\left(A(a)-\lambda)/B\right) = 1,
        \label{eq:compute_lambda_main}
    \end{align}
    where $A=Q(\pi_{k}) +\sum_{\tau=0}^{M-1}\alpha_{\tau}\phi^{\prime}(\pi_{k-\tau})$, $B=\sum_{\tau=0}^{M-1} \alpha_{\tau}$, $\lambda$ is the dual variable, $\psi^{\prime-1}$ is the inverse function of $\psi^{\prime}$ (the derivative of $\psi$), and $\pi(a)=\psi^{\prime-1}\left(A(a)-\lambda)/B\right)$.
\end{proposition}
This result is obtained via the Karush–Kuhn–Tucker (KKT) conditions of the Lagrange function obtained by applying the Lagrange multiplier $\lambda$ (i.e., the dual variable) to the problem (\ref{eq:gmd}). The full derivation can be found in Appendix~\ref{app:gmd}.

\textbf{Numerical Method for Computing $\lambda$.} Now we need to solve Eq. (\ref{eq:compute_lambda_main}) to obtain the value of $\lambda$. Unfortunately, this typically cannot be solved \textit{analytically}, rendering it less possible to derive the policy updating rule without the availability of the closed-form solution. To address this issue, we use the Newton method~\cite{ypma1995historical} to compute the value of $\lambda$: repeatedly executing $\lambda = \lambda - g(\lambda)/g^{\prime}(\lambda)$ for $C>0$ iterations, where $g(\lambda)=\left[\sum\nolimits_{a\in\mathcal{A}}\psi^{\prime-1}(A(a)-\lambda)/B)\right] - 1$, $g^{\prime}(\lambda)=\sum\nolimits_{a\in\mathcal{A}}-\frac{1}{B}[\psi^{\prime-1}]^{\prime}(A(a)-\lambda)/B)$, and $[\psi^{\prime-1}]^{\prime}$ is the derivative of $\psi^{\prime-1}$. The pseudo-code can be found in Algorithm~\ref{alg:newton} in Appendix~\ref{app:gmd}.

\textbf{Projection Operation.} After computing the value of $\lambda$, we can get the policy $\pi(a)$ by substituting it into the expression of $\pi(a)$ as presented in Eq. (\ref{eq:compute_lambda_main}). Furthermore, we employ a projection operation to ensure that $\pi(a)\ge \epsilon$. Specifically, we have: $\forall a \in \mathcal{A}$, $\pi_{k+1}(a) = \frac{\max\left\{\epsilon, \pi(a)\right\}}{\sum_{a^{\prime}\in\mathcal{A}} \max\left\{\epsilon, \pi(a^{\prime})\right\}}$.
 
\textbf{Different Bregman Divergences.} In addition to taking multiple historical policies into consideration, GMD\footnote{The term GMD is also used in~\cite{radhakrishnan2020linear}, which differs from our method.} further generalizes existing MD algorithms with the capability of exploring a broader class of Bregman divergence as, via the numerical method to compute the value of $\lambda$, it is capable of taking more possible convex functions into account. Table~\ref{tab:convex_functions} presents the functions considered in our work. See~\cite{boyd2004convex} for more examples. $x^2$ (i.e., $n=2$) and $x\ln x$ respectively correspond to the Euclidean norm and entropy. More details can be found in Appendix~\ref{app:gmd}. 

\begin{table}[ht]
\vspace{-10pt}
\centering
\caption{List of convex functions and related functions, $x\in (0,1]$.}
\label{tab:convex_functions}
\vskip 0.1in
\begin{tabular}{c|ccc}
\toprule
 $\psi(x)$ & $\psi^{\prime}(x)$ & $\psi^{\prime-1}(x)$ & $[\psi^{\prime-1}]^{\prime}(x)$ \\
\midrule
$x^{n}$, $n > 1$ & $nx^{n-1}$ & $(\frac{x}{n})^{\frac{1}{n-1}}$ & $\frac{1}{n-1}(\frac{x}{n})^{\frac{2-n}{n-1}}$ \\
\midrule
$x\ln x$ & $\ln x + 1$& $e^{x-1}$ & $e^{x-1}$ \\
\midrule
\makecell[c]{$-x^{n}$,\\$0<n<1$} & $-nx^{n-1}$& $(\frac{-x}{n})^{\frac{1}{n-1}}$ & $\frac{1}{1-n}(\frac{-x}{n})^{\frac{2-n}{n-1}}$ \\
\midrule
$e^{kx}$, $k>0$ & $ke^{kx}$ & $\ln(x/k)/k$ & $\frac{1}{x}$ \\
\bottomrule
\end{tabular}
\end{table}

\textbf{GMD Summary.} The pseudo-code of GMD is provided in Algorithm~\ref{alg:gmd}. Compared to existing MD algorithms, GMD could satisfy well the desiderata \textbf{D1} and \textbf{D2} as it not only takes multiple previous policies into account but is also capable of leveraging more possible Bregman divergences that may be better than existing choices such as KL divergence across different decision-making categories.

\begin{algorithm}[ht]
\caption{Generalized Mirror Descent (GMD)}
\label{alg:gmd}
\begin{algorithmic}[1]
\STATE Given $\psi$, initial policy $\pi_1$, $M$, $\bm{\alpha}$, $\epsilon$
\FOR{$k=1, \cdots, K$}
\STATE Compute $A$ and $B$ with $\pi_k$ and $\bm{\alpha}$\;
\STATE Compute $\lambda$ via Newton method (Algorithm~\ref{alg:newton})\;
\STATE Compute $\pi(a) = \psi^{\prime-1}(A(a)-\lambda)/B)$, $\forall a \in \mathcal{A}$\;
\STATE Compute $\pi_{k+1}(a)$ via projection operation, $\forall a \in \mathcal{A}$\;
\ENDFOR
\end{algorithmic}
\end{algorithm}

\subsection{Meta-Controller for Different Measures\label{sec:mc}}

While GMD can satisfy the desiderata \textbf{D1} and \textbf{D2}, it cannot satisfy well the last two desiderata \textbf{D3} and \textbf{D4}. The primary reason is that when each agent independently executes GMD, it is not immediately feasible to investigate different solution concepts and evaluation measures as no explicit objective regarding the different measures arises in such a \enquote{decentralized} execution process (different measures could lead to different solution concepts and henceforth, we will only focus on the different measures). To address this problem, we propose the configurable mirror descent (CMD) by introducing a meta-controller (MC) to adjust the hyper-parameters in GMD \textit{conditional on the evaluation measures}, which is a \enquote{centralized} process from the meta-controller's perspective as it considers all the agents (\textit{joint policy}) to optimize the targeted evaluation measure (and hence, the targeted solution concept), i.e., \textbf{D3} and \textbf{D4}.

\textbf{Zero-Order Meta-Controller.} As shown in Eq. (\ref{eq:gmd}), given the number of historical policies $M\ge 1$, the only controllable variable is the hyper-parameters $\bm{\alpha}=(\alpha_{\tau})_{0\leq\tau\leq M-1}$. At the iteration $k$, let $\bm{\pi}=\text{GMD}(\bm{\alpha})$\footnote{$\bm{\alpha}$ is applied to all the agents in multi-agent categories.} denote the \textit{joint policy} derived from the previous \textit{joint policy} $\bm{\pi}_k$ by using GMD with the given $\bm{\alpha}$, and the performance of this joint policy is denoted as $\mathcal{L}(\bm{\pi})$. Notably, optimizing $\bm{\alpha}$, unfortunately, is non-trivial as the evaluation measure $\mathcal{L}$ is non-differentiable with respect to $\bm{\alpha}$. To address this issue, we construct an efficient zero-order MC by leveraging a zero-order method to optimize $\bm{\alpha}$. As shown in Figure~\ref{fig:cmd-overview}, MC updates $\bm{\alpha}$ through three steps: i) it \textit{samples} $D$ candidates $\{\bm{\alpha}^j\}_{j=1}^{D}$ by perturbing the current $\bm{\alpha}$ and then derives $D$ new joint policies $\{\bm{\pi}^j=\text{GMD}(\bm{\alpha}^j)\}_{j=1}^{D}$ by employing GMD, ii) it \textit{evaluates} these new joint policies $\{\mathcal{L}(\bm{\pi}^j)\}_{j=1}^{D}$, and iii) it \textit{updates} $\bm{\alpha}$ based on the performance of these new joint policies.

\textbf{Direction-Guided Update.} Let $\bm{\alpha}^1$ and $\bm{\alpha}^2$ denote the two candidates sampled by perturbing the current $\bm{\alpha}$ and the corresponding joint policies $\bm{\pi}^1$ and $\bm{\pi}^2$ are obtained via GMD. Existing zero-order methods such as the random search (RS)~\cite{liu2020primer} typically update the $\bm{\alpha}$ directly based on the performance difference between the two candidates $\delta=\mathcal{L}(\bm{\pi}^1) - \mathcal{L}(\bm{\pi}^2)$, which could be ineffective as the value of $\mathcal{L}$ could be too small (as shown in our experiments) to derive an effective update. To address this problem, we propose to update $\bm{\alpha}$ based on the \textit{sign} of the performance difference. Precisely, $\delta$ only determines the \textit{update direction}, not the update magnitude, which is more effective when the value of $\mathcal{L}$ is too small. We call this simple yet effective technique the \textit{direction-guided update}. In our experiments, we construct an MC -- \textit{direction-guided random search (DRS)} -- by applying this method to the existing RS~\cite{wang2022zarts}. More details on different MCs can be found in Appendix~\ref{app:meta_controller}. 

\begin{algorithm}
\caption{Configurable Mirror Descent (CMD)}
\label{alg:cmd}
\begin{algorithmic}[1]
\STATE Given $\mathcal{L}$, $\psi$, initial (joint) policy $\bm{\pi}_1$, $M$, $D$, $\epsilon$
\FOR{$k=1, \cdots, K$}
\STATE Sample $D$ candidates $\{\bm{\alpha}^j\}_{j=1}^{D}$\;
\STATE Derive new joint policies $\{\bm{\pi}^j=\text{GMD}(\bm{\alpha}^j)\}_{j=1}^{D}$\;
\STATE Evaluate new joint policies $\{\mathcal{L}(\bm{\pi}^j)\}_{j=1}^{D}$\;
\STATE Update $\bm{\alpha}$ based on $\{\mathcal{L}(\bm{\pi}^j)\}_{j=1}^{D}$\;
\STATE Compute $\bm{\pi}_{k+1}$ via GMD with the updated $\bm{\alpha}$\;
\ENDFOR
\end{algorithmic}
\end{algorithm}
\textbf{CMD Summary.} By incorporating the MC into GMD, we establish the CMD. Intuitively, CMD can be configured to apply to different evaluation measures and hence, can satisfy the desiderata \textbf{D3} and \textbf{D4} while only minimal modifications are required: specifying the MC's optimization objective $\mathcal{L}$. The pseudo-code of CMD is shown in Algorithm~\ref{alg:cmd}.

\section{\textsc{GameBench}\label{sec:gamebench}}
\begin{figure}[ht]
\centering
\includegraphics[width=\columnwidth]{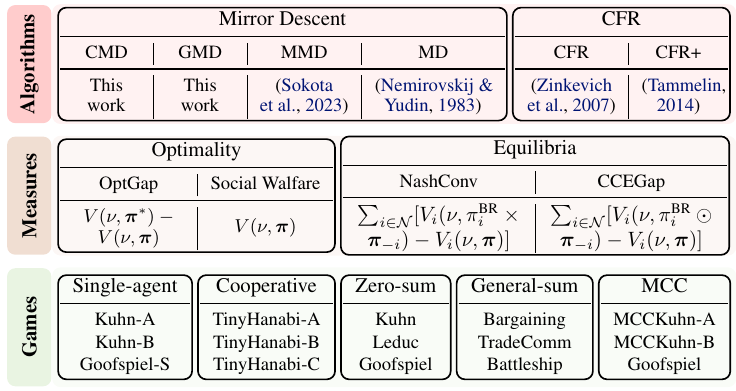}
\vspace{-15pt}
\caption{Overview of \textsc{GameBench}.}
\label{fig:game_bench}
\vspace{-10pt}
\end{figure}

    In this section, we present the \textsc{GameBench}, a novel benchmark which consists of 15 games covering all categories of decision making and includes different evaluation measures and different algorithms, which is shown in Figure~\ref{fig:game_bench}.

\textbf{Motivation.} Although various benchmarks have been suggested in literature, they are typically specialized for specific decision-making categories, e.g., Atari~\cite{bellemare2013arcade} for single-agent category, Hanabi~\cite{bard2020hanabi} for cooperative category, Hold'em poker~\cite{brown2018superhuman,brown2019superhuman} for competitive category, and football~\cite{kurach2020google} for mixed cooperative and competitive category. On the other hand, as MD algorithms require to execute the policy updating at each decision point at each iteration, running them on the existing benchmarks could be resource-intensive as the number of decision points in the environments could be extremely large (e.g., it is impractical to enumerate the observations in Atari as they are images).

\begin{figure*}[ht]
\centering
\includegraphics[width=\textwidth]{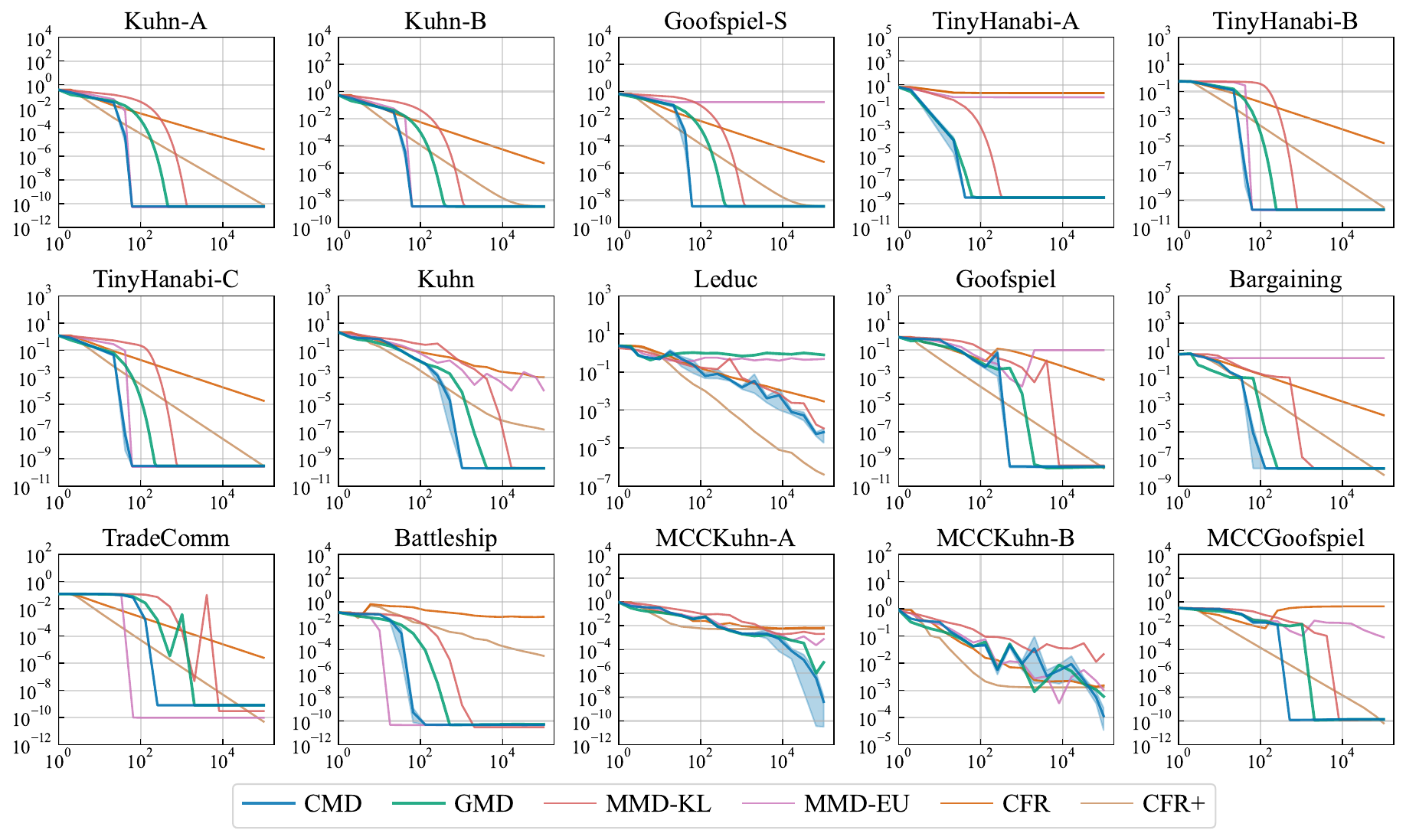}
\vspace{-15pt}
\caption{Summary of results. The first 6 figures correspond to single-agent and cooperative categories where the $y$-axis is \textit{OptGap}. The rest figures correspond to other categories where the $y$-axis is \textit{NashConv}. For all the figures, the $x$-axis is the number of iterations.}
\label{fig:result_summary}
\end{figure*}

\textbf{Desiderata.} Motivated by the above facts, we construct a new benchmark -- \textsc{GameBench}. It satisfies the two desiderata \textbf{D5} and \textbf{D6} presented in the Introduction. That is, it covers all categories of decision-making problems (comprehensive), and running MD algorithms (or other algorithms such as CFR~\cite{zinkevich2007regret}) on all the games does not require much computational resource (academic-friendly). The components of \textsc{GameBench} are given below.

\textbf{Games.} We curate the \textsc{GameBench} on top of the OpenSpiel~\cite{lanctot2019openspiel}. There are 15 games which are divided into 5 categories: single-agent, cooperative multi-agent, competitive multi-agent zero-sum, competitive multi-agent general-sum, and mixed cooperative and competitive (MCC) categories. In our \textsc{GameBench}, the original competitive category is further divided into two subcategories -- zero-sum and general-sum -- as they can involve different solution concepts and evaluation measures (given below). We construct all 15 games under two primary principles: i) these games involve as many aspects of decision making as possible, e.g., the number of agents (single or multiple) and the relationship between agents (cooperative, competitive, or mixed), and ii) these games are relatively simple, have a low barrier to entry, and yet complex enough, and hence, running algorithms on these games is less resource-intensive. The details of the constructions and the statistics of the 15 games can be found in Appendix~\ref{app:gb_games}.

\textbf{Measures.} As \textsc{GameBench} includes different categories of decision-making problems, it is indispensable to consider multiple evaluation measures. Roughly speaking, there are two types of measures: i) the notion of \textit{optimality}, including OptGap and social welfare, and ii) the notion of \textit{equilibrium}, including NashConv and CCEGap. Note that computing the equilibrium-type measures for the MCC category requires a new method to compute the team's best response (BR) (not a single agent's). The details can be found in Appendix~\ref{app:gb_eval_measure}.

\textbf{Algorithms.} We incorporate different MD algorithms into the \textsc{GameBench}, including the state-of-the-art baselines. The comparison between these MD algorithms can be found in Table~\ref{tab:compare_table}. In addition, we also include CFR-type algorithms as the baselines, including the CFR~\cite{zinkevich2007regret} and CFR+~\cite{tammelin2014solving}. Although these algorithms need to update the policy at each decision point, since the numbers of decision points of the games in \textsc{GameBench} are not too large, running these algorithms on these games is relatively easy (academic-friendly).

\section{Experiments}

In this section, we evaluate our method on \textsc{GameBench}. We first describe the experimental setup. Then, we present the results by answering several research questions (\textit{\textbf{RQs}})

\textbf{Setup.} We compare the following methods: i) CMD: our method where the hyper-parameters are determined by the meta-controller introduced in Section~\ref{sec:mc}, ii) GMD: the hyper-parameters are fixed to $\alpha_{\tau} = 1/M$, $0\leq\tau\leq M-1$ (a uniform distribution), iii) MMD-KL: the state-of-the-art method called magnetic mirror descent~\cite{sokota2023unified} where the policy updating rule is induced with KL divergence, iv) MMD-EU: similar to MMD-KL but the policy updating rule is induced with squared Euclidean norm (see Appendix~\ref{app:mmd_eu} for the derivation), v) CFR: the policy is updated based on regret~\cite{zinkevich2007regret}, and vi) CFR+: an advanced version of CFR~\cite{tammelin2014solving}. In CMD and GMD, we also include a magnet policy, which has been argued desirable~\cite{sokota2023unified,liu2023the}. Nevertheless, we note that this does not cause inconsistency with our method as we can equivalently obtain them by setting $M$ and $\alpha_{\tau}$ (see Appendix~\ref{app:gmd_connect_md}). Moreover, without explicitly specifying, the results are obtained under $\psi(x)=x\ln x$, $x\in(0,1]$. In \textit{\textbf{RQ4}}, we study more possible Bregman divergences by setting different $\psi$ in Table~\ref{tab:convex_functions}.

\subsection{Results and Analysis\label{sec:exp_results}}

\textit{\textbf{RQ1. (Desiderata D1 and D2)} Can CMD effectively tackle all categories of decision-making problems?} In Figure~\ref{fig:result_summary}, we show the learning performance of different methods across 15 games. From the results, we can see that, CMD can effectively solve all categories of decision-making problems: in single-agent and cooperative categories, CMD can find the approximate optimal (joint) policy (the OptGap converges to an extremely small value), and in other categories, CMD can find the approximate Nash equilibrium (the NashConv converges to an extremely small value). 

\textit{\textbf{RQ2. (Comparison with Baselines)} How does CMD perform compared with baselines?} As shown in Figure~\ref{fig:result_summary}, by comparison, we can obtain the following takeaways.
\begin{itemize}[left=0.2mm, itemsep=0mm, topsep=0mm]
    \item Incorporating multiple historical policies and dynamically adjusting the hyper-parameters could \textit{accelerate policy learning in terms of the number of iterations}. This can be verified by comparing CMD with MMD-KL where CMD can converge to a similar OptGap or NashConv value with MMD-KL using fewer iterations. This advantage holds even without tuning the hyper-parameters: in most of the games, GMD (the hyper-parameters are fixed) can also converge to a similar OptGap or NashConv value with MMD-KL using fewer iterations. For GMD with different heuristic hyper-parameter adjustment strategies such as linear decay can be found in Appendix~\ref{app:diff_gmd_alpha_sc}.
    \item Introducing the meta-controller is important as it not only accelerates policy learning but also could \textit{achieve competitive or better outcomes}. Specifically, in terms of NashConv, CMD outperforms the baselines in MCCKuhn-A and MCCKuhn-B and performs on par with the baselines in all other games. However, in the most difficult Leduc poker, GMD cannot effectively decrease the NashConv, showcasing the indispensability of the MC.
    \item Regarding the CFR-type algorithms, the results similar to previous works~\cite{sokota2023unified} are also observed: i) The vanilla CFR is typically inferior to CFR+ and CMD in all the games, and ii) in most of the games, CFR+ outperforms CMD over short iteration horizons but is quickly caught by CMD for longer horizons.
\end{itemize}

\textit{\textbf{RQ3. (Different MCs)} Is the direction-guided update in the meta-controller important?} In Figure~\ref{fig:diff-mc-1}, we compare DRS proposed in Section~\ref{sec:mc} with DGLDS, RS, GLDS, and GLD (see Appendix~\ref{app:meta_controller} for details on these MCs). As we can see, our proposed MC significantly outperforms others either in convergence rate or final performance. In Appendix~\ref{app:diff_mcs}, we visualize the evolution of the hyper-parameter values during the policy learning, verifying the intuition that our DRS can more efficiently adjust the hyper-parameters.

\begin{figure}[ht]
\centering
\includegraphics[width=0.49\columnwidth]{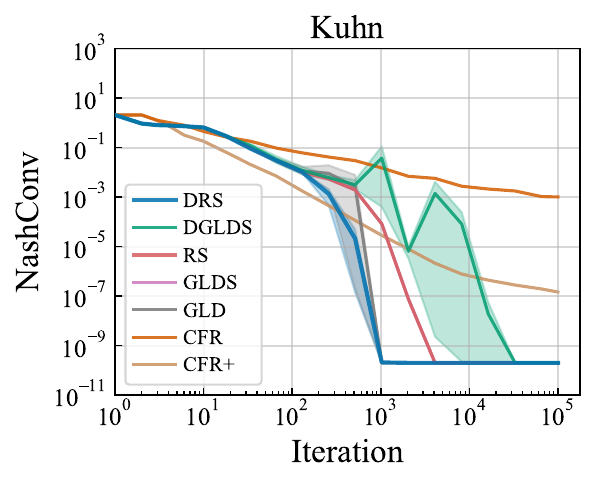}
\includegraphics[width=0.49\columnwidth]{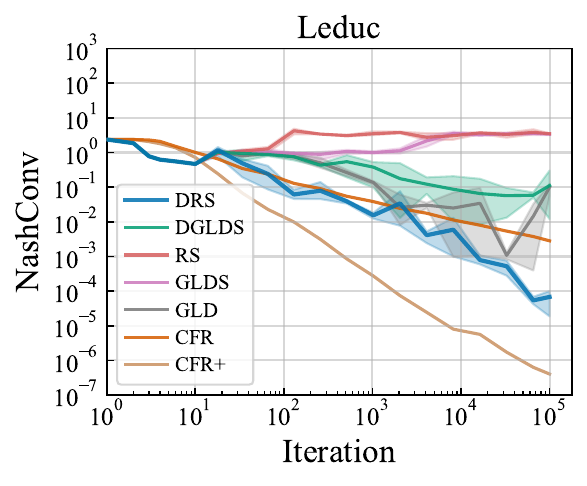}
\vspace{-5pt}
\caption{Results for different types of MC.}
\label{fig:diff-mc-1}
\vspace{-3pt}
\end{figure}

\begin{figure}[ht]
\centering
\includegraphics[width=0.49\columnwidth]{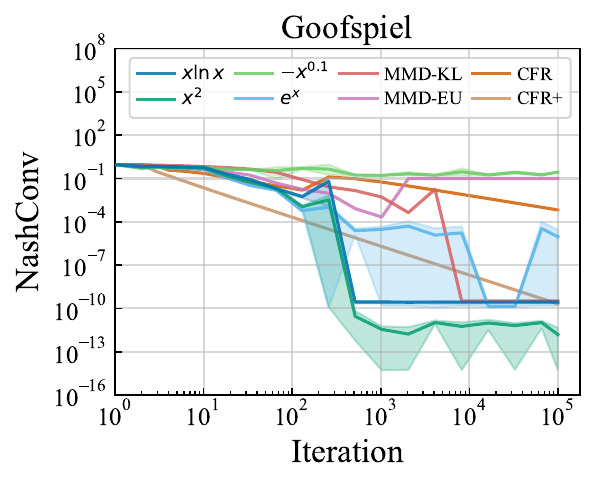}
\includegraphics[width=0.49\columnwidth]{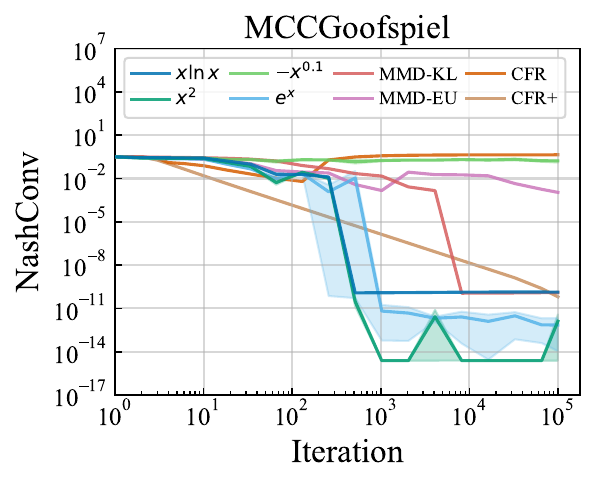}
\vspace{-5pt}
\caption{Results for different convex functions.}
\label{fig:diff-bregman-1}
\end{figure}

\textit{\textbf{RQ4. (Different Bregman Divergences)} How does CMD perform under different Bregman divergences?} In Figure~\ref{fig:diff-bregman-1}, we show the learning curves of different instances of CMD instantiated with different convex functions. From the results, we can see that the KL divergence ($\psi(x)=x\ln x$), though has been widely adopted, could be not the optimal choice across all the decision-making categories. To our knowledge, CMD is the first algorithm that is endowed with the capability of (empirically) exploring a broader class of Bregman divergence, a prominent feature compared with existing MD methods. See Appendix~\ref{app:diff_bregman} for more results.

\begin{figure}[ht]
\centering
\includegraphics[width=0.49\columnwidth]{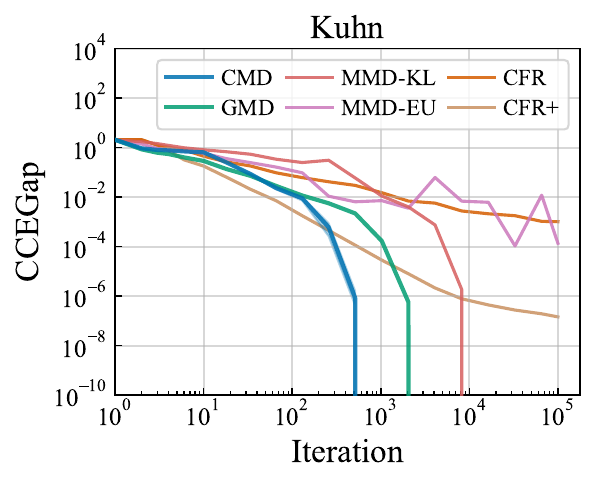}
\includegraphics[width=0.49\columnwidth]{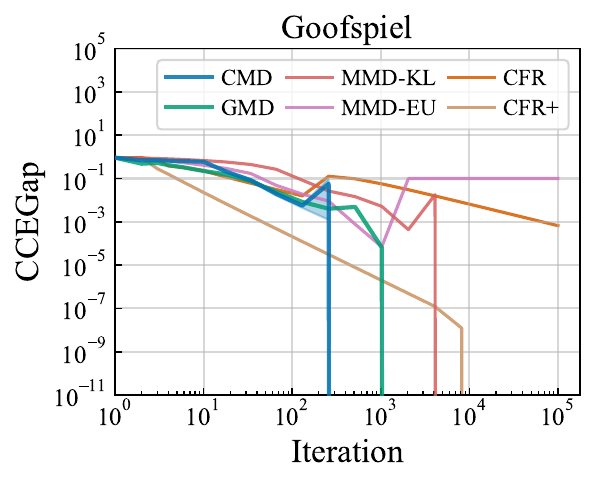}\\
\includegraphics[width=0.49\columnwidth]{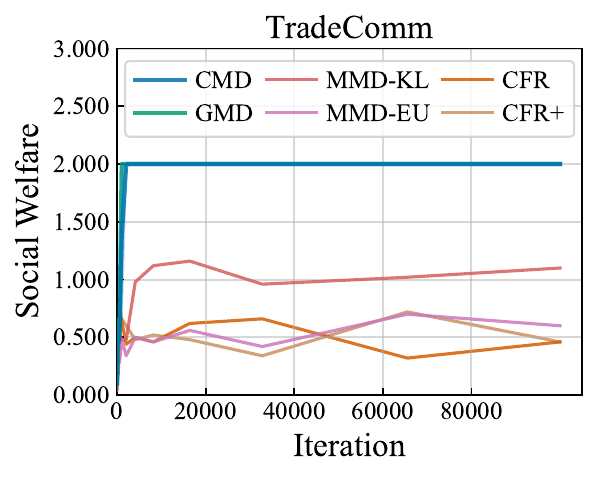}
\includegraphics[width=0.49\columnwidth]{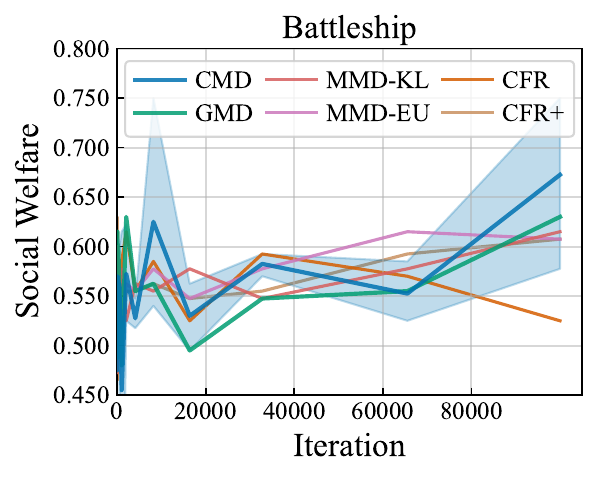}
\vspace{-5pt}
\caption{Results for the measures CCEGap and social welfare.}
\label{fig:cce-sw-result}
\vspace{-3pt}
\end{figure}

\textit{\textbf{RQ5. (Desiderata D3 and D4)} Can CMD generalize to consider different solution concepts and evaluation measures?} In Figure~\ref{fig:cce-sw-result}, we apply CMD to two different measures: CCEGap (top line) and social welfare (bottom line). The results verify the effectiveness of incorporating multiple historical policies and dynamically adjusting the hyper-parameter values conditional on the evaluation measures when considering other evaluation measures beyond NashConv. More results and analysis can be found in Appendix~\ref{app:diff_measure}. Notably, our CMD can be easily applied to different measures with minimal modifications: changing the MC's objective $\mathcal{L}$.

\textit{\textbf{RQ6. (Desiderata D5 and D6)} Is running the different algorithms computationally difficult?} We found that, although extra operations may be required, running the MD and CFR algorithms on \textsc{GameBench} does not cause much burden on the computational resource. The analysis of the computational complexity can be found in Appendix~\ref{app:runtime}.

\section{Limitations, Future Works, and Conclusions\label{sec:conclusion}}

In this section, we discuss the limitations of the current version of our approach and present the future directions, followed by conclusions of this work. 

\subsection{Limitations and Future Works}

This work aims to develop a single algorithm to effectively tackle all categories of decision-making problems: single-agent, cooperative multi-agent, competitive multi-agent, and mixed cooperative and competitive multi-agent categories. As a preliminary attempt, there are still some limitations that are worth investigating in future works.

Firstly, as the meta-controller determines the values of the hyper-parameter by sampling multiple candidates, extra computational cost is needed to evaluate the performance of these candidates. In Appendix~\ref{app:runtime}, we present the running time of an iteration of different methods. While requiring extra computational cost, we view this as one of the future directions: developing more computationally efficient hyper-parameter value updating methods without sacrificing performance. For example, in contrast to the current sampling method where the historical samples are entirely ignored after each update, these historical samples could be used to guide the selection of the new hyper-parameter values, e.g., via Bayesian optimization~\cite{lindauer2022smac3} or offline learning approaches~\cite{chen2022towards}. As a result, we may not need to sample multiple candidates, which could further reduce the extra computational cost.

Secondly, we evaluated CMD primarily from the empirical perspective, and the results demonstrate its promise in solving all categories of decision-making problems. Theoretical analysis of the behavior (e.g., the convergence rate) of CMD could be an interesting problem and may require novel tools that may be different from existing works since the policy updating rule of CMD is established with a numerical method, rather than depends on the closed-form solution to the regularized optimization problem in each decision point under some specific Bregman divergence~\cite{sokota2023unified,liu2023the,lee2021lastiterate}.

Thirdly, though our methods are capable of considering a broader class of Bregman divergence, they still require the mathematical formulation of the convex function $\psi$ as computing the value of the dual variable by using the Newton method requires a set of related functions derived from $\psi$ (as shown in Table~\ref{tab:convex_functions}). In other words, the number of Bregman divergences considered in our method is still limited. An interesting future direction is to develop a novel method to more effectively explore the entire space of the convex function, such as using a neural network to represent the convex function $\psi$, which leads to the neural Bregman divergence~\cite{lu2022neural,siahkamari2020learning,cilingir2020deep,amos2017input}. Moreover, using neural Bregman divergence could also be a possible solution for automatically choosing the Bregman divergences for different decision-making categories. Nevertheless, it could be non-trivial to integrate the neural Bregman divergence as training the neural network to well approximate the optimal convex function for the given category may not be easy and thus may require new treatment.

Finally, in its current version, \textsc{GameBench} consists of 15 academic-friendly games covering all categories of decision-making problems, different evaluation measures, and several baseline algorithms. We believe further extensions could be valuable. i) We could include more games with varying complexity (e.g., different numbers of decision points)~\cite{lanctot2023populationbased}. ii) We could include more evaluation measures such as fairness~\cite{rabin1993incorporating}. iii) We could include more algorithms. In particular, we may include deep learning-based algorithms~\cite{schulman2017proximal,yu2022surprising,lanctot2017unified} and investigate whether there exists a single deep learning-based algorithm that can effectively solve all categories of decision-making problems. iv) Recently, much attention has been drawn to studying the ability of large language models (LLMs) to solve various decision-making problems~\cite{hong2023metagpt,meta2022human,mao2023alympics}. Therefore, an interesting extension is to include LLMs as the baselines, and if necessary, develop new LLMs to more effectively solve different categories of decision-making problems.

\subsection{Conclusions}

In this work, we make the preliminary attempt to develop \textit{\textbf{a single algorithm} to tackle \textbf{ALL} categories of decision-making problems} and provide three contributions: i) the GMD, a generalization of exiting MD algorithms, which can be applied to different decision-making categories involving different numbers of agents and different relationships between agents (\textbf{D1} and \textbf{D2}), ii) the CMD which can be configured to apply to different solution concepts and evaluation measures (\textbf{D3} and \textbf{D4}), and iii) the comprehensive (\textbf{D5}) and academic-friendly (\textbf{D6}) benchmark -- \textsc{GameBench}. Extensive experiments demonstrate the effectiveness of CMD.

\section*{Acknowledgements}

This work is supported by the National Research Foundation, Singapore under its Industry Alignment Fund -- Pre-positioning (IAF-PP) Funding Initiative. Any opinions, findings and conclusions, or recommendations expressed in this material are those of the author(s) and do not reflect the views of National Research Foundation, Singapore. Hau Chan is supported by the National Institute of General Medical Sciences of the National Institutes of Health [P20GM130461], the Rural Drug Addiction Research Center at the University of Nebraska-Lincoln, and the National Science Foundation under grant IIS:RI \#2302999. The content is solely the responsibility of the authors and does not necessarily represent the official views of the funding agencies. This work is also supported by Shanghai Artificial Intelligence Laboratory.

\section*{Impact Statement}

This paper presents work whose goal is to advance the field of decision making. There are many potential societal consequences of our work, none of which we feel must be specifically highlighted here.


\bibliography{icml2024}
\bibliographystyle{icml2024}

\newpage
\appendix
\onecolumn

\section{Code Repository}

Code for experiments is available at \url{https://github.com/IpadLi/CMD}.

\section{More Related Works\label{app:more_related_works}}

\textbf{Single-Agent Category.} In the single-agent category, reinforcement learning (RL)~\cite{sutton2018reinforcement} has proved successful in many real-world applications. The power of RL is further amplified with the integration of deep neural networks, leading to various deep RL algorithms that have been successfully applied to various application domains such as video games~\cite{mnih2015human}, robot navigation~\cite{singh2022reinforcement}, and financial technology~\cite{sun2023reinforcement}. Among these algorithms, PPO~\cite{schulman2017proximal} is one of the most commonly used methods to solve single-agent RL problems. Recent works have shown that independent PPO~\cite{de2020independent,sun2023trust} can effectively solve single-agent and cooperative multi-agent RL problems. In addition, a variant of PPO is also shown to be effective in solving two-player zero-sum games when both players adopt this algorithm~\cite{sokota2023unified}. Nevertheless, it remains elusive whether these single-agent algorithms can be applied to solve other categories of decision-making problems which may involve different properties including different numbers of agents, different relationships between agents, different solution concepts, and different evaluation measures. In this work, we aim to develop a single algorithm that, when executed by each agent, provides an effective approach to address different categories of decision-making problems.

\textbf{Cooperative Multi-Agent Category.} Cooperative multi-agent RL (MARL) has been demonstrated successful in solving many real-world cooperative tasks such as traffic signal control~\cite{xu2021hierarchically,su2022emvlight}, power management~\cite{wang2021multi}, finance~\cite{fang2023learning}, and multi-robot cooperation~\cite{rizk2019cooperative}. In the past decade, a variety of MARL algorithms, e.g., QMIX~\cite{rashid2018qmix} and its variants~\cite{son2019qtran,rashid2020weighted,wang2021qplex}, MADDPG~\cite{lowe2017multi}, COMA~\cite{foerster2018counterfactual}, and MAPPO~\cite{yu2022surprising}, to name just a few, have been proposed and achieved significant performance in various multi-agent benchmarks, e.g., SMAC~\cite{samvelyan2019starcraft} and Dota II~\cite{berner2019dota}. These algorithms typically follow the principle of centralized training and decentralized execution (CTDE) where global information is only available during training. Despite their success, they cannot be directly applied to competitive and mixed cooperative and competitive categories. In this work, our proposed CMD can be applied to different decision-making categories and share similarities with the CTDE paradigm: the meta-controller takes all the agents (i.e., the joint policy) into account to optimize the hyper-parameters conditional on the targeted evaluation measure (a \enquote{centralized} process) while each agent in the environment independently execute the GMD with the given hyper-parameters to update the policy (a \enquote{decentralized} process). 

\textbf{Competitive Multi-Agent Category.} There has long been a history of researchers pursuing artificial intelligence (AI) agents that can achieve human-level or super-human-level performance in solving various competitive multi-agent games such as chess~\cite{campbell2002deep}, Go~\cite{silver2017mastering}, poker~\cite{brown2019superhuman}, and Stratego~\cite{perolat2022mastering}. Due to the competitive nature, the development of learning algorithms for solving these games is typically largely different from single-agent and cooperative MARL. Among others, counterfactual regret minimization (CFR)~\cite{zinkevich2007regret} and policy-space response oracles (PSRO)~\cite{lanctot2017unified} are two representative algorithms that have been widely used to solve complex games~\cite{schmid2023student}. Another category of algorithm that has drawn increasing attention recently is the mirror descent (MD)~\cite{nemirovskij1983problem,beck2003mirror}. In contrast to CFR and PSRO which are \enquote{average-iterate} algorithms, MD has proved the \enquote{last-iterate} convergence property in solving two-player zero-sum games~\cite{bailey2018multiplicative,kangarshahi2018let,wibisono2022alternating,kozuno2021learning,lee2021lastiterate,jain2022matrix,ao2023asynchronous,liu2023the,cen2023faster,sokota2023unified} and some classes of general-sum games~\cite{anagnostides2022last}. Moreover, MD has also been demonstrated effective in solving single-agent RL problems~\cite{tomar2022mirror}. Despite their success, existing MD algorithms typically focus on some specific Bregman divergences which may not be the optimal choices across different decision-making categories. Our proposed CMD generalizes existing MD algorithms to consider a broader class of Bregman divergence, which could achieve better learning performance in addressing different categories of decision-making problems.

\textbf{Mixed Cooperative and Competitive Category.} In some real-world scenarios, the relationship between agents could be neither purely cooperative nor purely competitive. For example, in a football game, the agents belonging to the same team need to cooperate while also competing with the other team~\cite{kurach2020google}. In hidden-role games~\cite{carminati2023hidden}, each agent tries to identify their (unknown) teammates and compete with other (unknown) adversaries~\cite{wang2018application,serrino2019finding,albrecht2022avalon}. However, in contrast to the other three categories (single-agent, purely cooperative, and purely competitive), mixed cooperative and competitive (MCC) games are largely unstudied~\cite{xu2023fictitious}. Furthermore, as MD algorithms typically require updating the policy at each decision point, running them on the current benchmark games such as football~\cite{kurach2020google} could be computationally prohibited. In the present work, we construct 3 MCC games that are academic-friendly -- their numbers of decision points are not too large and hence, running MD algorithms (and other algorithms such as CFR-type~\cite{zinkevich2007regret,tammelin2014solving}) on these games does not require much computational resource (e.g., running time and memory usage).

\textbf{Hyper-Parameter Tuning.} Existing works typically determine the hyper-parameter values of the MD algorithms depending on the domain knowledge~\cite{sokota2023unified,anagnostides2022last,hsieh2021adaptive,zhou2018learning,mertikopoulos2019optimistic,bailey2019fast,golowich2020tight}, which, though convenient for theoretical analysis, may not be easy to generalize to different games. On the other hand, as the evaluation measures, e.g., NashConv, could be non-differentiable with respect to the hyper-parameters, the gradient-based methods such as STAC~\cite{zahavy2020self} could also be less applicable. In this sense, a more feasible method is the zero-order hyper-parameter optimization which can update the parameters of interest without access to the true gradient, which has been extensively adopted in the adversarial robustness of deep neural networks~\cite{ilyas2018black}, meta-learning~\cite{song2020esmaml}, transfer learning~\cite{tsai2020transfer}, and neural architecture search (NSA)~\cite{wang2022zarts}. Nevertheless, we found that directly applying existing zero-order methods could be ineffective as when the value of the evaluation measure is too small, they may not be able to derive an effective update for the hyper-parameter. To address this issue, we propose a simple yet effective technique -- \textit{direction-guided update} -- where the performance difference between two candidates is used to only determine the \textit{update direction} of the hyper-parameters rather than the update magnitude, which is more effective than existing methods~\cite{wang2022zarts} when the value of the performance is extremely small.

\clearpage
\section{Notation Table}
\begin{table}[H]
    \centering
    \setlength\tabcolsep{8pt}
    \caption{Notation Table.}
    \label{tab:notation}
    \vskip 0.1in
    \begin{tabular}{cl}
    \toprule
    $\mathcal{N}$ & $\mathcal{N}=\{1, \cdots, N\}$, the set of $N$ agents. \\
    $\mathcal{S}$ & the finite set of states.\\
    $\mathcal{A}$ & $\mathcal{A}=\times_{i\in\mathcal{N}}\mathcal{A}_i$ where $\mathcal{A}_i$ is the finite set of actions of agent $i$.\\
    $\mathcal{O}$ & $\mathcal{O}=\times_{i\in\mathcal{N}}\mathcal{O}_{i}$ where $\mathcal{O}_{i}$ is the finite set of observations of agent $i$.\\
    $\Omega$ & $\Omega=\times_{i\in\mathcal{N}}\Omega_{i}$ where $\Omega_{i}:\mathcal{S}\times\mathcal{A}\rightarrow\mathcal{O}_{i}$ is the observation function of agent $i$.\\
    $P$ & $P:\mathcal{S}\times\mathcal{A}\rightarrow\Delta(\mathcal{S})$, the state transition function.\\
    $R$ & $R=\{r_{i}\}_{i\in\mathcal{N}}$ where $r_{i}:\mathcal{S}\times\mathcal{A}\rightarrow\mathbb{R}$ is the reward function of agent $i$.\\
    $\gamma$ & $\gamma\in[0, 1)$, the discount factor.\\
    $\nu$ & $\nu\in\Delta(\mathcal{S})$, the initial state distribution.\\
    $\tau_i^t$ & the decision point (action-observation history) of agent $i$ at time $t$, $\tau_{i}^{t}\in\mathcal{T}_{i}^{t}$. \\
    $\mathcal{T}_{i}^{t}$ & $\mathcal{T}_{i}^{t}=(\mathcal{O}_{i}\times\mathcal{A}_{i})^{t}$, the space of decision points of agent $i$ at time step $t$.\\
    $\Pi$ & $\Pi = \times_{i\in\mathcal{N}}\Pi_i$ where $\Pi_i$ is the policy space of agent $i$.\\
    $\bm{\pi}$ & $\bm{\pi}=\pi_1\odot\cdots\odot\pi_N$, the joint policy, $\bm{\pi}=\pi_1\times\cdots\times\pi_N$, the product policy.\\
    $V_i(s, \bm{\pi})$, $V_i(\nu, \bm{\pi})$ & the value functions of agent $i$, $V_i(\nu, \bm{\pi}):=\mathbb{E}_{s\sim \nu}[V_i(s,\bm{\pi})]$.\\
    $\mathcal{L}(\bm{\pi})$ & the evaluation measure of the joint policy $\bm{\pi}$.\\
    \midrule
    $\pi_k$ & the single agent's policy at the iteration $k$ of an algorithm.\\
    $\bm{\pi}_k$ & the joint policy at the iteration $k$ of an algorithm.\\
    $Q(\pi_{k})$ & $Q(\pi_{k})=(Q(a, \pi_{k}))_{a\in\mathcal{A}}$, the action-value vector of a single agent induced by $\pi_k$.\\
    $\mathcal{B}_{\phi}(x;y)$ & $\mathcal{B}_{\phi}(x;y)=\phi(x)-\phi(y)-\langle\nabla\phi(y), x-y\rangle$, the Bregman divergence with respect to $\phi$.\\
    $K$ & the number of iterations of an algorithm.\\
    \midrule
    $M$ & $M\ge1$, the number of historical policies. \\
    $\bm{\alpha}$ & $\bm{\alpha}=(\alpha_{\tau})_{0\leq\tau\leq M-1}$, $\alpha_{\tau}$ is the regularization intensity of $\pi_{k-\tau}$.\\
    $\epsilon$ & $\epsilon>0$, the smallest probability of an action. \\
    $\phi(\pi)$ & $\phi(\pi)=\sum_{a\in\mathcal{A}}\psi(\pi(a))$, $\psi$ is some convex function defined on $[0, 1]$.\\
    $\lambda$, $\bm{\beta}$ & $\bm{\beta}=(\beta_a)_{a\in\mathcal{A}}$, the dual variables. \\
    $A$, $B$ & $A=Q(\pi_{k}) +\sum_{\tau=0}^{M-1}\alpha_{\tau}\phi^{\prime}(\pi_{k-\tau})$, $B=\sum_{\tau=0}^{M-1} \alpha_{\tau}$, where $\phi^{\prime}$ is the derivative of $\phi$. \\
    $\psi^{\prime-1}$ & the inverse function of $\psi^{\prime}$ (the derivative of $\psi$).\\
    $C$ & $C>0$, the number of iterations for the Newton method.\\
    $g(\lambda)$, $g^{\prime}(\lambda)$ & $g(\lambda)=\left[\sum\nolimits_{a\in\mathcal{A}}\psi^{\prime-1}(A(a)-\lambda)/B)\right] - 1$, $g^{\prime}(\lambda)=\sum\nolimits_{a\in\mathcal{A}}-\frac{1}{B}[\psi^{\prime-1}]^{\prime}(A(a)-\lambda)/B)$.\\
    $[\psi^{\prime-1}]^{\prime}$ & the derivative of $\psi^{\prime-1}$.\\
    \midrule
    $D$ & the number of sampled candidate $\bm{\alpha}$'s.\\
    $\{\bm{\alpha}^j\}_{j=1}^{D}$ & $D$ candidates by perturbing the current $\bm{\alpha}$.\\
    $\{\bm{\pi}^j\}_{j=1}^{D}$ & $\{\bm{\pi}^j=\text{GMD}(\bm{\alpha}^j)\}_{j=1}^{D}$, $D$ new joint policies derived via GMD.\\
    $\mu$ & the smoothing parameter in DRS and RS.\\
    $[r_L, r_H]$ & the interval of the radiuses of the spheres in DGLDS, GLDS, and GLD.\\
    $\{\bm{u}^j\}_{j=1}^{D}$ & $D$ candidate updates sampled from a spherically symmetric distribution $\bm{u}^j \sim q$.\\
    $\bm{\alpha}_{+}^j$, $\bm{\alpha}_{-}^j$ & $\bm{\alpha}_{+}^j = \text{CLIP}_{\iota}^{1}(\bm{\alpha}+\mu\bm{u}^j)$, $\bm{\alpha}^j_{-} = \text{CLIP}_{\iota}^{1}(\bm{\alpha}-\mu\bm{u}^j)$, the candidates by perturbing the current $\bm{\alpha}$. \\
    $\bm{\pi}^j_{+}$, $\bm{\pi}^j_{-}$ & $\bm{\pi}^j_{+} = \text{GMD}(\bm{\alpha}^j_{+})$, $\bm{\pi}^j_{-} = \text{GMD}(\bm{\alpha}^j_{-})$, the new joint policies obtained via GMD.\\
    $\delta^j$ & $\delta^j = \mathcal{L}(\bm{\pi}^j_{+})-\mathcal{L}(\bm{\pi}^j_{-})$, the performance difference between $\bm{\pi}^j_{+}$ and $\bm{\pi}^j_{-}$.\\
    $\bm{u}^{*}$ & the final update.\\
    $\text{Sgn}$ & $\text{Sgn}(z)=1$ if $z>0$, $\text{Sgn}(z)=-1$ if $z<0$, otherwise, $\text{Sgn}(z)=0$.\\
    $\text{CLIP}_{\iota}^{1}$ & $\text{CLIP}_{\iota}^{1}(z)=\iota$ if $z<\iota$, $\text{CLIP}_{\iota}^{1}(z)=1$ if $z>1$, otherwise, $\text{CLIP}_{\iota}^{1}(z)=z$, where $0<\iota<1$. \\
    $\kappa$ & $\kappa\ge 1$, update the $\bm{\alpha}$ every $\kappa$ iterations.\\
    \midrule
    $\rho$ & the magnet policy in MMD.\\
    $\xi$ & $\xi>0$, the regularization intensity of the magnet policy.\\
    $\eta$ & $\eta>0$, the step size in MMD.\\
    $\tilde{\eta}$ & $\tilde{\eta}>0$, the step size of the magnet policy in MMD.\\
    \bottomrule
    \end{tabular}
\end{table}

\clearpage
\section{Configurable Mirror Descent\label{app:cmd}}

In this section, we present all the details of our methods. In Section~\ref{app:gmd}, we present the details of GMD. In Section~\ref{app:gmd_connect_md}, we establish some connections between GMD and existing MD algorithms. In Section~\ref{app:mmd_eu}, we restrict the convex function $\phi$ to the squared Euclidean norm and derive the closed-form solution under the MMD policy updating rule. Finally, in Section~\ref{app:meta_controller}, we present the details of different meta-controllers. 

\subsection{Generalized Mirror Descent\label{app:gmd}}

In this section, we present the proof of Proposition~\ref{prop:kkt_problem}, the pseudo-code of Newton's method for computing the value of the dual variable, and the convex functions considered in our work.
\begin{proof}[\textbf{Proof of Proposition~\ref{prop:kkt_problem}}]
Consider the optimization problem (\ref{eq:gmd}). By the definition of Bregman divergence, we have:
\begin{align}
\pi_{k+1} &= \arg\max\nolimits_{\pi\in\Pi} \langle Q(\pi_{k}), \pi \rangle - \sum\nolimits_{\tau = 0}^{M-1} \alpha_{\tau}\mathcal{B}_{\phi} (\pi, \pi_{k-\tau}), \\
\Rightarrow\pi_{k+1} &= \arg\max\nolimits_{\pi\in\Pi} \langle Q(\pi_{k}), \pi \rangle - \sum\nolimits_{\tau = 0}^{M-1} \alpha_{\tau}[\phi(\pi) - \phi(\pi_{k-\tau}) - \langle\phi'(\pi_{k-\tau}), \pi-\pi_{k-\tau}\rangle], \\
\Rightarrow\pi_{k+1} &= \arg\max\nolimits_{\pi\in\Pi} \langle Q(\pi_{k}) +\sum\nolimits_{\tau=0}^{M-1}\alpha_{\tau}\phi'(\pi_{k-\tau}), \pi\rangle - \phi(\pi)\sum\nolimits_{\tau=0}^{M-1} \alpha_{\tau} + \text{const.},
\end{align}
where \enquote{const.} summarizes all terms that are irrelevant to $\pi$. Let $A=Q(\pi_{k}) +\sum_{\tau=0}^{M-1}\alpha_{\tau}\phi^{\prime}(\pi_{k-\tau})$ and $B=\sum_{\tau=0}^{M-1} \alpha_{\tau}$ which are fixed at the current iteration $k$. Then, we can convert Eq. (\ref{eq:gmd}) to the following optimization problem:
\begin{equation}
\begin{aligned}
    & \pi_{k+1} = \operatorname*{arg\,max}\nolimits_{\pi \in \Pi} \langle A, \pi\rangle  - B\phi(\pi) + \text{const.},\\
    & \textit{s.t. } \sum\nolimits_{a\in\mathcal{A}}\pi_k(a) = 1 \text{ and } \pi_k(a)\geq 0, \forall a \in\mathcal{A}.
\end{aligned}
\label{eq:converted-gmd}
\end{equation}
To solve the constrained optimization problem (\ref{eq:converted-gmd}), we can apply the Lagrange multiplier, which gives us:
\begin{align}
&L(\pi, \lambda, \bm{\beta})=\langle A, \pi\rangle - B\phi(\pi) + \text{const.} - \lambda \left(\sum\nolimits_{a\in\mathcal{A}}\pi(a) - 1\right) +  \sum\nolimits_{a\in\mathcal{A}}\beta_{a}\pi(a),
\end{align}
where $\lambda$ and $\bm{\beta}=(\beta_{a})_{a\in\mathcal{A}}$ are the dual variables. For such problems, we can get the Karush–Kuhn–Tucker (KKT) conditions for each component (action) $a\in\mathcal{A}$ as follows:
\begin{subequations}
\begin{align}
    A(a) + B\phi^{\prime}(\pi)(a) - \lambda + \beta_{a} = 0,\label{eq:kkt-main} \\ 
    \sum\nolimits_{a\in\mathcal{A}}\pi(a) = 1,\label{eq:kkt-prob-sum}\\
    \beta_{a} \pi(a) = 0,\\
    \pi(a)\geq 0, \beta_{a}\geq 0.\label{eq:kkt-last}
\end{align}
\end{subequations}
Then the problem is to find a policy $\pi$ such that it satisfies all the above conditions, which could be difficult owing to two reasons: i) it simultaneously involves the two dual variables $\lambda$ and $\beta_a$, and ii) in Eq. (\ref{eq:kkt-main}), computing the value $\phi^{\prime}(\pi)(a)$ involves all the components (actions) as $\phi$ is defined on the policy $\pi$, not the individual component (action) $a\in\mathcal{A}$. 

To address the challenges, we apply the two conditions: $\pi(a) \geq \epsilon$ and $\phi(\pi)=\sum_{a\in\mathcal{A}}\psi(\pi(a))$. Then, we have $\phi^{\prime}(\pi)(a)=\psi^{\prime}(\pi(a))$. As a result, the problem (\ref{eq:kkt-main}--\ref{eq:kkt-last}) is simplified to the following problem:
\begin{subequations}
\begin{align}
    A(a) + B\psi^{\prime}(\pi(a)) - \lambda = 0,\label{eq:reduced-kkt-main}\\
    \sum\nolimits_{a\in\mathcal{A}}\pi(a) = 1,\label{eq:reduced-kkt-sum-prob}\\
    \pi(a)\geq \epsilon.\label{eq:reduced-kkt-last}
\end{align}
\end{subequations}
From Eq. (\ref{eq:reduced-kkt-main}), we can get that: $\forall a\in\mathcal{A}$,
\begin{align}
    \pi(a) = \psi^{\prime-1}\left(\frac{A(a)-\lambda}{B}\right),\label{eq:pi_a}
\end{align}
where $\psi^{\prime-1}$ is the inverse function of $\psi^{\prime}$. Substituting the above expression for $\pi(a)$ into the constraint (\ref{eq:kkt-prob-sum}), we have:
\begin{align}
    \sum\nolimits_{a\in\mathcal{A}}\pi(a) = \sum\nolimits_{a\in\mathcal{A}}\psi^{\prime-1}\left(\frac{A(a)-\lambda}{B}\right) = 1,
    \label{eq:compute_lambda}
\end{align}
which completes the proof.
\end{proof}

\textbf{Numerical Method for Computing $\lambda$.} Notably, Eq. (\ref{eq:compute_lambda}) typically cannot be solved \textit{analytically}. To address this problem, we propose to use a numerical method to compute the value of $\lambda$, offering the possibility of exploring a broader class of Bregman divergence. Specifically, for any convex function $\psi$, we employ the Newton method~\cite{ypma1995historical} to compute the value of $\lambda$, which is shown in Algorithm~\ref{alg:newton}, where $C$ is the number of iterations.
\begin{algorithm}
\caption{Newton method for computing the value of $\lambda$}
\label{alg:newton}
\begin{algorithmic}[1]
\STATE Given $\psi$, $A$, and $B$. Randomly initialize the value of $\lambda$
\FOR{$C$ iterations}
\STATE $\lambda = \lambda - \frac{g(\lambda)}{g^{\prime}(\lambda)}$\;
\ENDFOR
\end{algorithmic}
\end{algorithm}

\begin{figure}[ht]
\centering
\vspace{-5pt}
\subfigure[$x\ln x$]{\includegraphics[width=0.24\textwidth]{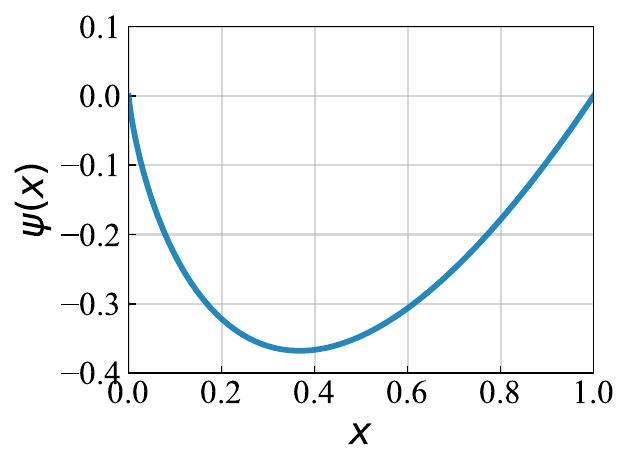}}
\subfigure[$x^2$]{\includegraphics[width=0.24\textwidth]{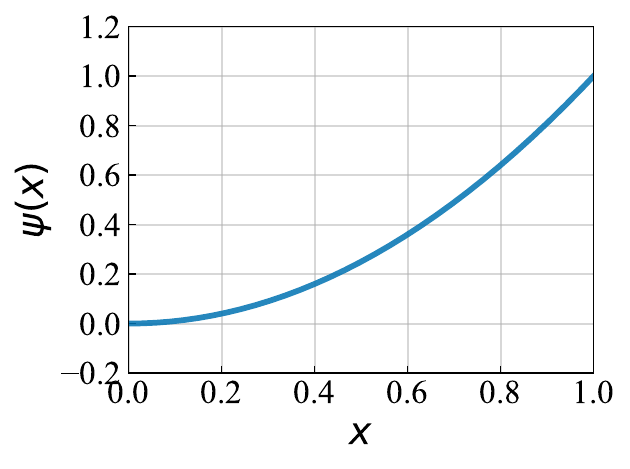}}
\subfigure[$-x^{0.1}$]{\includegraphics[width=0.24\textwidth]{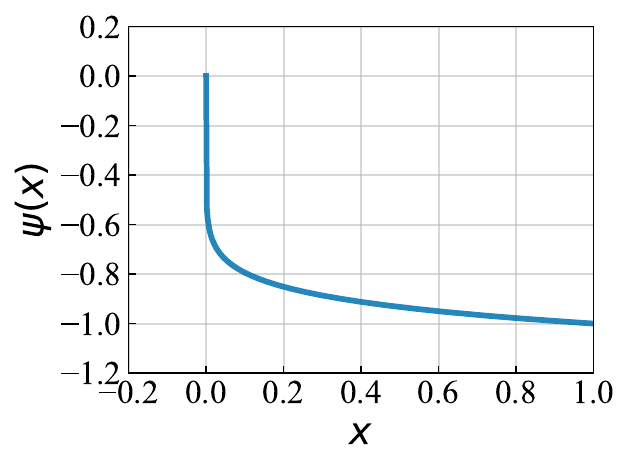}}
\subfigure[$e^x$]{\includegraphics[width=0.24\textwidth]{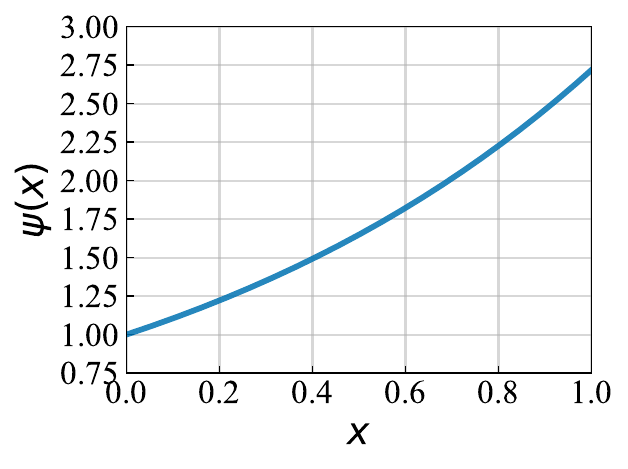}}
\vspace{-5pt}
\caption{Plots for different convex functions $\psi$.}
\label{fig:diff-convex-function}
\vspace{-5pt}
\end{figure}

\textbf{Different Bregman Divergences.} In Table~\ref{tab:convex_functions}, we list the convex functions considered in our work. To be more intuitive, we plot these convex functions in Figure~\ref{fig:diff-convex-function}.

\subsection{Connection Between GMD and Existing MD Algorithms~\label{app:gmd_connect_md}}

In this section, we present some discussion on the connection between GMD and existing MD algorithms. In Table~\ref{tab:gmd_connect_md}, we present the conditions for converting GMD to different MD algorithms and their formulations.
\begin{table}[ht]
\centering
\caption{Connection between GMD and existing MD algorithms.}
\label{tab:gmd_connect_md}
\vskip 0.1in
\begin{tabular}{l|l|l}
\toprule
 Method & Conditions & Formulation \\
\midrule
MD & $M=1$, $\alpha_0\in(0, 1]$ & $\pi_{k+1} = \arg\max_{\pi\in\Pi} \langle Q(\pi_{k}), \pi\rangle  - \alpha_0 \mathcal{B}_{\phi}(\pi, \pi_{k})$ \\
\midrule
MMD & \makecell[l]{$M=k$, $\alpha_{k-1}, \alpha_0\in(0, 1]$,\\$\alpha_{\tau}=0$, $0<\tau<k-1$} & $\pi_{k+1} = \arg\max_{\pi\in\Pi} \langle Q(\pi_{k}), \pi\rangle - \alpha_{k-1} \mathcal{B}_{\phi}(\pi, \pi_1) - \alpha_0 \mathcal{B}_{\phi}(\pi, \pi_{k})$ \\
\bottomrule
\end{tabular}
\end{table}

\textbf{GMD $\to$ MD.} It is trivial to get the MD algorithm~\cite{nemirovskij1983problem,beck2003mirror} by setting $M=1$ and $\alpha_0>0$, that is, MD only considers the current policy $\pi_k$ when deriving the new policy $\pi_{k+1}$.

\textbf{GMD $\to$ MMD.} To obtain MMD~\cite{sokota2023unified}, we can set $M=k$ and then only let $\alpha_{k-1}$ and $\alpha_0$ to be positive, while all other terms are 0. That is, MMD considers two previous policies -- the initial policy $\pi_1$ and the current policy $\pi_k$ -- when deriving the new policy. In the MMD's terminology, the initial policy $\pi_1$ serves as the magnet policy.

In practice, we can get more variants by setting the $M$ and $\bm{\alpha}$, which shows that GMD is a general method. For example, we can consider both $M<k$ previous policies and the initial policy $\pi_1$ (i.e., adding a magnet policy) when deriving the new policy $\pi_{k+1}$, which is taken as the default choice for instantiating the GMD in our experiments, that is, 
\begin{equation}
    \pi_{k+1} = \arg\max_{\pi\in\Pi} \langle Q(\pi_{k}), \pi\rangle - \alpha_{k-1} \mathcal{B}_{\phi}(\pi, \pi_1) - \sum_{\tau = 0}^{M-1} \alpha_{\tau}\mathcal{B}_{\phi} (\pi, \pi_{k-\tau}).\label{eq:gmd_instance}
\end{equation}
In Appendix~\ref{app:effect_of_magnet}, we perform an ablation study to show the effectiveness of adding such a magnet policy. Nevertheless, it is worth noting that this particular choice should not be confused with the original MMD even when $M=1$ as the policy updating rule is derived via a numerical method, instead of relying on the closed-form solution~\cite{sokota2023unified}.

\subsection{Derivation of MMD-EU\label{app:mmd_eu}}

In this section, we present the details of the baseline, MMD-EU, used in our experiments. This baseline follows the spirit of MMD-KL~\cite{sokota2023unified}. Consider the following problem:
\begin{equation}
    \pi_{k+1} = \arg\max_{\pi\in\Pi} \langle Q(\pi_{k}), \pi\rangle - \xi \mathcal{B}_{\phi}(\pi, \rho) - \frac{1}{\eta} \mathcal{B}_{\phi}(\pi, \pi_{k}),
\end{equation}
where $\xi>0$ is the regularization intensity, $\eta>0$ is the step size, and $\rho$ is the magnet policy. Let $\phi(\pi)=\sum_{a\in\mathcal{A}}\frac{1}{2}\Vert\pi(a)\Vert^2$, i.e., the squared Euclidean norm. Then, we need to optimize the following objective:
\begin{align}
 \langle Q(\pi_k), \pi\rangle  - \frac{\xi}{2}||\pi - \rho||^{2}_{2} - \frac{1}{2\eta}||\pi - \pi_{k}||^{2}_{2},
\end{align}
with the constraint $\sum_{a\in\mathcal{A}}\pi(a)=1$ and $\pi(a)>0$. We can use the Lagrange multiplier to get the following objective:
\begin{align}
 \langle Q(\pi_k), \pi\rangle  - \frac{\xi}{2}||\pi - \rho||^{2}_{2} - \frac{1}{2\eta}||\pi - \pi_{k}||^{2}_{2} + \lambda \left(1-\sum_{a\in\mathcal{A}}\pi(a)\right).
\end{align}
Taking the derivative of both $\pi$ and $\lambda$, we have:
\begin{align}
    &Q(a, \pi_k) - \xi (\pi(a) - \rho(a)) - \frac{1}{\eta} (\pi(a)-\pi_{k}(a)) - \lambda = 0, \forall a\in\mathcal{A}, \label{eq:pi_i}\\ 
    &\sum_{a\in\mathcal{A}}\pi(a)=1. \label{eq:sum_one}
\end{align}
Therefore from Eq.~(\ref{eq:pi_i}), we have:
\begin{align}
    \pi(a) = \frac{\xi\rho(a) + \frac{1}{\eta}\pi_{k}(a) + Q(a,\pi_k) - \lambda}{(\xi + \frac{1}{\eta})}.
\end{align}
Substituting the above equation to Eq.~(\ref{eq:sum_one}), we have:
\begin{align}
    &\sum_{a\in\mathcal{A}}\frac{\xi\rho(a) + \frac{1}{\eta}\pi_{k}(a) + Q(a,\pi_k) - \lambda}{(\xi + \frac{1}{\eta})} = 1, \\ 
    \Rightarrow&\sum_{a\in\mathcal{A}}\left[\xi\rho(a) + \frac{1}{\eta}\pi_{k}(a) + Q(a,\pi_k)\right] = (\xi + \frac{1}{\eta}) + \sum_{a\in\mathcal{A}}\lambda.
\end{align}
Note that $\sum_{a\in\mathcal{A}}\left[\xi\rho(a) + \frac{1}{\eta}\pi_{k}(a)\right]=\xi + \frac{1}{\eta}$, we have:
\begin{align}
    \lambda =\frac{\sum_{a\in\mathcal{A}}Q(a,\pi_k)}{\vert\mathcal{A}\vert}.
\end{align}
Then we can compute the new policy as follows:
\begin{equation}
\begin{aligned}
    \pi(a) &= \frac{\xi\rho(a) + \frac{1}{\eta}\pi_{k}(a) + Q(a,\pi_k) - \frac{1}{\vert\mathcal{A}\vert} \sum_{a'\in\mathcal{A}}Q(a^{\prime},\pi_k)}{(\xi + \frac{1}{\eta})}.
\end{aligned}
\end{equation}
Theoretically, we note that by choosing the suitable values for $\xi$ and $\eta$, we can always ensure that $\pi$ is well-defined, i.e., $\pi(a)\geq 0, \forall a\in\mathcal{A}$. In experiments, we can use a projection operation to ensure this condition ($\zeta=1e-10$ is used to avoid division by zero):
\begin{equation}
    \pi_{k+1}(a) = \frac{\max\{0, \pi(a)\} + \zeta}{\sum_{a^{\prime}\in\mathcal{A}} \max\{0, \pi(a^{\prime})\} + \zeta}.
\end{equation}
In addition, similar to MMD-KL, the magnet policy is updated as (i.e., moving magnet):
\begin{equation}
    \rho_{k+1}(a)=(1-\tilde{\eta})\rho_k(a) + \tilde{\eta}\pi_{k+1}(a),
\end{equation}
where $\tilde{\eta}>0$ is the learning rate for the magnet policy $\rho$. In practice, the initial magnet policy $\rho_1$ can be set to the initial policy $\pi_1$ which is typically a uniform policy.

\subsection{Meta-Controller for Different Measures\label{app:meta_controller}}

GMD generalizes exiting MD algorithms in two aspects: i) it takes multiple previous policies into account and can recover some of the existing MD algorithms by setting the $M$ and $\bm{\alpha}$, and ii) it can consider a broader class of Bregman divergence by setting $\phi$ to more possible convex functions (Table~\ref{tab:convex_functions}). As a consequence, we argue that when GMD is executed by each agent independently, it could satisfy the first two desiderata \textbf{D1} and \textbf{D2} presented in the Introduction. However, as mentioned in Section~\ref{sec:mc}, since there is no explicit objective regarding different evaluation measures (and different solution concepts) arises in this \enquote{decentralized} execution process, GMD itself cannot satisfy well the last two desiderata \textbf{D3} and \textbf{D4}. To address the challenges, our solution is the zero-order meta-controller (MC) which dynamically adjusts the hyper-parameters conditional on the evaluation measures (Section~\ref{sec:mc}). In this section, we present the details of different MCs.

\textbf{Direction-Guided Random Search (DRS).} Our DRS method is obtained by applying the \textit{direction-guided update} (Section~\ref{sec:mc}) to the existing RS method presented in~\cite{wang2022zarts}. Specifically, at the iteration $k$, we first sample $D$ candidate updates $\{\bm{u}^j\}_{j=1}^{D}$ from a spherically symmetric distribution $\bm{u}^j \sim q$. Then, we update $\bm{\alpha}$ as follows:
\begin{equation}
\label{eq:drs}\tag{\textbf{DRS}}
\begin{aligned}
    & \bm{\alpha}_{+}^j = \text{CLIP}_{\iota}^{1}(\bm{\alpha}+\mu\bm{u}^j), \; \bm{\alpha}^j_{-} = \text{CLIP}_{\iota}^{1}(\bm{\alpha}-\mu\bm{u}^j), \; 1\leq j\leq D,\\
    & \bm{\pi}^j_{+} = \text{GMD}(\bm{\alpha}^j_{+}), \; \bm{\pi}^j_{-} = \text{GMD}(\bm{\alpha}^j_{-}), \; 1\leq j\leq D,\\
    & \delta^j = \mathcal{L}(\bm{\pi}^j_{+})-\mathcal{L}(\bm{\pi}^j_{-}), \; 1\leq j\leq D,\\
    & \bm{u}^{*} = -\sum\nolimits_{j=1}^{D} \text{Sgn}(\delta^j)\bm{u}^j,\\
    &  \bm{\alpha}\gets\text{CLIP}_{\iota}^{1}(\bm{\alpha}+\bm{u}^{*}).
\end{aligned}
\end{equation}
$\text{Sgn}(z)$ is defined as: $\text{Sgn}(z)=1$ if $z>0$, $\text{Sgn}(z)=-1$ if $z<0$, otherwise, $\text{Sgn}(z)=0$. $\mu$ is the smoothing parameter determining the radius of the sphere. $\text{CLIP}_{\iota}^{1}$ is the element-wise clipping operation defined as: $\text{CLIP}_{\iota}^{1}(z)=\iota$ if $z<\iota$, $\text{CLIP}_{\iota}^{1}(z)=1$ if $z>1$, otherwise, $\text{CLIP}_{\iota}^{1}(z)=z$, where $0<\iota<1$. Note that the clipping operation which bounds $\bm{\alpha}$ above $\iota>0$ is necessary as the term $B$ is used as the denominator in Eq. (\ref{eq:pi_a}). In addition, the operation $\text{Sgn}(\cdot)$ plays an important role and differentiates our DRS from the vanilla RS~\cite{wang2022zarts}. Intuitively, in the games where the value of the evaluation measure $\mathcal{L}$ is extremely small and converges quickly, the magnitude of $\delta^j$ would be too small to derive an effective update. In contrast, by using the operation $\text{Sgn}(\cdot)$, the difference between the performance of $\bm{\alpha}^j_{+}$ and $\bm{\alpha}^j_{-}$ will only determine the \textit{update direction}, not the update magnitude, which could be more effective. 

\textbf{Random Search (RS).} The vanilla RS which is adapted from~\cite{wang2022zarts}. The only difference from DRS is it updates $\bm{\alpha}$ directly based on the performance difference $\delta^j$. Precisely, we have:
\begin{equation}
\label{eq:rs}\tag{\textbf{RS}}
\begin{aligned}
    & \bm{\alpha}_{+}^j = \text{CLIP}_{\iota}^{1}(\bm{\alpha}+\mu\bm{u}^j), \; \bm{\alpha}^j_{-} = \text{CLIP}_{\iota}^{1}(\bm{\alpha}-\mu\bm{u}^j), \; 1\leq j\leq D,\\
    & \bm{\pi}^j_{+} = \text{GMD}(\bm{\alpha}^j_{+}), \; \bm{\pi}^j_{-} = \text{GMD}(\bm{\alpha}^j_{-}), \; 1\leq j\leq D,\\
    & \delta^j = \mathcal{L}(\bm{\pi}^j_{+})-\mathcal{L}(\bm{\pi}^j_{-}), \; 1\leq j\leq D,\\
    & \bm{u}^{*} = -\sum\nolimits_{j=1}^{D} \delta^j\bm{u}^j,\\
    &  \bm{\alpha}\gets\text{CLIP}_{\iota}^{1}(\bm{\alpha}+\bm{u}^{*}).
\end{aligned}
\end{equation}

\textbf{GradientLess Descent (GLD).} This method is adapted from~\cite{wang2022zarts}. At the iteration $k$, we first sample $D$ candidate updates $\{\bm{u}^j\}_{j=1}^{D}$. Different from RS which samples the candidates from a fixed radius (the smoothing parameter $\mu$ in DRS and RS), we independently sample the candidates on spheres with various radiuses uniformly sampled from the interval $[r_L, r_H]$. Then, we update $\bm{\alpha}$ as follows:
\begin{equation}
\label{eq:gld}\tag{\textbf{GLD}}
\begin{aligned}
    & \bm{\alpha}_{+}^j = \text{CLIP}_{\iota}^{1}(\bm{\alpha}+\bm{u}^j), \; \bm{\pi}^j_{+} = \text{GMD}(\bm{\alpha}^j_{+}), \; 1\leq j\leq D,\\
    & j^{*} = \arg\min_j\{\mathcal{L}(\bm{\pi}^j_{+})\}_{j=1}^{D},\\
    & \bm{u}^{*} = \bm{u}^{j^{*}},\\
    &  \bm{\alpha}\gets\text{CLIP}_{\iota}^{1}(\bm{\alpha}+\bm{u}^{*}).
\end{aligned}
\end{equation}
Intuitively, by comparing the performance of $D$ candidates, $\bm{\alpha}$ is updated by the candidate with the smallest value of $\mathcal{L}$. 

\textbf{GradientLess Descent with Summation (GLDS).} Different from GLD which uses only one of the $D$ samples to update the $\bm{\alpha}$, we can follow the idea of RS/DRS to take all the candidates into account by summation. Specifically, let $\mathcal{L}(\pi_k)$ denote the performance of the current policy $\pi_k$, then we have:
\begin{equation}
\label{eq:glds}\tag{\textbf{GLDS}}
\begin{aligned}
    & \bm{\alpha}_{+}^j = \text{CLIP}_{\iota}^{1}(\bm{\alpha}+\bm{u}^j), \; \bm{\pi}^j_{+} = \text{GMD}(\bm{\alpha}^j_{+}), \; 1\leq j\leq D,\\
    & \delta^j = \mathcal{L}(\bm{\pi}^j_{+})-\mathcal{L}(\bm{\pi}_{k}), \; 1\leq j\leq D,\\
    & \bm{u}^{*} = -\sum\nolimits_{j=1}^{D} \delta^j\bm{u}^j,\\
    &  \bm{\alpha}\gets\text{CLIP}_{\iota}^{1}(\bm{\alpha}+\bm{u}^{*}).
\end{aligned}
\end{equation}

\textbf{Direction-Guided GLDS (DGLDS).} Applying the direction-guided update to the GLDS, we can get this method. Precisely, let $\mathcal{L}(\pi_k)$ denote the performance of the current policy $\pi_k$, then we have:
\begin{equation}
\label{eq:dglds}\tag{\textbf{DGLDS}}
\begin{aligned}
    & \bm{\alpha}_{+}^j = \text{CLIP}_{\iota}^{1}(\bm{\alpha}+\bm{u}^j), \; \bm{\pi}^j_{+} = \text{GMD}(\bm{\alpha}^j_{+}), \; 1\leq j\leq D,\\
    & \delta^j = \mathcal{L}(\bm{\pi}^j_{+})-\mathcal{L}(\bm{\pi}_{k}), \; 1\leq j\leq D,\\
    & \bm{u}^{*} = -\sum\nolimits_{j=1}^{D} \text{Sgn}(\delta^j)\bm{u}^j,\\
    &  \bm{\alpha}\gets\text{CLIP}_{\iota}^{1}(\bm{\alpha}+\bm{u}^{*}).
\end{aligned}
\end{equation}

As the meta-controller needs to evaluate the performance of the candidates, extra computational cost is required. In our experiments, to trade-off between the learning performance and running time, we update $\bm{\alpha}$ every $\kappa \ge 1$ iteration. In addition, during the first $M-1$ iterations, i.e., $k < M$, as there are only $k<M$ historical policies, we set $\alpha_{\tau} = \frac{1}{k}$ for $0\leq\tau \leq k-1$. In other words, MC will start to update $\bm{\alpha}$ only after $M$ iterations. Algorithm~\ref{alg:cmd} in the main text is the simplified version which shows the primary principle of CMD. In Algorithm~\ref{alg:cmd_app}, we present the full details of CMD.
\begin{algorithm}[ht]
\caption{Configurable Mirror Descent (CMD)}
\label{alg:cmd_app}
\begin{algorithmic}[1]
\STATE Given $\mathcal{L}$, $\psi$, initial (joint) policy $\bm{\pi}_1$, $M$, $D$, $\epsilon$, $\iota$,
\FOR{$k=1, \cdots, K$}
\IF{$k\leq M$}
\STATE $\alpha_{\tau} = \frac{1}{k}$, $\forall 0\leq\tau\leq k-1$\;
\ELSE
\IF{$k\%\kappa=0$}
\STATE Sample $D$ candidates $\{\bm{\alpha}^j\}_{j=1}^{D}$\;
\STATE Derive new joint policies $\{\bm{\pi}^j=\text{GMD}(\bm{\alpha}^j)\}_{j=1}^{D}$\;
\STATE Evaluate new joint policies $\{\mathcal{L}(\bm{\pi}^j)\}_{j=1}^{D}$\;
\STATE Update $\bm{\alpha}$ based on $\{\mathcal{L}(\bm{\pi}^j)\}_{j=1}^{D}$\;
\ENDIF
\ENDIF
\STATE Compute $\bm{\pi}_{k+1}$ via GMD with the updated $\bm{\alpha}$\;
\ENDFOR
\end{algorithmic}
\end{algorithm}

\clearpage
\section{\textsc{GameBench}\label{app:gamebench}}

In this section, we present the details of \textsc{GameBench} (see Figure~\ref{fig:game_bench} for an overview). In Section~\ref{app:gb_motivate_desiderata}, we discuss the motivation and desiderata by briefly reviewing the games that have been employed to test existing MD algorithms. In Section~\ref{app:gb_games}, we present the details of the construction of all 15 games. Finally, in Section~\ref{app:gb_eval_measure}, we present the evaluation measures considered in this work.

\subsection{Motivation and Desiderata\label{app:gb_motivate_desiderata}}
As mentioned in Section~\ref{sec:gamebench}, existing benchmarks for decision making are typically specialized for some specific categories. Furthermore, running MD algorithms on these benchmarks could be computationally prohibitive as the number of decision points in the environments could be extremely large. On the other hand, though MD algorithms have been demonstrated powerful in single-agent RL~\cite{tomar2022mirror} and two-player zero-sum games~\cite{wibisono2022alternating,kozuno2021learning,lee2021lastiterate,liu2022equivalence,jain2022matrix,ao2023asynchronous,liu2023the,cen2023faster,sokota2023unified} in recent works, their experiments are typically conducted on a handful of games. It remains elusive how will these MD algorithms perform when applied to other categories of decision-making problems. In Table~\ref{tab:game_review}, we briefly review the games that have been used in some recent works.
\begin{table}[ht]
\centering
\caption{The games that have been used in recent works on MD algorithms. Note that this list does not include the games that are used to benchmark deep learning-based algorithms in these references. $^1$This game is made to be a general-sum game via a tie-breaking mechanism in this reference. $^2$This game is made to be a zero-sum game in this reference.}
\label{tab:game_review}
\vskip 0.1in
\begin{tabular}{l|l|l}
\toprule
Reference & Game & Category\\
\midrule
\multirow{4}{*}{\cite{sokota2023unified}} & Kuhn Poker & Two-Player Zero-Sum\\
 & Leduc Poker & Two-Player Zero-Sum\\
 & 2x2 Abrupt Dark Hex & Two-Player Zero-Sum\\
 & 4-Sided Liar’s Dice & Two-Player Zero-Sum\\
\midrule
\multirow{2}{*}{\cite{liu2023the}} & Kuhn Poker & Two-Player Zero-Sum \\
 & Leduc Poker & Two-Player Zero-Sum\\
\midrule
\multirow{2}{*}{\cite{anagnostides2022last}} & Kuhn Poker & Two-Player Zero-Sum\\
 & Leduc Poker & Two-Player Zero-Sum\\
\midrule
\multirow{4}{*}{\cite{anagnostides2022optimistic}} & Sheriff & Two-Player General-Sum\\
 & Battleship & Two-Player General-Sum \\
 & Goofspiel$^1$ & Two-Player General-Sum \\
 & Liar’s Dice & Two-Player Zero-Sum \\
\midrule
\multirow{3}{*}{\cite{lee2021lastiterate}} & Kuhn Poker & Two-Player Zero-Sum \\
 & Leduc Poker & Two-Player Zero-Sum \\
 & Pursuit-Evasion & Two-Player Zero-Sum\\
\midrule
\multirow{4}{*}{\cite{liu2022equivalence}} & Leduc Poker & Two-Player Zero-Sum\\
 & Goofspiel & Two-Player Zero-Sum \\
 & Liar's Dice & Two-Player Zero-Sum \\
 & Battleship$^2$ & Two-Player Zero-Sum \\
\bottomrule
\end{tabular}
\end{table}

In view of the above facts, we aim to construct a novel benchmark which should satisfy two desiderata (\textbf{D5} and \textbf{D6} presented in the Introduction): i) it should cover all categories of decision making (comprehensive), and ii) the games are relatively simple and running MD algorithms on these games does not require much computational resource (academic-friendly).

\subsection{Games\label{app:gb_games}}

In this section, we present the details of the construction of all 15 games in our \textsc{GameBench}. All the games are divided into 5 categories: single-agent, cooperative multi-agent, competitive multi-agent zero-sum (zero-sum), competitive multi-agent general-sum (general-sum), and mixed cooperative and competitive (MCC) categories. In Table~\ref{tab:sel_games}, we give an overview of all the games. We curate the \textsc{GameBench} on top of OpenSpiel~\cite{lanctot2019openspiel}. For cooperative, zero-sum, and general-sum categories, we construct the game by passing the configurations to the games implemented in OpenSpiel. The configurations for these games are deliberately selected such that the instances of these games are academic-friendly (i.e., their numbers of decision points are not too large). For single-agent and MCC categories, we obtain the games by modifying the original games in OpenSpiel. In the following, we present the details of each category.

\begin{table}[ht]
    \centering
    \caption{The games and their statistics in \textsc{GameBench}. $N$ is the number of players and \enquote{\#DP} stands for the number of decision points.}
    \label{tab:sel_games}
    \vskip 0.1in
    \begin{tabular}{lllccl}
    \toprule
    Category & Name of Game w/ Config. & Shorthand & $N$ & \#DP & Evaluation Measure\\
    \midrule
    \multirow{3}{*}{Single-Agent} &single\_agent\_kuhn\_a & Kuhn-A & 1 & 6 & OptGap\\
    &single\_agent\_kuhn\_b & Kuhn-B & 1 & 6 & OptGap\\
    &single\_agent\_goofspiel & Goofspiel-S & 1 & 8 & OptGap\\
    \midrule
    \multirow{3}{*}{Cooperative} &tiny\_hanabi\_game\_a & TinyHanabi-A & 2 & 8 & OptGap\\
    &tiny\_hanabi\_game\_b & TinyHanabi-B & 2 & 6 & OptGap\\
    &tiny\_hanabi\_game\_c & TinyHanabi-C & 2 & 6 & OptGap\\
    \midrule
    \multirow{3}{*}{Zero-Sum} &kuhn\_poker(players=3) & Kuhn & 3 & 48 & NashConv, CCEGap\\
    &leduc\_poker(players=2) & Leduc & 2 & 936 & NashConv\\
    &goofspiel(players=3) & Goofspiel & 3 & 30 & NashConv, CCEGap\\
    \midrule
    \multirow{3}{*}{General-Sum} &bargaining(max\_turns=2) & Bargaining & 2 & 178 & NashConv, SW\\
    &trade\_comm(num\_items=2) & TradeComm & 2 & 22 & NashConv, SW\\
    &battleship & Battleship & 2 & 210 & NashConv, SW\\
    \midrule
    \multirow{3}{*}{MCC} &mix\_kuhn\_3p\_game\_a & MCCKuhn-A & 3 & 48 & NashConv\\
    &mix\_kuhn\_3p\_game\_b & MCCKuhn-B & 3 & 48 & NashConv\\
    &mix\_goofspiel\_3p & MCCGoofspiel & 3 & 30 & NashConv\\
    \bottomrule
    \end{tabular}
\end{table}

\textbf{Single-Agent.} We construct three single-agent games: Kuhn-A, Kuhn-B, and Goofspiel-S, from the original two-player Kuhn poker and Goofspiel in OpenSpiel. Consider a two-player Kuhn poker game. To obtain a single-agent counterpart, we fix one player's policy as the uniform policy (called the background player) while only updating the other player's policy (called the focal player) at each iteration. In Kuhn-A, player 1 is selected as the focal player while in Kuhn-B, player 2 is chosen as the focal player, as the two players are asymmetric~\cite{kuhn1950simplified}. Similarly, we can get Goofspiel-S. As the two players are symmetric in Goofspiel~\cite{ross1971goofspiel}, we choose player 1 as the focal player without loss of generality.

\begin{figure}[ht]
\centering
\includegraphics[width=0.6\textwidth]{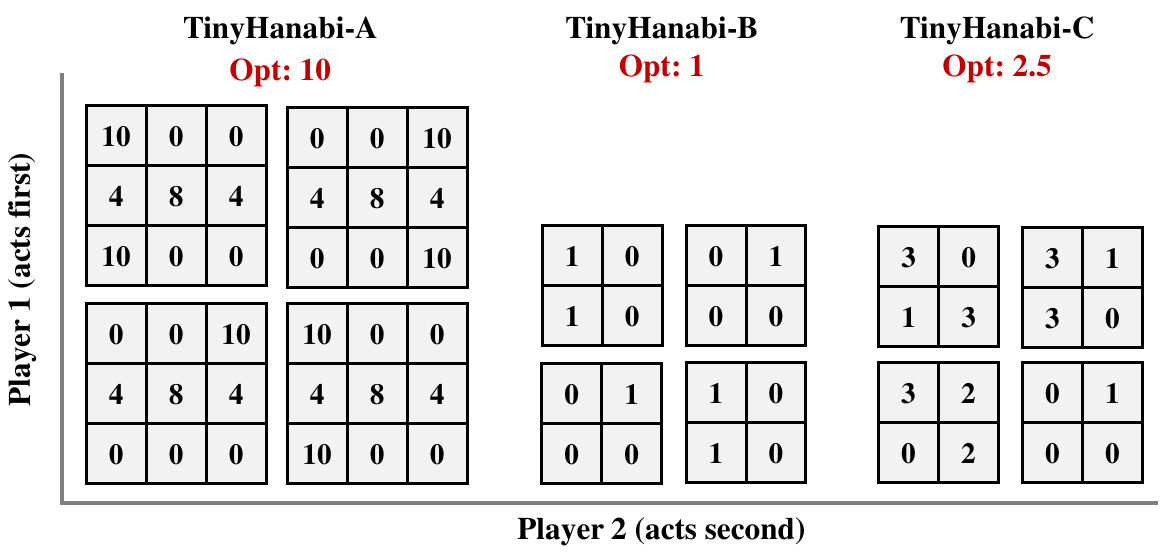}
\caption{Payoff matrices and optimal values of the three tiny Hanabi games.}
\label{fig:payoff_coop}
\end{figure}

\textbf{Cooperative.} For cooperative games, we consider the following three \textit{two-player} tiny Hanabi games~\cite{foerster2019bayesian,sokota2021solving}: TinyHanabi-A, TinyHanabi-B, and TinyHanabi-C. The payoff matrices along with the optimal values of these games are given in Figure~\ref{fig:payoff_coop}. These games are easy to obtain in OpenSpiel by setting the three parameters: \texttt{num\_chance}, \texttt{num\_actions}, and \texttt{payoff}. For \texttt{num\_chance}, they are 2, 2, and 2, respectively. For \texttt{num\_actions}, they are 3, 2, and 2, respectively.

\textbf{Competitive Zero-Sum and General-Sum.} We consider the following three zero-sum games: \textit{three-player} Kuhn, \textit{two-player} Leduc, and \textit{three-player} Goofspiel, and the following three general-sum games: \textit{two-player} Battleship~\cite{farina2020coarse}, \textit{two-player} TradeComm~\cite{sokota2021solving}, and \textit{two-player} Bargaining~\cite{lewis2017deal}, which are implemented in OpenSpiel. The configurations of these games are given in the second column in Table~\ref{tab:sel_games}. Note that in contrast to most of the existing works which only focus on two-player games, we set the number of players to more than two players in some of the games: Kuhn and Goofspiel are three-player games.

\textbf{Mixed Cooperative and Competitive (MCC).} We construct the following \textit{three-player} MCC games: MCCKuhn-A, MCCKuhn-B, and MCCGoofspiel, from the original three-player Kuhn poker and three-player Goofspiel in OpenSpiel. Consider a three-player Kuhn poker game. To obtain an MCC counterpart, we partition the three players into two teams: Team 1 includes two players while Team 2 only consists of one player (i.e., two \textit{vs.} one). When computing the rewards of the players, in Team 1, each player will get the average reward of the team. Precisely, let $r^{\text{team}} = r^1 + r^2$ denote the team reward which is the sum of the original rewards of the two team members. Then, the true rewards of the two players are $\tilde{r}^1 = \tilde{r}^2 = r^{\text{team}} / 2$. In MCCKuhn-A, Team 1 includes players 1 and 2 (i.e., \{1, 2\} \textit{vs.} 3), while in MCCKuhn-B, Team 1 includes players 1 and 3 (i.e., \{1, 3\} \textit{vs.} 2). Similarly, we can get MCCGoofspiel in the same manner. 

As shown in Table~\ref{tab:sel_games}, the number of decision points (\#DP) varies across different categories, which shows that \textsc{GameBench} includes diverse enough environments as, to some extent, the number of decision points reflects the difficulty of the game.

\subsection{Evaluation Measures\label{app:gb_eval_measure}}

As shown in Figure~\ref{fig:game_bench}, we consider multiple evaluation measures in \textsc{GameBench}. There are two types of measures: i) the notion of \textit{optimality}, including OptGap and social welfare, and ii) the notion of \textit{equilibrium}, including NashConv and CCEGap. In the last column of Table~\ref{tab:sel_games}, we present the measures employed in each game. In single-agent and cooperative categories, we use OptGap as the measure which captures the distance of the current (joint) policy to the optimal (joint) policy. In the other three categories, the primary measure is NashConv which captures the distance of the current joint policy to the \textit{Nash equilibrium}. In addition, we also consider other solution concepts and evaluation measures in some of the games. For zero-sum Kuhn and Goofspiel, as there are three players, we also consider the measure CCEGap which captures the distance of the current joint policy to the coarse correlated equilibrium (CCE). For general-sum games, we also consider the social welfare (SW) of all the agents.

Except for the MCC category, all the measures can be easily computed by using the built-in implementation functions in OpenSpiel. However, to compute the NashConv in the MCC games, we need to compute the best response policy of the team, i.e., a joint policy of the team members, rather than the policy of a single agent. This is incompatible with the built-in implementation in OpenSpiel, which only computes the best response policy of a single agent. In other words, if we directly adopt the built-in implementation, the NashConv will correspond to the original three-player game, not the modified game. Unfortunately, computing the exact joint policy of the team members is not easy in practice. Nevertheless, it is worth noting that from our experiments, we found that MMD-KL can effectively solve cooperative decision-making problems. As a result, we can apply MMD-KL to compute the approximate best response of the team as it is a purely cooperative environment from the team's perspective (the other team's policy is fixed when computing the best response of the team). For a team that only has a single player, we use the built-in implementation in OpenSpiel to compute the exact best response policy of the player. In summary, during the policy learning process, when the evaluation of the current joint policy is needed, we use MMD-KL as a subroutine to compute a team's approximate best response while using built-in implementation to compute a single player's exact best response. In the MMD-KL subroutine, the starting point of the best response is set to the current joint policy of the team members. In experiments, to balance the accuracy of the approximate best response and running time, the number of updates in the MMD-KL subroutine is set to 100 (the returned joint policy can be also called a better response).

For example, in MCCKuhn-A, suppose the current joint policy is $\bm{\pi}=\bm{\pi}_{\text{team}}\times\pi_3$ where $\bm{\pi}_{\text{team}}=\pi_1\odot\pi_2$ is the team's joint policy. The built-in implementation in OpenSpiel can only compute the best response policy for every single agent and hence, the resulting $\text{NashConv}(\bm{\pi})=\sum_{i=1}^{3}[V_i(\nu, \pi_i^{\text{BR}}\times\bm{\pi}_{-i}) - V_i(\nu, \bm{\pi})]$ corresponds to the original three-player game. In contrast, in our method, we use MMD-KL to compute the team's best response rather than the single agent's. Therefore, the NashConv of $\bm{\pi}$ is:
\begin{equation}
\begin{aligned}
    \text{NashConv}(\bm{\pi})&=V_{\text{team}}(\nu, \bm{\pi}_{\text{team}}^{\text{BR}}\times\pi_3) - V_{\text{team}}(\nu, \bm{\pi}) \\
    & + V_3(\nu, \bm{\pi}_{\text{team}}\times\pi_3^{\text{BR}}) - V_3(\nu, \bm{\pi}),
\end{aligned}
\end{equation}
where $\bm{\pi}_{\text{team}}^{\text{BR}}$ is the team's BR policy computed via MMD-KL given that player 3 is fixed to $\pi_3$ (that is, player 3 is a part of the environment from the team's perspective). As players 1 and 2 are fully cooperative, they share the same value $V_{\text{team}}$.

\clearpage
\section{More Experimental Results\label{app:more_results}}

In this section, we provide more experimental details, results, and analysis. We briefly summarize each section below.
\begin{itemize}[left=0.2mm, itemsep=0mm, topsep=0mm]
    \item Section~\ref{app:exp_setup}. More details on the experimental setup, including the hyper-parameter settings for different methods.
    \item Section~\ref{app:ablation_K_mu}. Searching of $M$ (the number of previous policies) and $\mu$ (the smoothing parameter in DRS) (\textbf{D1} and \textbf{D2}).
    \item Section~\ref{app:perf_vs_num_joint_acts}. Investigation of performance w.r.t. the number of joint actions (\textbf{D1} and \textbf{D2}).
    \item Section~\ref{app:diff_gmd_alpha_sc}. Investigation of GMD with different heuristic strategies for adjusting $\bm{\alpha}$ (\textbf{D1} and \textbf{D2}).
    \item Section~\ref{app:diff_mcs}. Investigation of different meta-controllers (\textbf{D1} and \textbf{D2}).
    \item Section~\ref{app:diff_bregman}. Investigation of different Bregman divergences (\textbf{D1} and \textbf{D2}).
    \item Section~\ref{app:effect_of_magnet}. Investigation of the effectiveness of adding the magnet policy (\textbf{D1} and \textbf{D2}).
    \item Section~\ref{app:diff_measure}. Investigation of different evaluation measures and different solution concepts (\textbf{D3} and \textbf{D4}).
    \item Section~\ref{app:runtime}. Analysis of the computational complexity for running different algorithms on \textsc{GameBench} (\textbf{D5} and \textbf{D6}).
\end{itemize}

\subsection{Experimental Setup\label{app:exp_setup}}

\textbf{Hyper-parameters.} Table~\ref{tab:hyper-parameters} provides the default values of hyper-parameters used in different methods. In the RS-type meta-controllers (RS and DRS), the spherically symmetric distribution $q$ is a standard multivariate normal distribution $\mathbb{N}(\bm{0}, \textbf{I})$. For CMD/GMD, there are two critical hyper-parameters: the number of previous policies $M\ge1$ and the smoothing parameter $\mu$ in DRS. In Section~\ref{app:ablation_K_mu}, we perform an ablation study to determine their default values (given in Table~\ref{tab:hyper-parameters}), which will be fixed in other experiments. The specific setups for each experiment will be given in each of the following sections.

\textbf{Baselines.} We consider the MMD-type (MMD-KL and MMD-EU) and CFR-type (CFR and CFR+) algorithms as the baselines. It is worth noting that CFR-type algorithms can be also applied to single-agent and cooperative categories.

\textbf{Computational Resources.} Experiments are performed on a machine with a 24-core i9 and NVIDIA A4000. For CMD, the results are obtained with 3 random seeds. For other methods, as there is no randomness, no multiple runs are needed.

\begin{table}[ht]
    \centering
    \caption{Default values of the hyper-parameters in different methods. All the hyper-parameters in GMD -- $C$, $\iota$, and $M$ -- are also used in CMD. For CMD, its hyper-parameters also include i) $D$ (the number of samples) and $\kappa$ (update interval), which are shared for different MCs, ii) $\mu$ in the DRS and RS, and iii) $r_L$ and $r_H$ in the GLD, GLDS, and DGLDS.}
    \label{tab:hyper-parameters}
    \vskip 0.1in
    \begin{tabular}{lcc|ccc|ccccc|ccc}
    \toprule
    & & & \multicolumn{8}{c|}{CMD} & \multicolumn{3}{c}{MMD-KL/-EU}\\
    \cmidrule(r){4-11}
    \cmidrule(r){12-14}
    & & & \multicolumn{3}{c|}{GMD} & \multicolumn{2}{c}{Shared} & (D)RS & \multicolumn{2}{c|}{(D)GLD(S)} & & & \\
    Game & $K$ & $\epsilon$ & $C$ & $\iota$ & $M$ & $D$ & $\kappa$ & $\mu$ & $r_L$ & $r_H$ & $\xi$ & $\eta$ & $\tilde{\eta}$ \\
    \midrule
    Kuhn-A & 100000 & 1e-10 & 50 & 1e-6 & 1 & 5 & 10 & 0.05 & 0.01 & 0.05 & 1 & 0.1 & 0.05 \\
    Kuhn-B & 100000 & 1e-10 & 50 & 1e-6 & 1 & 5 & 10 & 0.05 & 0.01 & 0.05 & 1 & 0.1 & 0.05\\
    Goofspiel-S & 100000 & 1e-10 & 50 & 1e-6 & 1 & 5 & 10 & 0.05 & 0.01 & 0.05 & 1 & 0.1 & 0.05\\
    \midrule
    TinyHanabi-A & 100000 & 1e-10 & 50 & 1e-6 & 3 & 5 & 10 & 0.05 & 0.01 & 0.05 & 1 & 0.1 & 0.05\\
    TinyHanabi-B & 100000 & 1e-10 & 50 & 1e-6 & 1 & 5 & 10 & 0.05 & 0.01 & 0.05 & 1 & 0.1 & 0.05\\
    TinyHanabi-C & 100000 & 1e-10 & 50 & 1e-6 & 1 & 5 & 10 & 0.05 & 0.01 & 0.05 & 1 & 0.1 & 0.05\\
    \midrule
    Kuhn & 100000 & 1e-10 & 50 & 1e-6 & 5 & 5 & 10 & 0.01 & 0.01 & 0.05 & 1 & 0.1 & 0.05\\
    Leduc & 100000 & 1e-10 & 50 & 1e-6 & 3 & 5 & 10 & 0.05 & 0.01 & 0.05 & 1 & 0.1 & 0.05\\
    Goofspiel & 100000 & 1e-10 & 50 & 1e-6 & 3 & 5 & 10 & 0.01 & 0.01 & 0.05 & 1 & 0.1 & 0.05\\
    \midrule
    Bargaining & 100000 & 1e-10 & 50 & 1e-6 & 5 & 5 & 10 & 0.05 & 0.01 & 0.05 & 1 & 0.1 & 0.05\\
    TradeComm & 100000 & 1e-10 & 50 & 1e-6 & 1 & 5 & 10 & 0.01 & 0.01 & 0.05 & 1 & 0.1 & 0.05\\
    Battleship & 100000 & 1e-10 & 50 & 1e-6 & 1 & 5 & 10 & 0.05 & 0.01 & 0.05 & 1 & 0.1 & 0.05\\
    \midrule
    MCCKuhn-A & 100000 & 1e-10 & 50 & 1e-6 & 1 & 5 & 10 & 0.01 & 0.01 & 0.05 & 1 & 0.1 & 0.05\\
    MCCKuhn-B & 100000 & 1e-10 & 50 & 1e-6 & 1 & 5 & 10 & 0.01 & 0.01 & 0.05 & 1 & 0.1 & 0.05\\
    MCCGoofspiel & 100000 & 1e-10 & 50 & 1e-6 & 1 & 5 & 10 & 0.01 & 0.01 & 0.05 & 1 & 0.1 & 0.05\\
    \bottomrule
    \end{tabular}
\end{table}

\clearpage
\subsection{Number of Historical Policies and Smoothing Parameter\label{app:ablation_K_mu}}

In this section, we explore the influence of the number of previous policies $M$ and the smoothing parameter $\mu$ in DRS on the learning performance. We consider $M\in\{1, 3, 5\}$ and $\mu\in\{0.01, 0.05\}$ and thus, there are 6 combinations of $(M, \mu)$. Note that it would be impractical to enumerate all the combinations as $M$ can be any integer greater than 0 and $\mu$ can be any real number greater than 0. 

\begin{figure}[ht]
\centering
\includegraphics[width=\textwidth]{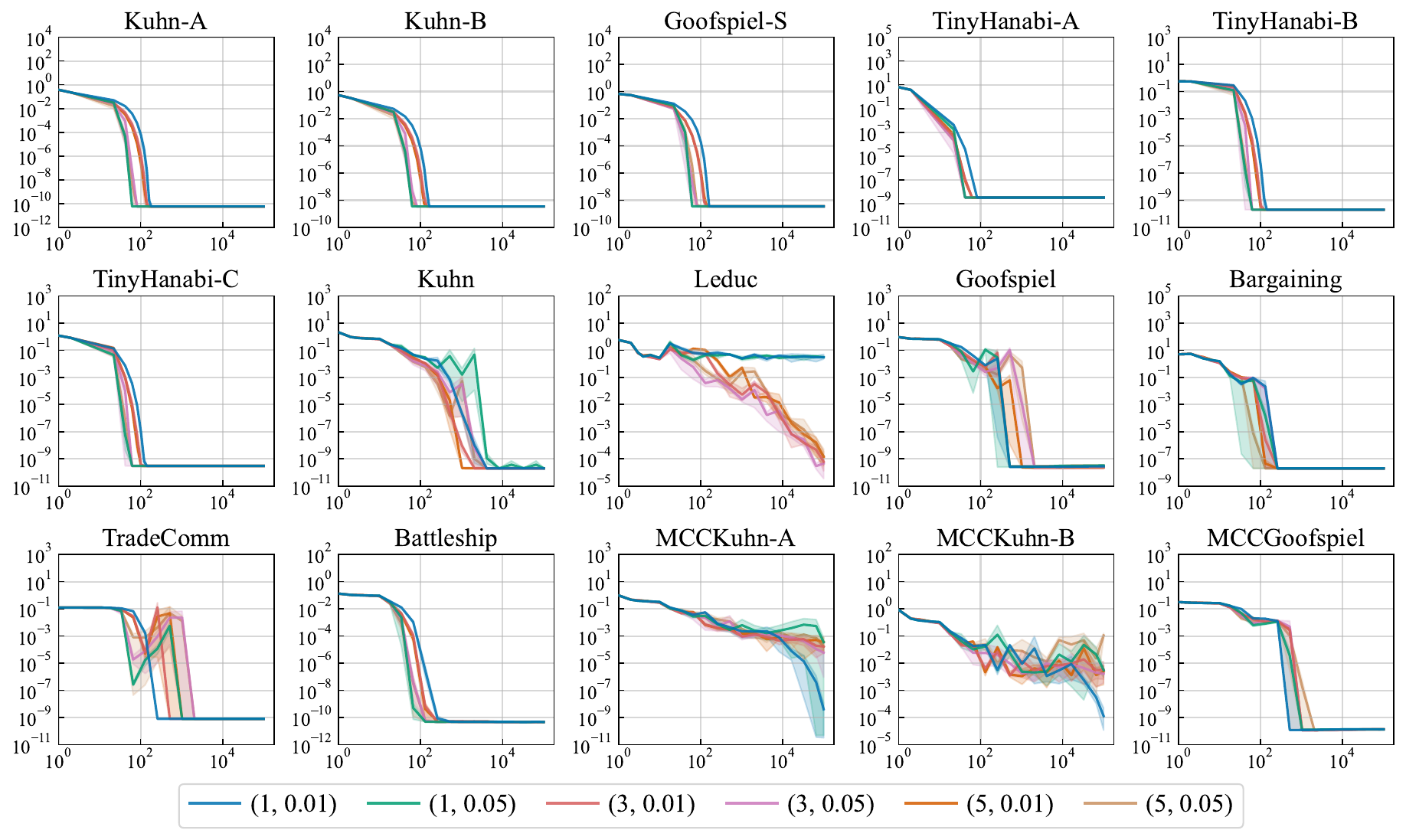}
\caption{Experimental results for the combinations of $(M, \mu)$. The first 6 figures correspond to single-agent and cooperative categories where the $y$-axis is \textit{OptGap}. The rest figures correspond to other categories where the $y$-axis is \textit{NashConv}. For all the figures, the $x$-axis is the number of iterations.}
\label{fig:k-mu}
\end{figure}

The experimental results are shown in Figure~\ref{fig:k-mu}. From the results, we can see that different decision-making problems may require different $M$ and $\mu$. Notably, $M=1$, i.e., only considering the current policy when deriving the new policy which is common in existing MD algorithms, is not always the optimal choice across different decision-making problems. For example, in the most difficult Leduc poker game, when $M=1$, CMD cannot decrease the NashConv, meaning that only considering the current policy is ineffective in solving this game. By comparison, we determine the default values of $M$ and $\mu$ for different games, which are given in Table~\ref{tab:hyper-parameters} and will be fixed in other experiments.

\clearpage
\subsection{Performance w.r.t. the Number of Joint Actions\label{app:perf_vs_num_joint_acts}}
\begin{figure}[ht]
\centering
\includegraphics[width=\textwidth]{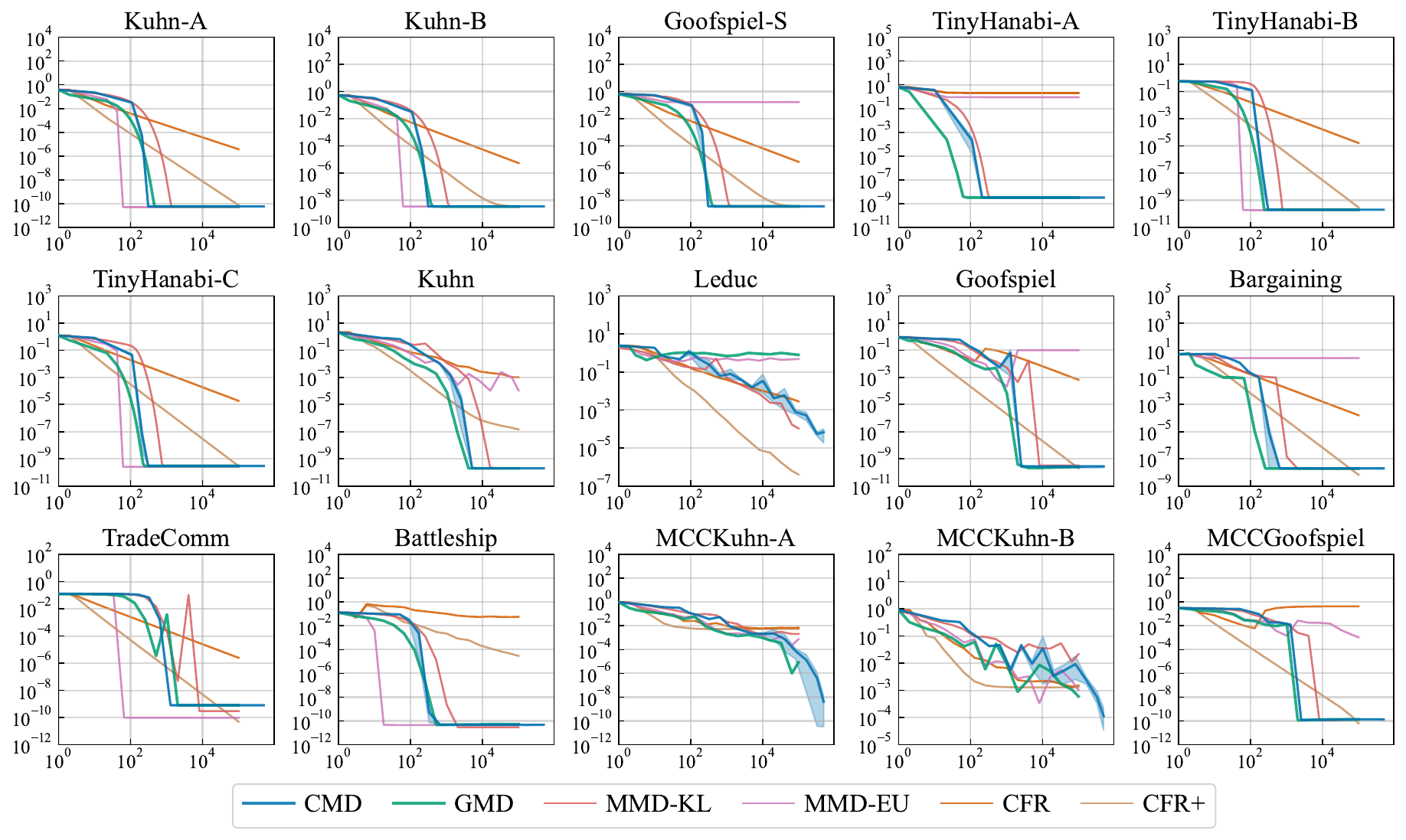}
\caption{Performance of different methods w.r.t. the number of joint actions. The first 6 figures correspond to single-agent and cooperative categories where the $y$-axis is \textit{OptGap}. The rest figures correspond to other categories where the $y$-axis is \textit{NashConv}. For all the figures, the $x$-axis is the number of iterations.}
\label{fig:perf_vs_num_joint_act}
\end{figure}

In Figure~\ref{fig:perf_vs_num_joint_act}, we plot the performance of different methods with respect to the number of joint actions involved in each iteration. As both MD-type and CFR-type algorithms will traverse the whole game tree, the number of joint actions for a given joint policy is the same. Therefore, in the figure, we only need to change the scale of the $x$-axis for the CMD by multiplying the constant $D$ (the number of joint policies evaluated at each iteration), while keeping the scales of other methods unchanged. We note that as $D$ is small in our experiments ($D=5$, i.e., sample 5 candidate joint policies), the conclusions in terms of the number of iterations presented in the main text still hold in terms of the number of joint actions.

As discussed in Section~\ref{sec:conclusion}, one of the future directions of our work would be the development of a more efficient method for updating the $\bm{\alpha}$, e.g., a method that only needs to sample one candidate (in this case, the number of joint actions will be the same for both CMD and other baselines). Nevertheless, compared to baseline MD and CFR-type algorithms, CMD provides a feasible way to study different solution concepts and evaluation measures, though, in the current version, it requires evaluating multiple candidates at each iteration.

\clearpage
\subsection{Different Heuristic Strategies for Adjusting $\bm{\alpha}$ in GMD\label{app:diff_gmd_alpha_sc}}
\begin{figure}[ht]
\centering
\includegraphics[width=\textwidth]{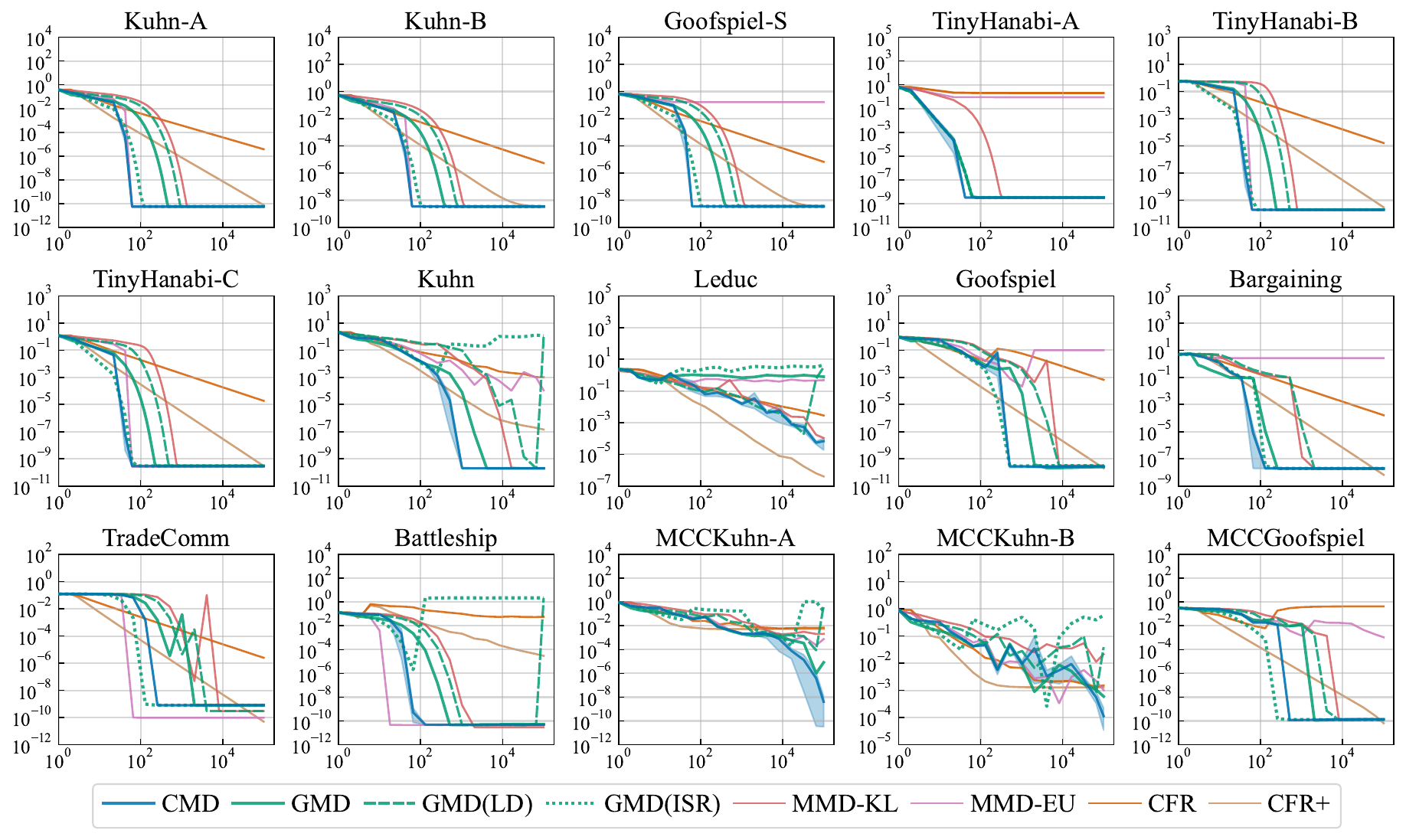}
\caption{Experimental results for GMD with different heuristic strategies for adjusting $\bm{\alpha}$. The first 6 figures correspond to single-agent and cooperative categories where the $y$-axis is \textit{OptGap}. The rest figures correspond to other categories where the $y$-axis is \textit{NashConv}. For all the figures, the $x$-axis is the number of iterations.}
\label{fig:diff_heu_alpha_sc}
\end{figure}

In the main text, the baseline method GMD employs a fixed strategy -- a uniform distribution -- to determine the value of $\bm{\alpha}$. In this section, we consider two more heuristic strategies: i) \enquote{GMD(LD)} denotes that the $\bm{\alpha}$ is linearly decayed with the iteration, and ii) \enquote{GMD(ISR)} denotes that the $\bm{\alpha}$ is decayed with the iteration in the form of inverse square root function $\alpha_{\tau}=\frac{1}{\sqrt{k}}$, where $k$ is the $k$-th iteration. The results are shown in Figure~\ref{fig:diff_heu_alpha_sc}. From the results, we can see that different heuristic strategies can perform differently in different decision-making scenarios; one can beat others in some scenarios while it can also be beaten by others in other scenarios. 

\clearpage
\subsection{Different Meta-Controllers\label{app:diff_mcs}}
\begin{figure*}[ht]
\centering
\includegraphics[width=\textwidth]{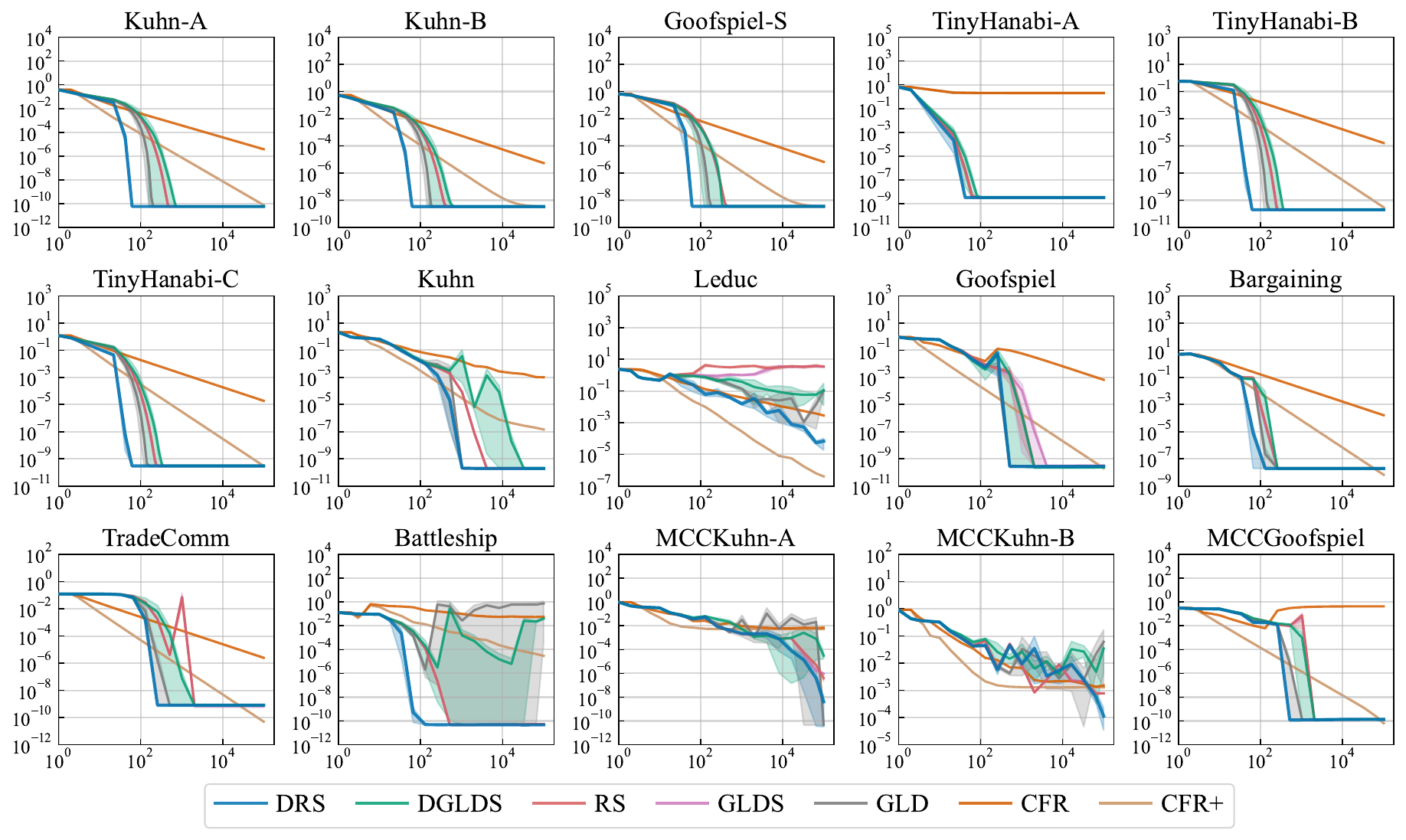}
\caption{Experimental results for different MCs. The first 6 figures correspond to single-agent and cooperative categories where the $y$-axis is \textit{OptGap}. The rest figures correspond to other categories where the $y$-axis is \textit{NashConv}. For all the figures, the $x$-axis is the number of iterations.}
\label{fig:diff-mc}
\end{figure*}

In this section, we investigate the effectiveness of different MCs, and the results are shown in Figure~\ref{fig:diff-mc}. From the results, we can see that DRS can consistently outperform all the other baseline MCs across almost all of the decision-making problems. Particularly, in Leduc and MCCKuhn-B, DRS achieves a significantly better convergent performance than other baseline MCs. Although in MCCKuhn-A, GLD finally converges to a lower NashConv than DRS, it can perform much worse in other games, e.g., in Battleship, GLD cannot decrease the NashConv, in Leduc and MCCKuhn-B, it only converges to a high value of NashConv. In other words, GLD cannot consistently work well across all the decision-making categories. In addition, in most of the games, the RS-type MCs typically perform better than the GLD-type MCs. We hypothesize that the RS-type MCs are more efficient in exploring the parameter space as they use more samples ($\bm{\alpha}_{+}^j$ and $\bm{\alpha}_{-}^j$ for each $\bm{u}^j$) to obtain the final update for the hyper-parameters.

\begin{figure*}[ht]
\centering
\includegraphics[width=0.91\textwidth]{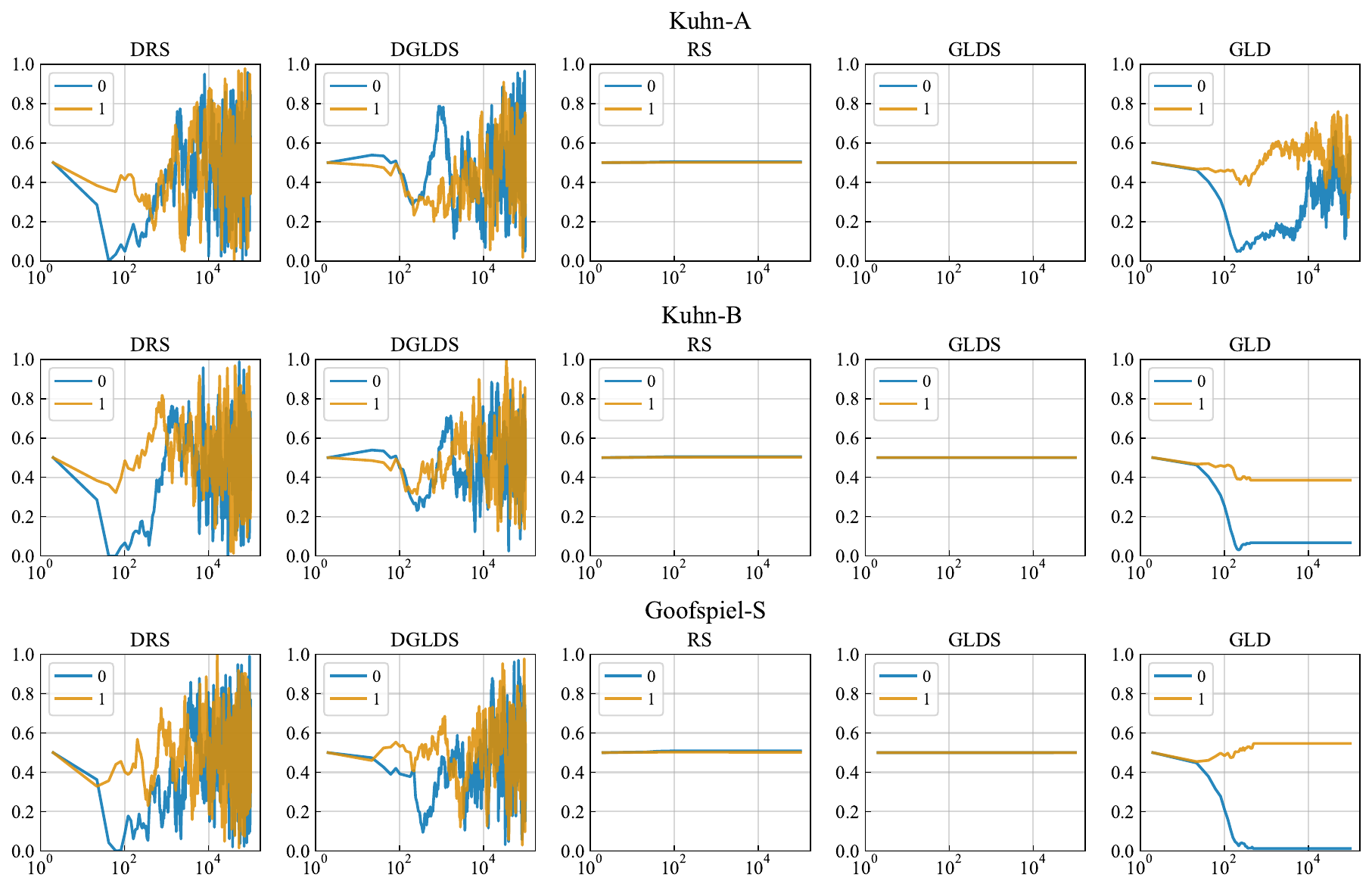}
\caption{The evolution of the hyper-parameter values of different MCs in the \textbf{Single-Agent} category. The $y$-axis is the value of $\bm{\alpha}$. The $x$-axis is the number of iterations.}
\label{fig:alpha-sc-single-agent}
\end{figure*}

\begin{figure*}[ht]
\centering
\includegraphics[width=0.91\textwidth]{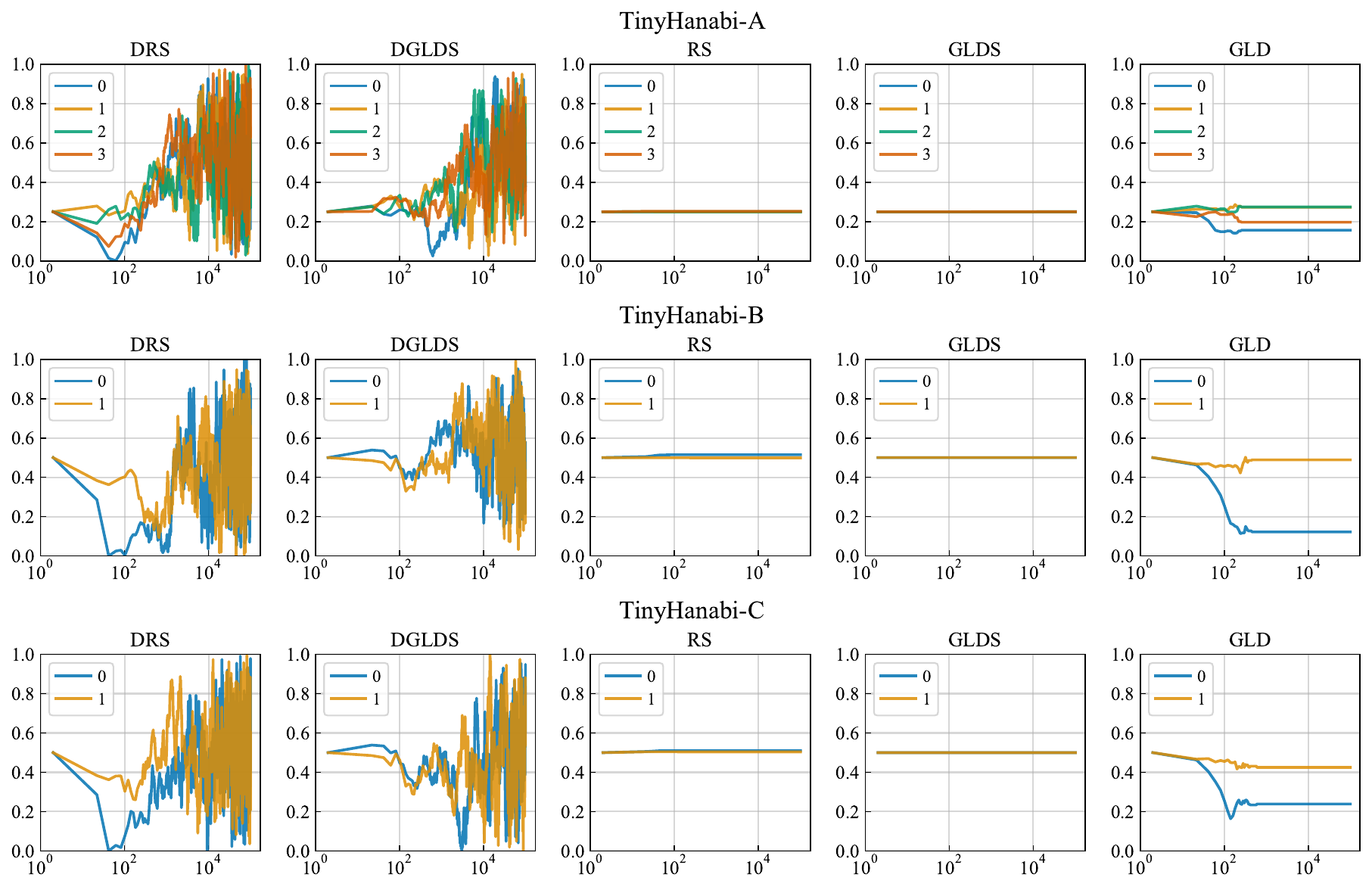}
\caption{The evolution of the hyper-parameter values of different MCs in the \textbf{Cooperative} category. The $y$-axis is the value of $\bm{\alpha}$. The $x$-axis is the number of iterations.}
\label{fig:alpha-sc-coop}
\end{figure*}

\begin{figure*}[ht]
\centering
\includegraphics[width=0.91\textwidth]{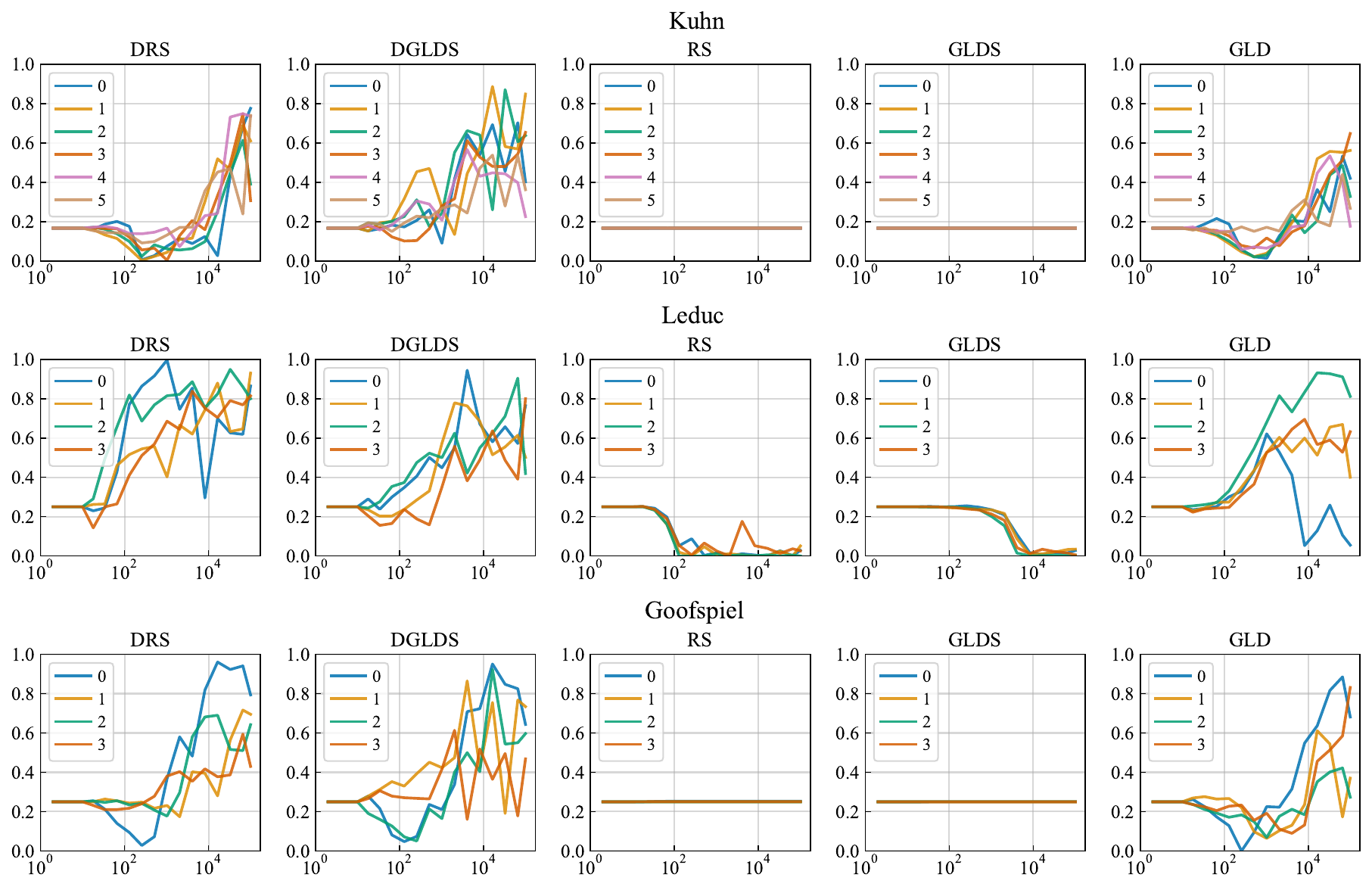}
\caption{The evolution of the hyper-parameter values of different MCs in the \textbf{Zero-Sum} category. The $y$-axis is the value of $\bm{\alpha}$. The $x$-axis is the number of iterations.}
\label{fig:alpha-sc-zero-sum}
\end{figure*}

\begin{figure*}[ht]
\centering
\includegraphics[width=0.91\textwidth]{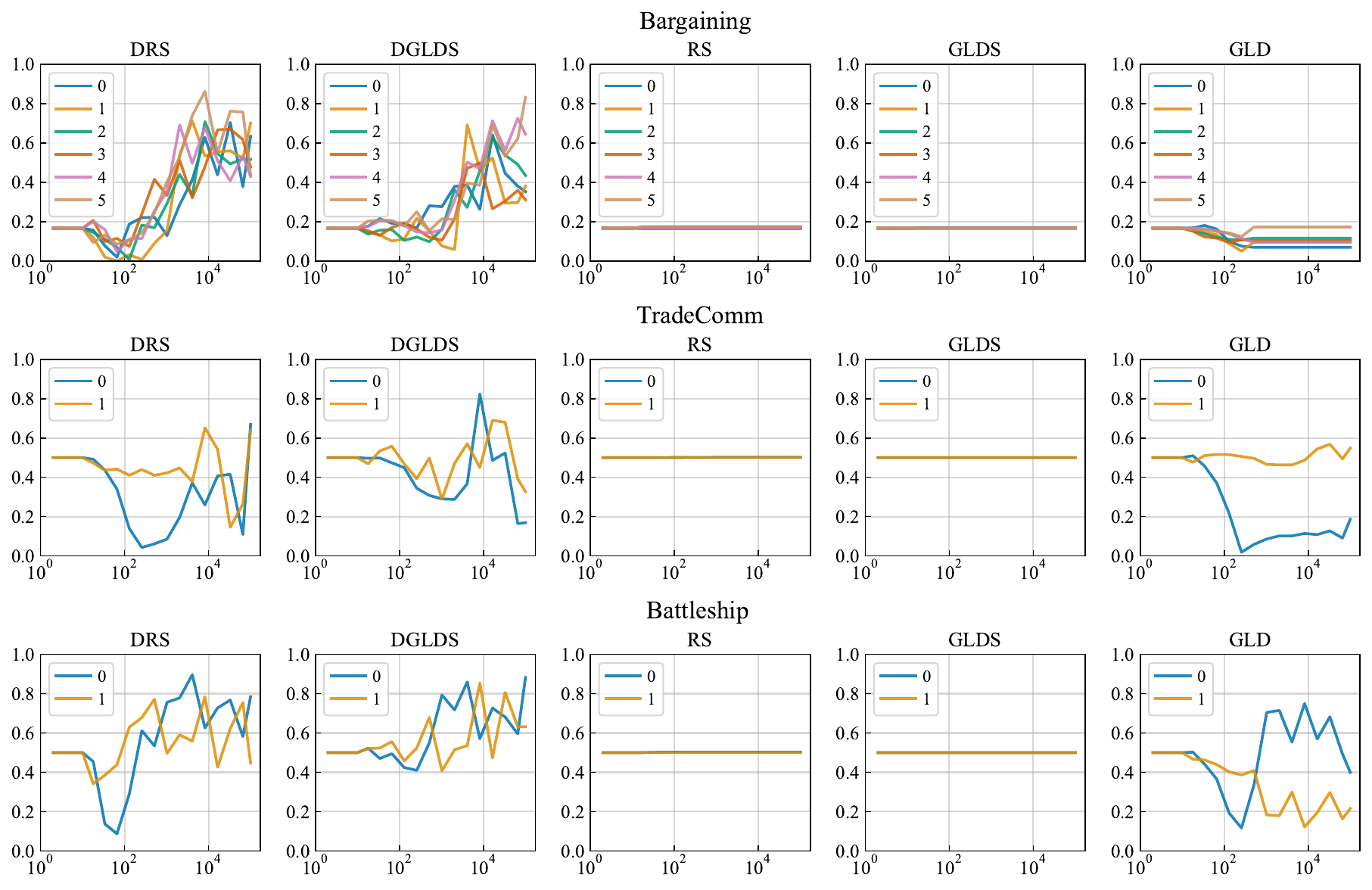}
\caption{The evolution of the hyper-parameter values of different MCs in the \textbf{General-Sum} category. The $y$-axis is the value of $\bm{\alpha}$. The $x$-axis is the number of iterations.}
\label{fig:alpha-sc-gene-sum}
\end{figure*}

\begin{figure*}[ht]
\centering
\includegraphics[width=0.91\textwidth]{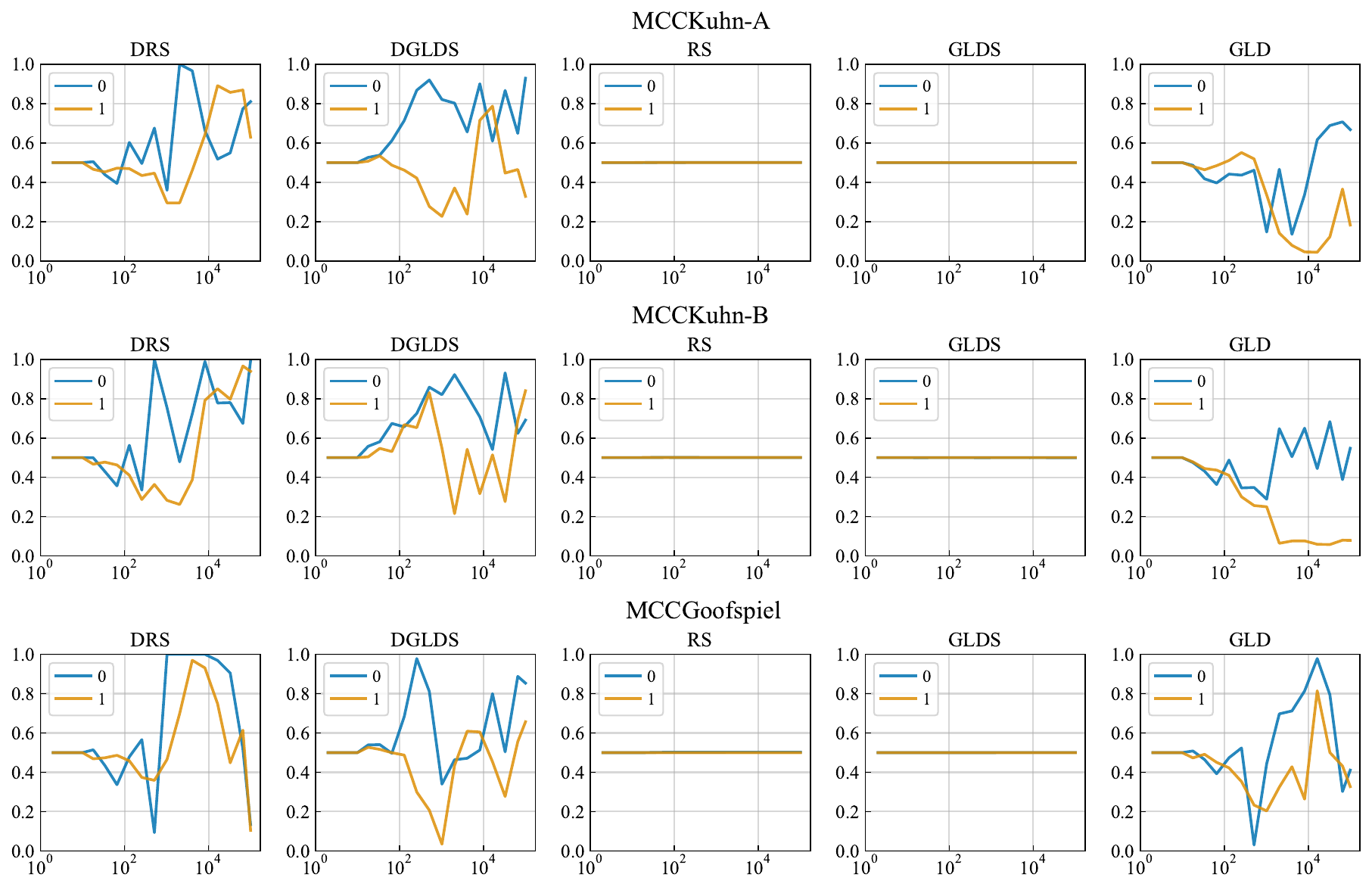}
\caption{The evolution of the hyper-parameter values of different MCs in the \textbf{MCC} category. The $y$-axis is the value of $\bm{\alpha}$. The $x$-axis is the number of iterations.}
\label{fig:alpha-sc-mcc}
\end{figure*}

\textbf{Evolution of Hyper-Parameters.} The critical observation that supports our proposed DRS is that the value of the evaluation measure $\mathcal{L}$ is extremely small and converges relatively quickly, which makes the original zero-order methods struggle in adjusting the hyper-parameters as they typically adjust the hyper-parameters directly based on the performance. To further verify this intuition, we visualize the evolution of the hyper-parameter $\bm{\alpha}$ over the learning process, which is shown in Figure~\ref{fig:alpha-sc-single-agent}--Figure~\ref{fig:alpha-sc-mcc} (respectively corresponds to single-agent, cooperative, zero-sum, general-sum, and MCC categories). We use index 0 to represent the magnet policy and the recent $M$ historical policies are indexed by $\{1, \cdots, M\}$. From the results, we can see that in all the games except Leduc, the value of $\bm{\alpha}$ determined by RS almost does not change over the learning process (the same phenomenon is observed for GLDS as it follows the same idea of RS). In Leduc, this value tends to decrease to 0 over the learning process. In other words, the regularization is vanishing, which explains why RS and GLDS cannot decrease the NashConv in this game as adding regularization has been proven important to solve two-player zero-sum games~\cite{sokota2023unified,liu2023the}. In all the games, DRS and DGLDS share some similarities in determining the value of $\bm{\alpha}$ and differ from GLD. Nevertheless, the convergence results in Figure~\ref{fig:diff-mc} show that DRS is the best choice among them as it can consistently work well across all categories of decision-making problems.

\clearpage

\clearpage
\subsection{Different Bregman Divergences\label{app:diff_bregman}}
One of the prominent features of our CMD (GMD) is that it is capable of exploring more possible Bregman divergences. In this section, we investigate how CMD performs under the different Bregman divergences induced by different convex functions in Table~\ref{tab:convex_functions} (the plots for these convex functions are shown in Figure~\ref{fig:diff-convex-function}).

\begin{figure*}[ht]
\centering
\includegraphics[width=\textwidth]{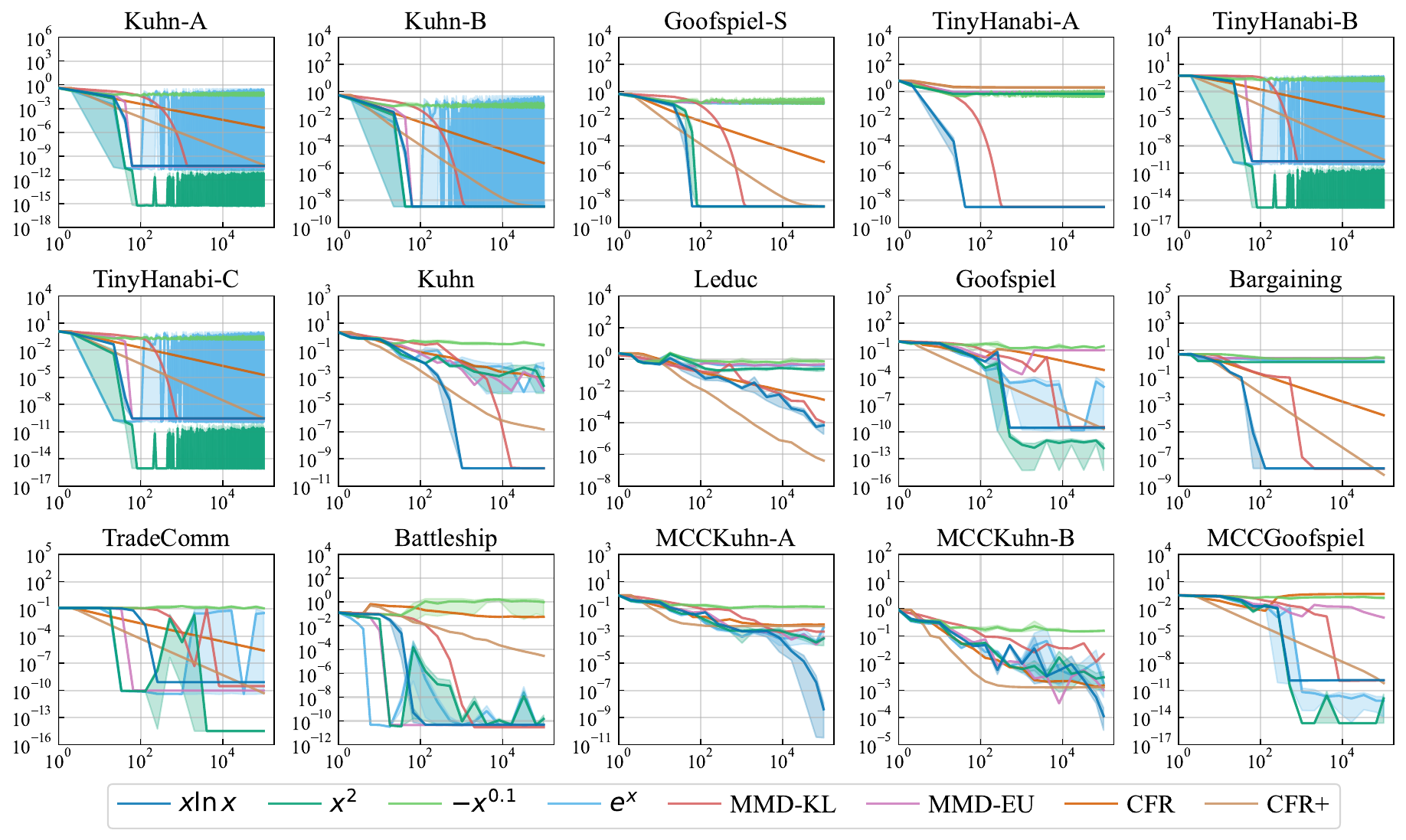}
\caption{Experimental results for different Bregman divergences. The first 6 figures correspond to the single-agent and cooperative categories where the $y$-axis is \textit{OptGap}. The rest figures correspond to other categories where the $y$-axis is \textit{NashConv}. For all the figures, the $x$-axis is the number of iterations.}
\label{fig:diff-bregman}
\end{figure*}

The experimental results are shown in Figure~\ref{fig:diff-bregman}. From the results, we can see that the entropy function $x\ln x$ is still a good choice across all the games. Nevertheless, in some games, there exist other convex functions that are better choices. For example, in Kuhn-A, TinyHanabi-B, TinyHanabi-C, Goofspiel, TradeComm, and MCCGoofspiel, $x^2$ is better than $x\ln x$. Furthermore, in MCCGoofspiel, $e^x$ is also better than $x\ln x$, which verifies that the KL divergence ($x\ln x$) or squared Euclidean norm ($x^2$) could be not always the best choice across different games. On the other hand, even under the entropy function $x\ln x$, our CMD could also outperform MMD-KL in some games such as MCCKuhn-A and MCCKuhn-B.

\clearpage
\subsection{Effectiveness of Magnet\label{app:effect_of_magnet}}
As mentioned in Section~\ref{app:gmd_connect_md}, in our experiments, when instantiating CMD, we by default add a magnet policy (the initial policy) into the policy updating as it has been demonstrated that adding a magnet policy is powerful in solving two-player zero-sum games~\cite{sokota2023unified,liu2023the}. To verify this, we conduct an ablation study where \enquote{CMD w/o Mag} denotes the method that only considers the most recent $M$ historical policies without adding the initial policy. 

\begin{figure}[ht]
\centering
\includegraphics[width=\textwidth]{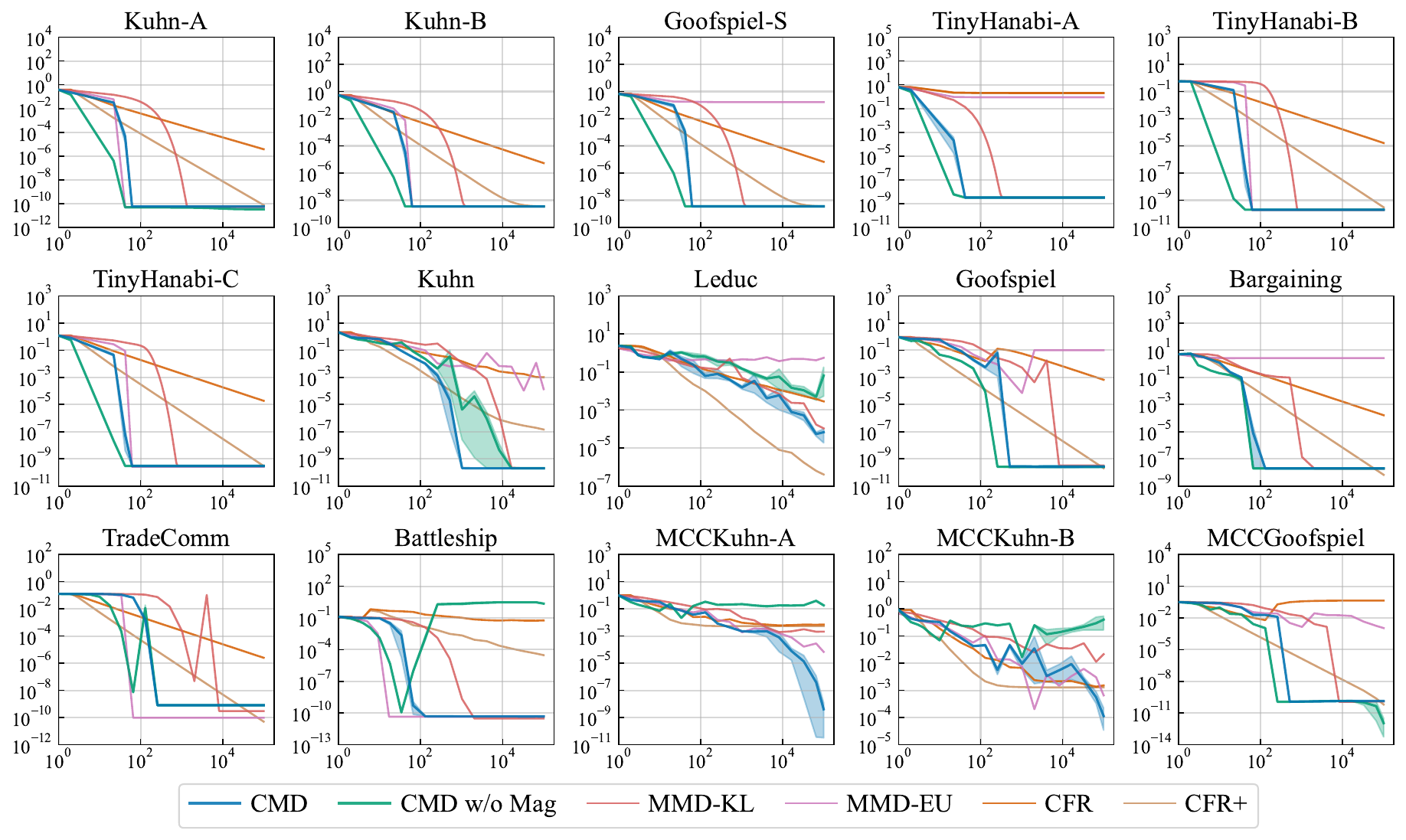}
\caption{Experimental results for the effectiveness of adding the magnet policy. The first 6 figures correspond to single-agent and cooperative categories where the $y$-axis is \textit{OptGap}. The rest figures correspond to other categories where the $y$-axis is \textit{NashConv}. For all the figures, the $x$-axis is the number of iterations.}
\label{fig:mag}
\end{figure}

The experimental results are shown in Figure~\ref{fig:mag}. From the results, we can see that i) For single-agent and cooperative categories, adding the magnet policy could result in a slightly slower convergence rate; we hypothesize that this may be due to the fact that the single-agent and cooperative games are relatively simpler than the other games (as shown in Table~\ref{tab:sel_games}, the numbers of decision points of single-agent and cooperative games are smaller than the other games). ii) For the other three categories, adding the magnet policy is necessary for CMD to work consistently well across all the games; though in MCCGoofspiel, CMD finally converges to a lower NashConv without the magnet, it could perform worse in some other games, e.g., in Battleship, it finally diverges without the magnet, and in Leduc, MCCKuhn-A, and MCCKuhn-B, it converges to a high NashConv without the magnet. Nevertheless, as pointed out in Section~\ref{app:gmd_connect_md}, this default instance of CMD (GMD) should not be confused with the original MMD even when $M=1$ as the policy updating rule is derived via a numerical method, instead of relying on the closed-form solution~\cite{sokota2023unified}.

\clearpage
\subsection{Different Measures\label{app:diff_measure}}

In this section, we apply our CMD to different solution concepts and evaluation measures (i.e., the desiderata \textbf{D3} and \textbf{D4} presented in the Introduction). Note that when running CMD for different evaluation measures, only minimal modifications are required: changing the MC's optimization objective $\mathcal{L}$. We first investigate the CCEGap~\cite{moulin1978strategically,marris2021multi} in Section~\ref{app:ccegap} and then the social welfare~\cite{davis1962externalities} in Section~\ref{app:sw}.

\subsubsection{CCEGap\label{app:ccegap}}

Note that in two-player zero-sum games, NE and CCE can be shown to be payoff equivalent~\cite{v1928theorie}. Therefore, we conduct experiments on the \textit{three-player} Kuhn and Goofspiel. We follow the same experimental pipeline of OptGap and NashConv: i) investigating the combination of $M$ and $\mu$, ii) investigating different MCs, iii) investigating different Bregman divergences, and iv) investigating the effectiveness of magnet. 

\begin{figure*}[htbp]
\centering
\includegraphics[width=0.38\textwidth]{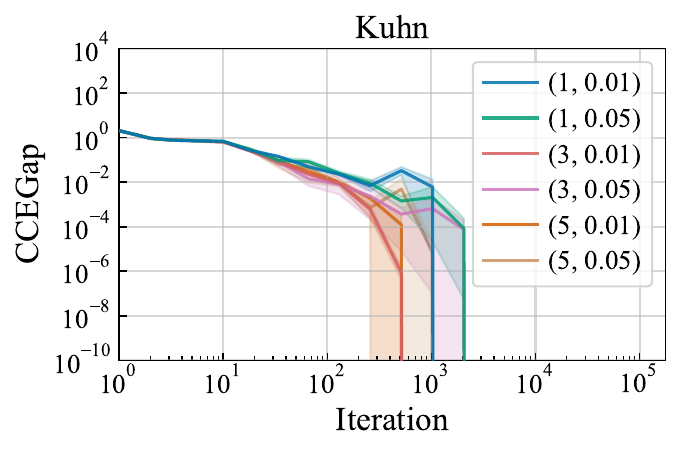}
\includegraphics[width=0.38\textwidth]{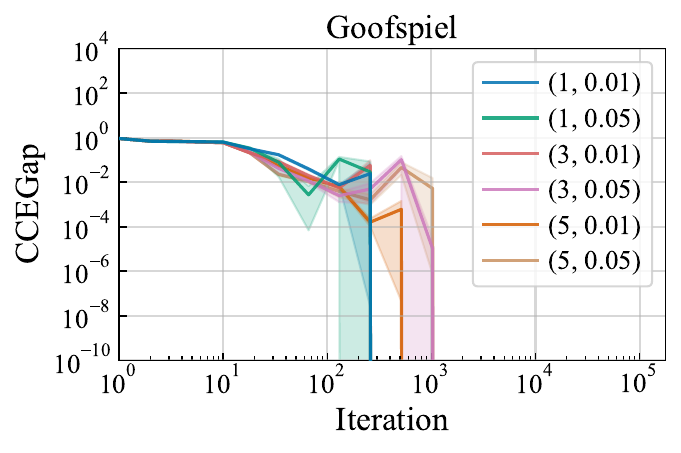}
\caption{Experimental results for the combinations of $(M,\mu)$ under the measure CCEGap.}
\label{fig:cce-K-mu}
\end{figure*}

\textbf{Number of Historical Policies and Smoothing Parameter.} We first investigate the influence of the number of previous policies $M$ and the smoothing parameter $\mu$ in DRS on the learning performance. Similar to OptGap/NashConv, we consider $M\in\{1, 3, 5\}$ and $\mu\in\{0.01, 0.05\}$, and the results are shown in Figure~\ref{fig:cce-K-mu}. We can get the same conclusion: different games may require different $M$ and $\mu$. By comparison, we determine their default values which will be fixed in the other experiments: $(M,\mu)=(3, 0.01)$ for both Kuhn and Goofspiel. All the other hyper-parameter settings are the same as OptGap/NashConv given in Table~\ref{tab:hyper-parameters}.

\begin{figure*}[htbp]
\centering
\includegraphics[width=0.38\textwidth]{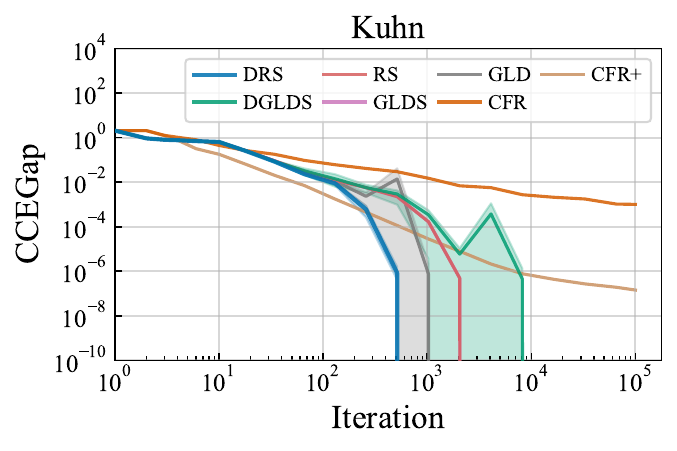}
\includegraphics[width=0.38\textwidth]{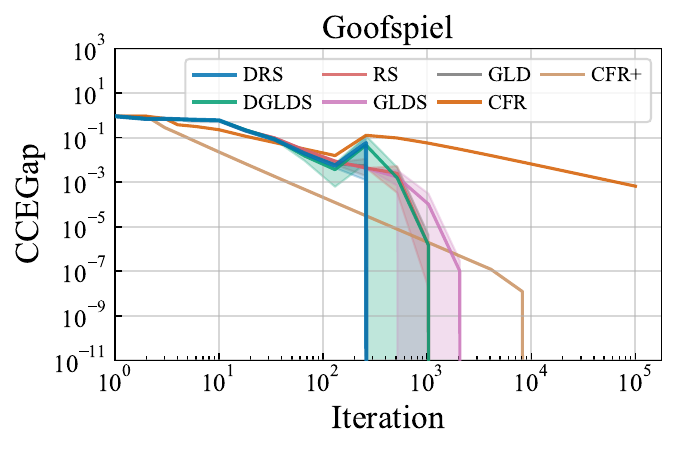}
\caption{Experimental results for different meta-controllers under the measure CCEGap.}
\label{fig:cce-mcs}
\end{figure*}

\begin{figure*}[htbp]
\centering
\includegraphics[width=0.85\textwidth]{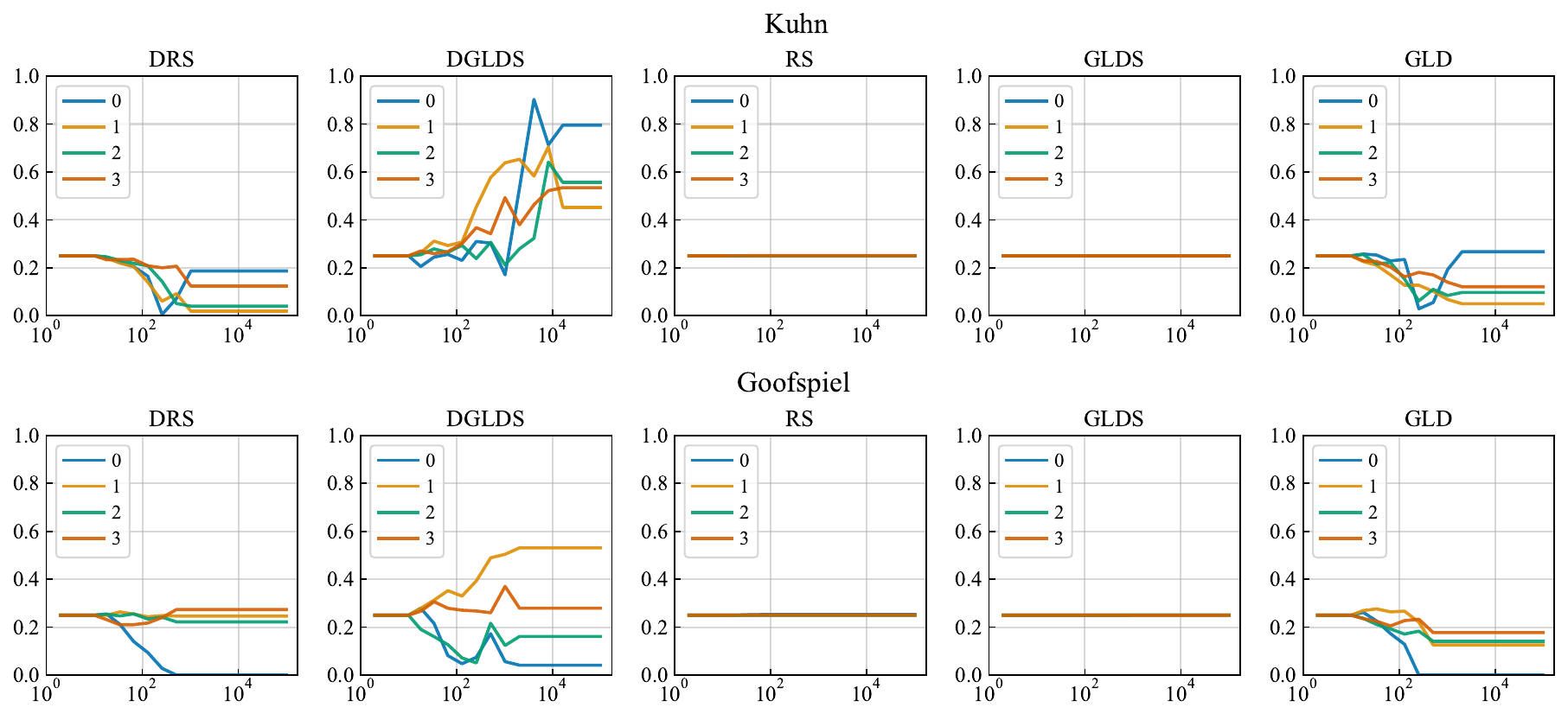}
\vspace{-8pt}
\caption{Evolution of the hyper-parameters of different MCs. The $y$-axis is the value of $\bm{\alpha}$. The $x$-axis is the number of iterations.}
\vspace{-6pt}
\label{fig:cce-alpha}
\end{figure*}

\textbf{Different Meta-Controllers.} Then, we investigate the effectiveness of different MCs. In Figure~\ref{fig:cce-mcs}, we present the learning curves of the performance of different MCs, and in Figure~\ref{fig:cce-alpha}, we present the evolution of the value of $\bm{\alpha}$ determined by different MCs over the learning process. We can get the same conclusion as OptGap/NashConv: DRS is the best choice among the 5 MCs. From the evolution of $\bm{\alpha}$ we observe that DRS and DGLDS follow two different patterns to determine the value of $\bm{\alpha}$, which is not the case for OptGap/NashConv (see Figure~\ref{fig:alpha-sc-single-agent}--Figure~\ref{fig:alpha-sc-mcc}). In contrast, we found that since GLD follows a similar pattern to DRS, it performs on par with or better than DGLDS.

\begin{figure*}[htbp]
\centering
\includegraphics[width=0.37\textwidth]{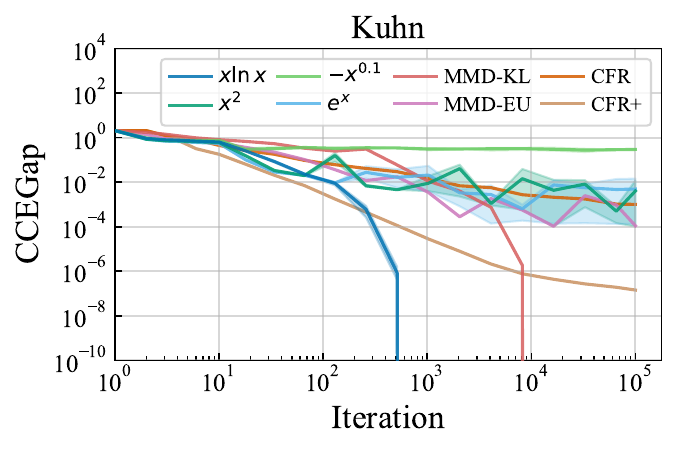}
\includegraphics[width=0.37\textwidth]{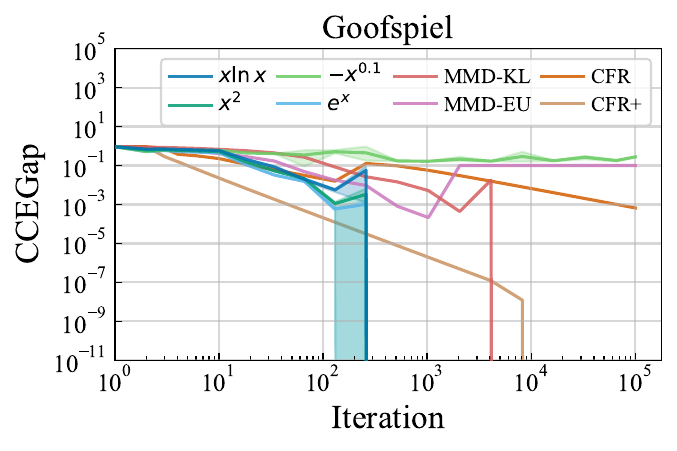}
\vspace{-8pt}
\caption{Experimental results for different Bregman divergences under the measure CCEGap.}
\vspace{-6pt}
\label{fig:cce-divergences}
\end{figure*}

\textbf{Different Bregman Divergences.} Next, we investigate how CMD performs under different Bregman divergences induced by different convex functions in Table~\ref{tab:convex_functions}, and the results are given in Figure~\ref{fig:cce-divergences}. We can get the same conclusions as for OptGap/NashConv: i) the entropy function $\psi(x)=x\ln x$ is still a good choice in different games, ii) there could exist other convex functions that are better than the entropy function, e.g., $x^2$ and $e^x$ in Goofspiel, iii) even under the entropy function $\psi(x)=x\ln x$, our CMD can converge faster than the SOTA MMD-KL in terms of the number of iterations.

\begin{figure*}[htbp]
\centering
\includegraphics[width=0.37\textwidth]{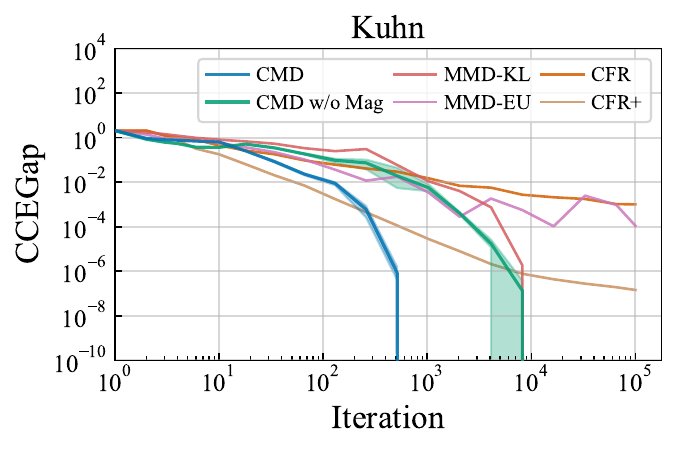}
\includegraphics[width=0.37\textwidth]{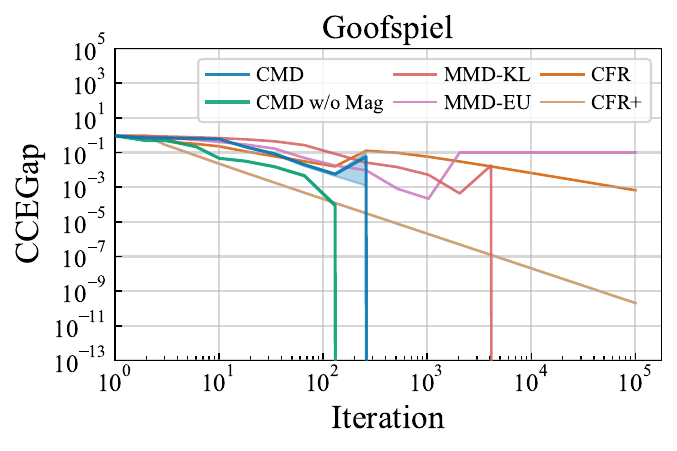}
\vspace{-8pt}
\caption{Experimental results for the effectiveness of the magnet policy under the measure CCEGap.}
\label{fig:cce-mag}
\end{figure*}

\textbf{Effectiveness of Magnet.} Finally, we investigate the effectiveness of adding the magnet policy to the policy updating, and the results are presented in Figure~\ref{fig:cce-mag}. We can observe a similar phenomenon to NashConv: in Kuhn, CMD converges remarkably faster (in terms of the number of iterations) than all the other methods, and in Goofspiel, it converges remarkably faster (in terms of the number of iterations) than all the other methods except the \enquote{CMD w/o Mag}. Nevertheless, we can still conclude that adding the magnet policy is necessary for our CMD.

\clearpage
\subsubsection{Social Welfare\label{app:sw}}
\begin{figure}[ht]
\centering
\includegraphics[width=0.3\textwidth]{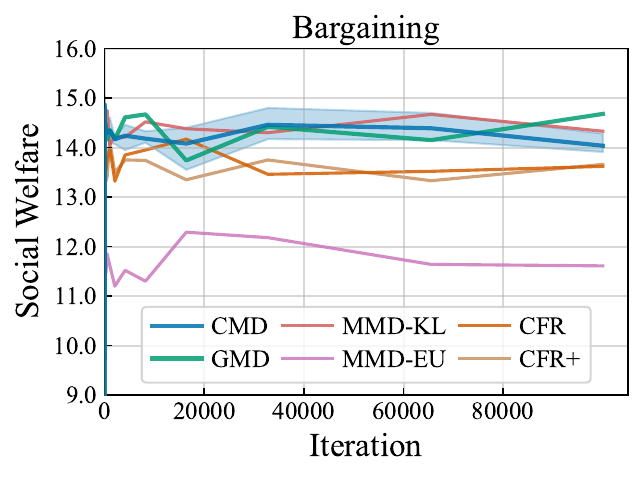}
\includegraphics[width=0.3\textwidth]{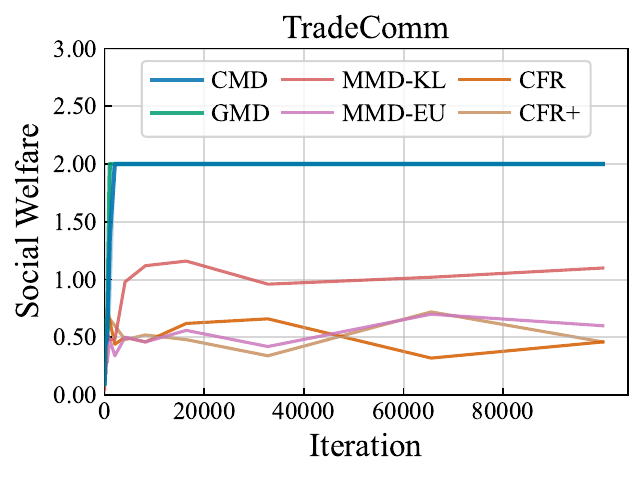}
\includegraphics[width=0.3\textwidth]{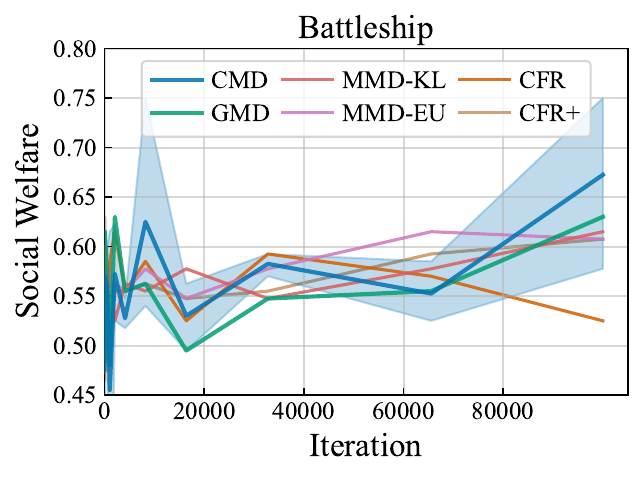}\\
\includegraphics[width=0.3\textwidth]{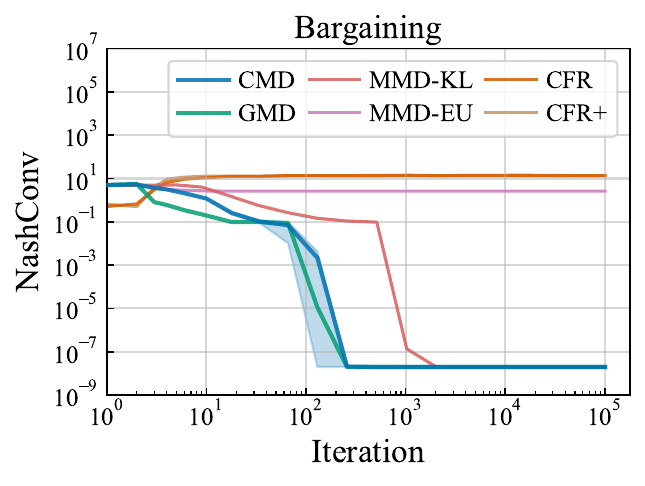}
\includegraphics[width=0.3\textwidth]{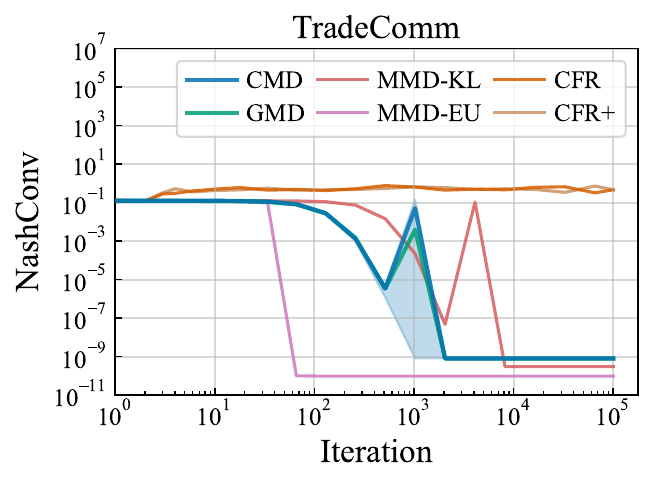}
\includegraphics[width=0.3\textwidth]{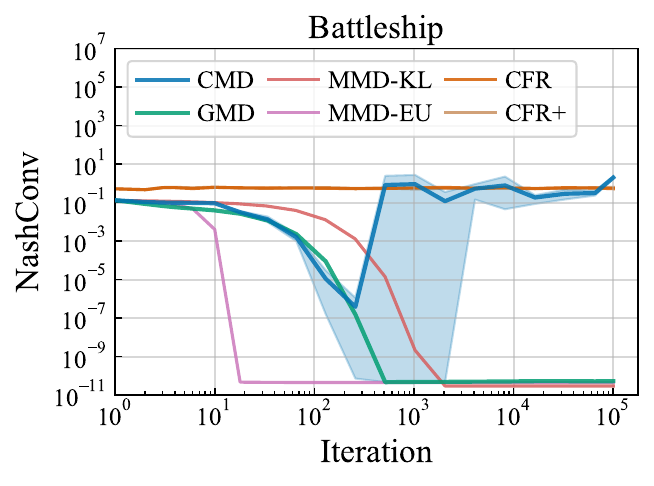}\\
\includegraphics[width=0.3\textwidth]{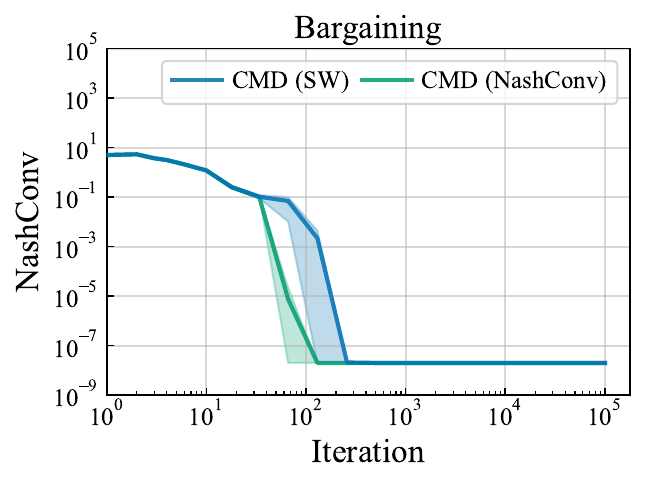}
\includegraphics[width=0.3\textwidth]{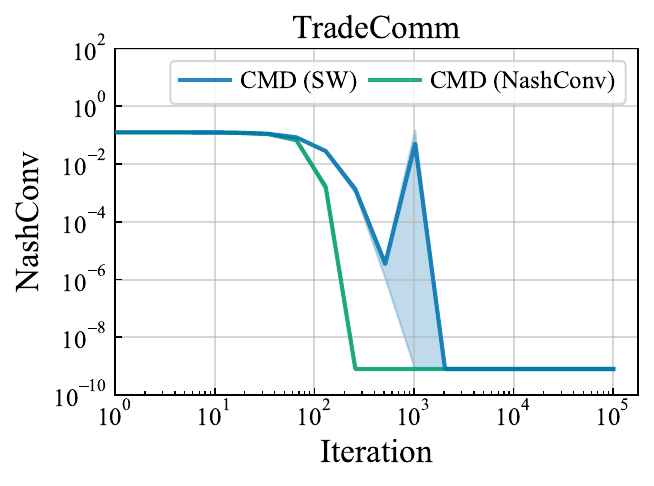}
\includegraphics[width=0.3\textwidth]{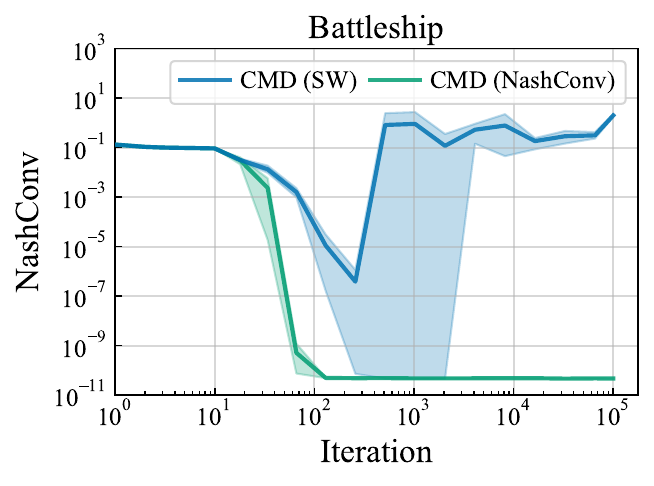}
\caption{Experimental results for the evaluation measure--social welfare.}
\label{fig:nc-sw-result}
\end{figure}

In this section, we apply our methods to the evaluation measure -- social welfare~\cite{davis1962externalities}. We conduct experiments on the general-sum games, and the results are shown in Figure~\ref{fig:nc-sw-result}. In this experiment, we use the default values in Table~\ref{tab:hyper-parameters} for the hyper-parameters.

From the top line of the figure, we can see that CMD/GMD can empirically achieve competitive or better social welfare compared to other baselines, demonstrating the effectiveness of our method when considering different measures.

In the median line of the figure, we plot the NashConv when the MC's objective $\mathcal{L}$ is social welfare. We can see that in Bargaining and TradeComm, the learning still can converge to the approximate NE even though the MC's objective is social welfare, not the NashConv. In Battleship, while CMD can get a higher (average) social welfare, the final joint policy is not an NE as its NashConv cannot converge. In other words, an efficient (in terms of social welfare) joint policy could not necessarily be an NE. This suggests one of the future directions: how to efficiently learn the \textit{NE with maximum social welfare}, which involves the equilibrium selection problem~\cite{harsanyi1988general}.

Another intuitive consequence of setting the MC's optimization objective to social welfare is that the learning could not converge to the NE or converge slower than the case where the MC's objective is directly the NashConv. As shown in the bottom line of the figure, we plot the NashConv of the two cases. In Bargaining and TradeComm, CMD with NashConv as the MC's objective can converge to the NE faster than that with SW as the MC's objective. In Battleship, CMD with SW as the MC's objective, though could achieve a higher social welfare, cannot converge the Nash equilibrium.

\clearpage
\subsection{Computational Complexity\label{app:runtime}}
In this section, we give some remarks on the computational complexity of different methods. In Section~\ref{app:running_time}, we focus on the running time of different methods, and in Section~\ref{app:memory_usage} we present the memory consumption of different methods. Note that i) these numbers are obtained under the default values of hyper-parameters (Table~\ref{tab:hyper-parameters}), and ii) these numbers are not absolute and depend on the property of the game (see Table~\ref{tab:sel_games}) and the computational resources used to run the experiments (see Section~\ref{app:exp_setup}); they only provide intuition on the computational complexity of different methods to show that our \textsc{GameBench} can satisfy the desiderata \textbf{D5} and \textbf{D6} mentioned in the Introduction.

\subsubsection{Running Time\label{app:running_time}}

The running time of different methods in different games is shown in Table~\ref{tab:runtime_all}. From the results, we can see that in most of the games, running all the algorithms does not cause a very long running time (the desiderata \textbf{D5} and \textbf{D6} presented in the Introduction), even for our methods where a set of extra operations is required: GMD requires computing the value of the dual variable via a numerical method and CMD further requires to evaluate multiple candidates of the hyper-parameters. Notably, we emphasize that: i) Our methods (CMD/GMD) provide the capability of exploring more dimensions of decision making, though they require extra computational cost (the major limitation of the current version of our methods); ii) Comparing CMD and GMD, we can see that the major cost comes from evaluating multiple samples. Therefore, as pointed out in Section~\ref{sec:conclusion}, we view this as a future direction: developing more computationally efficient hyper-parameter value updating methods without sacrificing performance. In this regard, other techniques such as Bayesian optimization~\cite{lindauer2022smac3} or offline hyper-parameter optimization approaches~\cite{chen2022towards} may be required.

\begin{table}[ht]
    \centering
    \caption{The running time of \textbf{one iteration} of different methods in different games (second).}
    \label{tab:runtime_all}
    \vskip 0.1in
    \begin{tabular}{lcccccccccc}
    \toprule
    \multirow{2}{*}{Game}&\multirow{2}{*}{CFR}&\multirow{2}{*}{CFR+}&\multirowcell{2}{MMD\\-KL}&\multirowcell{2}{MMD\\-EU}&\multirow{2}{*}{GMD}&\multicolumn{5}{c}{CMD}\\
    \cmidrule(r){7-11}
    &  &  & & & & DRS & RS & DGLDS & GLDS & GLD \\
    \midrule
    Kuhn-A       & 0.0004 & 0.0004 & 0.0003 & 0.0003 & 0.0034 & 0.0372 & 0.0372 & 0.0204 & 0.0204 & 0.0203\\ 
    Kuhn-B       & 0.0004 & 0.0004 & 0.0003 & 0.0003 & 0.0033 & 0.0370 & 0.0365 & 0.0202 & 0.0201 & 0.0202\\ 
    Goofspiel-S  & 0.0007 & 0.0006 & 0.0004 & 0.0004 & 0.0046 & 0.0491 & 0.0489 & 0.0269 & 0.0275 & 0.0270\\
    \midrule
    TinyHanabi-A & 0.0006 & 0.0006 & 0.0004 & 0.0004 & 0.0045 & 0.0474 & 0.0472 & 0.0262 & 0.0268 & 0.0260\\ 
    TinyHanabi-B & 0.0004 & 0.0004 & 0.0003 & 0.0003 & 0.0033 & 0.0366 & 0.0370 & 0.0198 & 0.0192 & 0.0197\\ 
    TinyHanabi-C & 0.0004 & 0.0004 & 0.0003 & 0.0003 & 0.0032 & 0.0364 & 0.0359 & 0.0195 & 0.0195 & 0.0199\\ 
    \midrule
    Kuhn         & 0.0084 & 0.0082 & 0.0022 & 0.0021 & 0.0267 & 0.4102 & 0.4058 & 0.2273 & 0.2282 & 0.2288\\ 
    Leduc        & 0.0942 & 0.0961 & 0.0422 & 0.0412 & 0.5146 & 6.8897 & 6.9749 & 3.8188 & 3.8116 & 3.8744\\ 
    Goofspiel    & 0.0072 & 0.0073 & 0.0014 & 0.0014 & 0.0167 & 0.2879 & 0.2899 & 0.1636 & 0.1610 & 0.1625\\ 
    \midrule
    Bargaining   & 0.0279 & 0.0273 & 0.0130 & 0.0116 & 0.1093 & 1.5311 & 1.5308 & 0.8428 & 0.8579 & 0.8516\\ 
    TradeComm    & 0.0028 & 0.0029 & 0.0011 & 0.0010 & 0.0121 & 0.1704 & 0.1699 & 0.0957 & 0.0942 & 0.0939\\ 
    Battleship   & 0.0245 & 0.0248 & 0.0097 & 0.0094 & 0.1125 & 1.5570 & 1.5771 & 0.8833 & 0.8774 & 0.8835\\
    \midrule
    MCCKuhn-A    & 0.0083 & 0.0084 & 0.0021 & 0.0021 & 0.0264 & 6.9054 & 6.8377 & 4.0453 & 4.1461 & 4.1034\\ 
    MCCKuhn-B    & 0.0080 & 0.0083 & 0.0021 & 0.0021 & 0.0260 & 6.8100 & 6.7847 & 4.1047 & 4.0944 & 4.1104\\ 
    MCCGoofspiel & 0.0070 & 0.0073 & 0.0014 & 0.0014 & 0.0167 & 5.3541 & 5.3852 & 3.1817 & 3.1945 & 3.1836\\
    \bottomrule
    \end{tabular}
\end{table} 

\clearpage
\subsubsection{Memory Usage\label{app:memory_usage}}
The memory usage of different methods in different games is provided in Table~\ref{tab:memory_all}. From the results, we can see that running all the algorithms does not cause much memory consumption, which shows that our \textsc{GameBench} is academic-friendly.
\begin{table}[ht]
    \centering
    \caption{The memory usage of different methods in different games (MB).}
    \label{tab:memory_all}
    \vskip 0.1in
    \begin{tabular}{lcccccccccc}
    \toprule
    \multirow{2}{*}{Game}&\multirow{2}{*}{CFR}&\multirow{2}{*}{CFR+}&\multirowcell{2}{MMD\\-KL}&\multirowcell{2}{MMD\\-EU}&\multirow{2}{*}{GMD}&\multicolumn{5}{c}{CMD}\\
    \cmidrule(r){7-11}
    &  &  & & & & DRS & RS & DGLDS & GLDS & GLD \\
    \midrule
    Kuhn-A       & 0.8750 & 0.9805 & 0.3711 & 0.3750 & 0.9492 & 1.0352 & 1.0859 & 0.3750 & 0.3750 & 0.3750\\ 
    Kuhn-B       & 0.8672 & 0.8672 & 0.4297 & 0.3750 & 0.7461 & 1.1289 & 1.0352 & 0.3750 & 0.3750 & 0.3164\\ 
    Goofspiel-S  & 1.1875 & 1.2969 & 1.2617 & 1.2539 & 1.2578 & 1.2578 & 1.2578 & 1.6992 & 1.6953 & 1.6992\\
    \midrule
    TinyHanabi-A & 1.0352 & 1.0156 & 0.4805 & 0.4258 & 0.4883 & 1.0312 & 1.1836 & 0.4453 & 0.4297 & 0.4883\\ 
    TinyHanabi-B & 0.9922 & 1.0898 & 0.4922 & 0.4844 & 0.4336 & 0.4922 & 0.5430 & 0.4922 & 0.4297 & 0.4922\\ 
    TinyHanabi-C & 0.9805 & 0.9922 & 0.4336 & 0.4297 & 0.4336 & 0.4922 & 0.4336 & 0.4336 & 0.4336 & 0.4336\\ 
    \midrule
    Kuhn         & 1.9961 & 2.0078 & 3.0352 & 3.1367 & 3.5586 & 3.7734 & 3.7227 & 3.5156 & 3.5352 & 3.5273\\ 
    Leduc        & 26.262 & 26.344 & 51.664 & 52.465 & 58.273 & 59.063 & 58.555 & 52.688 & 51.949 & 51.359\\ 
    Goofspiel    & 2.4844 & 2.4297 & 3.5039 & 3.4961 & 4.3555 & 4.3359 & 4.3047 & 4.0391 & 4.0430 & 4.0898\\ 
    \midrule
    Bargaining   & 10.633 & 10.578 & 24.816 & 24.852 & 35.129 & 35.422 & 35.020 & 31.941 & 32.344 & 31.781\\ 
    TradeComm    & 1.4961 & 1.5430 & 2.0156 & 2.0703 & 2.0664 & 2.0078 & 2.0117 & 2.2461 & 2.2461 & 2.3633\\ 
    Battleship   & 6.9102 & 6.9023 & 13.539 & 13.422 & 13.543 & 13.543 & 13.484 & 12.098 & 12.148 & 12.332\\
    \midrule
    MCCKuhn-A    & 2.5742 & 2.5195 & 2.0312 & 2.0273 & 2.5586 & 2.6719 & 2.7266 & 2.3047 & 2.2930 & 2.2930\\ 
    MCCKuhn-B    & 2.4688 & 2.5703 & 2.0273 & 1.9648 & 1.9727 & 2.6562 & 2.7617 & 2.2383 & 2.2305 & 2.1797\\ 
    MCCGoofspiel & 2.7500 & 2.7500 & 3.0078 & 3.0820 & 3.0781 & 3.0195 & 3.1289 & 2.8203 & 2.8047 & 2.8672\\
    \bottomrule
    \end{tabular}
\end{table}


\end{document}